\documentclass[11pt,onecolumn,a4paper]{article}

\usepackage{amsmath}
\usepackage{amsfonts}
\usepackage{amssymb}
\usepackage{amsthm}
\usepackage{commath}
\usepackage{caption}
\usepackage[ansinew]{inputenc}
\usepackage{graphicx,color}
\usepackage{url}
\usepackage{latexsym,enumerate}
\usepackage{booktabs}
\usepackage{framed}
\usepackage{subfigure}
\usepackage{multirow} 
\usepackage{color}
\usepackage{colortbl}
\definecolor{gray1}{rgb}{0.8,0.8,0.8}
\definecolor{gray2}{rgb}{0.95,0.95,0.95}

\newcommand{\rb}[1]{\raisebox{1.5ex}[-1.5ex]{#1}}

\newcommand{\cF}{{\mathcal F}}

\newcommand{\cB}{{\mathcal B}}

\newcommand{\cR}{{\mathcal R}}
\newcommand{\cH}{{\mathcal H}}
\newcommand{\pvc}{{*_{pv}}} %principal value convolution

\theoremstyle{plain}

\theoremstyle{definition}

\theoremstyle{remark}

\begin{document}

\title{Filter Design and Performance Evaluation for Fingerprint Image Segmentation}

\date{}%
\author{Duy Hoang Thai\thanks{D.H. Thai, S. Huckemann, and C. Gottschlich are with the Institute for Mathematical Stochastics, 
University of G\"ottingen, Germany.
S. Huckemann and C. Gottschlich are also with the Felix-Bernstein-Institute for Mathematical Statistics in the Biosciences,
University of G\"ottingen, Germany.
Email: \{duy.thai, huckeman, gottschlich\}@math.uni-goettingen.de%
}%
, Stephan Huckemann, and Carsten Gottschlich}%

\maketitle

\begin{abstract}

Fingerprint recognition plays an important role in many commercial applications and is used by millions of people every day, e.g. for unlocking mobile phones. Fingerprint image segmentation is typically the first processing step of most fingerprint algorithms and it divides an image into foreground, the region of interest, and background. Two types of error can occur during this step which both have a negative impact on the recognition performance: 'true' foreground can be labeled as background and features like minutiae can be lost, or conversely 'true' background can be misclassified as foreground and spurious features can be introduced. The contribution of this paper is threefold: firstly, we propose a novel factorized directional bandpass (FDB) segmentation method  for texture extraction based on the directional Hilbert transform of a Butterworth bandpass (DHBB) filter interwoven with soft-thresholding. Secondly, we provide a manually marked ground truth segmentation for 10560 images as an evaluation benchmark. Thirdly, we conduct a systematic performance comparison between the FDB method and four of the most often cited fingerprint segmentation algorithms showing that the FDB segmentation method clearly outperforms these four widely used methods. The benchmark and the implementation of the FDB method are made publicly available.

\end{abstract}

\section*{Keywords}
Fingerprint recognition, fingerprint image segmentation, evaluation benchmark, manually marked ground truth,
directional Hilbert transform, Riesz transform, Butterworth bandpass filter, texture extraction

\section{Introduction}

Nowadays, fingerprint recognition is used by millions of people 
in their daily life for verifying a claimed identity 
in commercial applications ranging from check-in at work places or libraries,
access control at amusement parks or zoos to unlocking notebooks, tablets or mobile phones.
Most fingerprint recognition systems are based on minutiae 
as features for comparing fingerprints \cite{HandbookFingerprintRecognition2009}.
Typical processing steps prior to minutiae extraction are 
fingerprint segmentation, orientation field estimation and image enhancement. 
The segmentation step divides an image into foreground, the region of interest (ROI), and background.
 Two types of error can occur in this step 
and both have a negative impact on the recognition rate:
'true' foreground can be labelled as background and features like minutiae can be lost,
or 'true' background can be misclassified as foreground and spurious features may be introduced. It is desirable to have a method that controls both errors.

% our contribution
\subsection{The Factorized Directional Bandpass Method,  Benchmark and Evaluation} 
In order to balance both errors we take the viewpoint that -- loosely speaking -- fingerprint images are highly determined by patterns that have frequencies only
in a specific band of the Fourier spectrum (prior knowledge). Focusing on these \emph{frequencies occuring in true fingerprint images} (FOTIs), we aim at the following goals:
\begin{enumerate}
 \item[1)] Equally preserving all FOTIs while attenuating all non-FOTIs.
\item[2)] Removing all image artifacts in the FOTI spectrum, not due to the true fingerprint pattern.
\item[3)] Returning a (smooth) texture image containing only FOTI features from the true fingerprint pattern.
 \item[4)] Morphological methods returning the ROI.
\end{enumerate}
In order to meet these goals we have developed a \emph{factorized directional bandpass} (FDB) segmentation method. 
\paragraph{The FDB method}
At the core of the FDB method is a classical Butterworth bandpass filter which guarantees Goal 1. Notably Goal 1 cannot fully be met by Gaussian based filtering methods such as the Gabor filter. Obviously, due to the Gaussian bell shaped curve, FOTIs would not be filtered alike. Because straightforward Fourier methods cannot cope with
curvature (as could e.g. curved Gabor filters \cite{Gottschlich2012}) we perform separate filtering into a few isolated orientations only, via directional Hilbert transformations. The composite \emph{directional Hilbert Butterworth bandpass filter} (DHBB) incorporates our prior knowledge about the range of possible values 
of ridge frequencies (between $1/3$ and $1/25$ pixels) 
or interridge distances (between 3 and 25 pixels) \cite{Gottschlich2012},
assuming a sensor resolution of 500 DPI and that adult fingerprints are processed.
In the case of adolescent fingerprints \cite{GottschlichHotzLorenzBernhardtHantschelMunk2011}
or sensors with a different resolution, the images can be resized 
to achieve an age and sensor independent size -- not only for the first segmentation step,
but also for all later processing stages. 
Our parameters can be tuned to reach an optimal tradeoff between treating all realistic frequencies alike and avoiding Gibbs effects. Moreover we use a data friendly rectangular spectral shape of the bandpass filter employed which preserves the rectangular shape of the spatial image.

A second key ingredient is the factorization of the filter into two factors in the spectral domain, between which a thresholding operation is inserted. After preserving all FOTIs and removing all non-FOTIs in 
application of the first factor, all FOTI features not due to the true fingerprint pattern (which are usually less pronounced) 
are removed via a shrinkage operator: soft-thresholding. Note that albeit removing less pronounced FOTI features, thresholding introduces new unwanted high frequencies. These are removed, however, by application of the second factor, which also compensates for a possible phase shift due to the first factor, thus producing a smoothed image with pronounced FOTI features only.

At this stage, non-prominent FOTI features have been removed, not only outside the ROI, but also some due to true fingerprint features inside the ROI. In the final step, these ``lost'' regions are restored via morphological operations (convex hull after binarization and two-scale  opening and closing).

The careful combination of the above ingredients in our proposed FDB method yields segmentation results far superior to existing segmentation methods. 

\paragraph{Benchmark}
In order to verify this claim, because of the lack of a suitable benchmark in the literature,
we provide a manually marked ground truth segmentation for all 12 databases 
of FVC2000 \cite{FVC2000}, FVC2002 \cite{FVC2002} and FVC2004 \cite{FVC2004}.
Each databases consists of 80 images for training and 800 images for testing. 
Overall this benchmark consists of 10560 marked segmentation images.
This ground truth benchmark is made publicly available, 
so that other researchers can evaluate segmentation algorithms on it.

\paragraph{Evaluation against existing methods}
We conduct a systematic performance comparison of widely used segmentation algorithms on this benchmark. In total, more than 100 methods for fingerprint segmentation can be found in literature. 
However, it remains unclear how these methods compare with each other in terms of segmentation performance and
which methods can be considered as state-of-the-art.
In order to remedy the current situation
we chose four of the most often cited fingerprint segmentation methods and compared their performance:
a method based on mean and variance of gray level intensities 
and the coherence of gradients as features 
and a neural network as a classifier \cite{BazenGerez2001}, 
a method using Gabor filter bank responses~\cite{ShenKotKoo2001},
a Harris corner response based method~\cite{WuTulyakovGovindaraju2007} 
and an approach using local Fourier analysis~\cite{ChikkerurCartwrightGovindaraju2007}.

\subsection{Related Work}

Early methods for fingerprint segmentation include Mehtre \textit{et al.} \cite{MehtreMurthyKapoorChatterjee1987}
who segment an image based on histograms of local ridge orientation
and in \cite{MehtreChatterjee1989} additionally the gray-level variance is considered.
A method proposed by Bazen and Gerez \cite{BazenGerez2001} uses the local mean and variance of gray-level intensities 
and the coherence of gradients as features and a neural network as a classifier.
Similarly Chen \textit{et al.} \cite{ChenTianChengYang2004} use block based features 
including the mean and variance in combination with a linear classifier.
Both methods perform morphology operations for postprocessing.
A method by Shen \textit{et al.} is based on Gabor filter bank responses of blocks \cite{ShenKotKoo2001}.
In \cite{Gottschlich2012}, all pixels are regarded as foreground 
for which a valid ridge frequency based on curved regions can be estimated.
Wu \textit{et al.} \cite{WuTulyakovGovindaraju2007} proposed a Harris corner response based method  
and they apply Gabor responses for postprocessing. 
Wang \textit{et al.} \cite{WangSuoDai2005} proposed to use Gaussian-Hermite moments 
for fingerprint segmentation.
The method of Zhu \textit{et al.} \cite{ZhuYinHuZhang2006} uses 
a gradient based orientation estimation as the main feature,
and a neural network detects wrongly estimated orientation and classifies the corresponding blocks as background.
Chikkerur \textit{et al.} \cite{ChikkerurCartwrightGovindaraju2007} applied 
local Fourier analysis for fingerprint image enhancement. 
The method performs implicitly fingerprint segmentation, orientation field and ridge frequency estimation.
Further approaches for fingerprint enhancement in the Fourier domain include 
Sherlock \textit{et al.} \cite{SherlockMonroMillard1994}, 
Sutthiwichaiporn and Areekul \cite{SutthiwichaipornAreekul2013} and 
Bart\r{u}n\v{e}k \textit{et al.} 
\cite{BartunekNilssonNordbergClaesson2006,BartunekNilssonNordbergClaesson2008,BartunekNilssonSallbergClaesson2013}.

\subsection{Setup of Paper}
The paper is organized as follows: 
in the next section, we describe the proposed method
beginning with the design of the DHBB filter for texture extraction in Section \ref{secFilterDesign}.
Subsequently, the extracted and denoised texture is utilized for estimating the segmentation 
as described in \ref{secSegmentation} which summarizes the FDB segmentation procedure.
In Section \ref{secBenchmarkResults}, the manually marked ground truth benchmark is introduced 
and applied for evaluating the segmentation performance of four widely used algorithms
and for comparing them to the proposed FDB segmentation method.
The results are discussed in Section \ref{secConclusions}.

\begin{figure*} 
\begin{center}
  \includegraphics[width=0.8\textwidth]{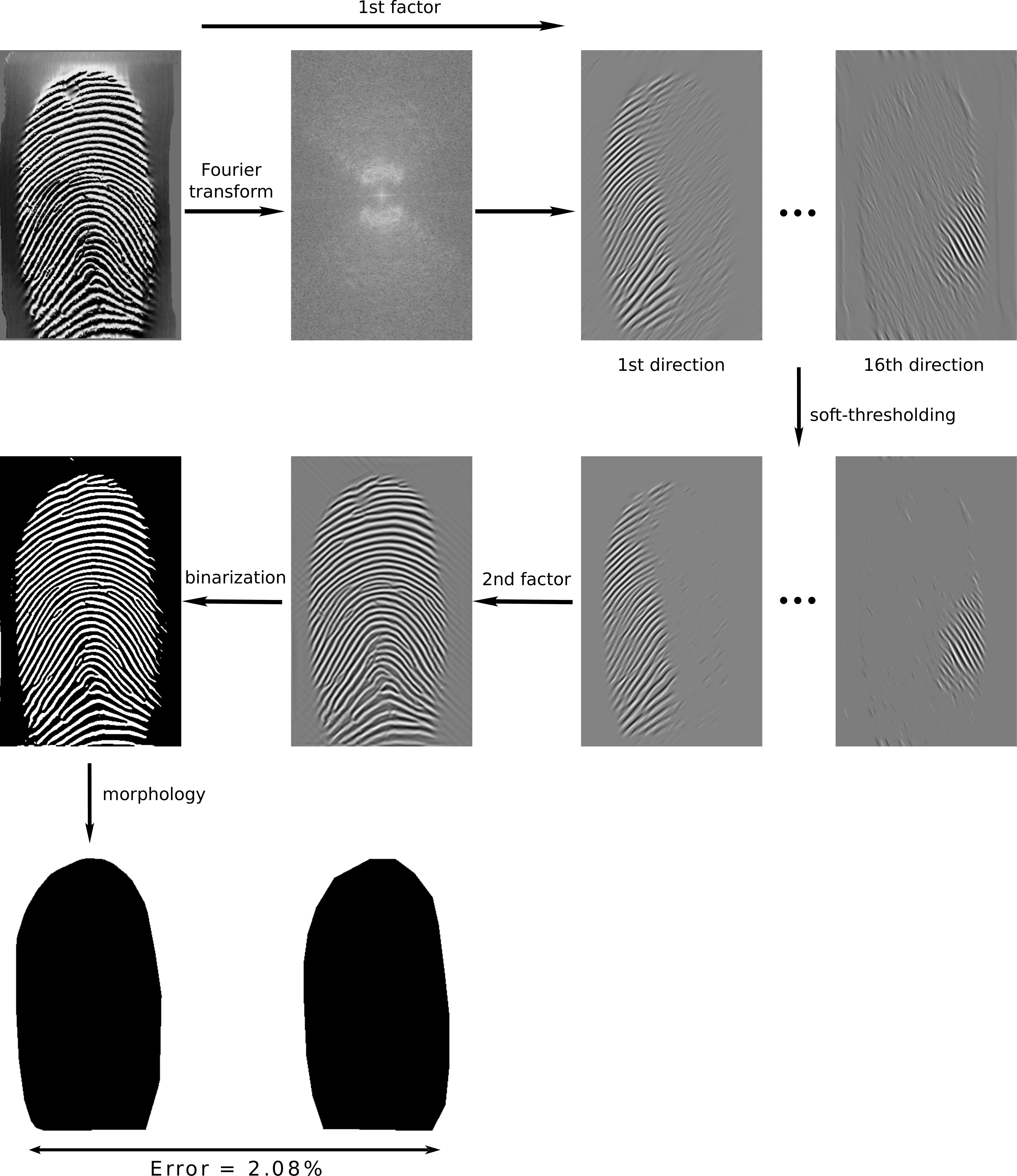}
  \caption{
           Overview over the segmentation by the FDB method:           
           In the analysis step, the original image (top row, left) is transformed into the Fourier domain (second column) and filtered by the first DHBB factor obtaining 16 directional subbands (third and fourth columns). 
           Next soft-thresholding is applied to 
           remove spurious patterns (second row, third and fourth columns). 
           In the synthesis step, the feature image (second column) is reconstructed 
           from these subbands using the second DHBB factor.
           Finally, the feature image is binarized and the ROI is obtained by morphological operations.
           The estimated ROI (third row, left) is compared to manually marked ground truth segmentation (third row, right)
           in order to evaluate the segmentation performance.
	   \label{figAlgorithmOverview}
          }    
\end{center}
\end{figure*}

\section{Fingerprint Segmentation by FDB Methods}

Our segmentation method uses a filter transforming an input 2D image $f(\cdot)\in L_2(\mathbb R^2)$ into a \emph{feature image}  
\begin{equation} \label{eq8}
 \tilde f(\boldsymbol x) ~:=~
 \sum_{l=0}^{L-1} \sum_{\boldsymbol x \neq \boldsymbol m \in \mathbb Z^2} 
 \underbrace{ \mathbf T \bigg\{
 \underbrace{ \langle f(\cdot), \phi_l^{\gamma, n}(\cdot - \boldsymbol m) \rangle^{pv}_{L_2} }_{ c_l[\boldsymbol m] } , \beta \bigg\} }_
 { d_l[\boldsymbol m] }
 \cdot \phi_l^{\gamma, n} (\boldsymbol x - \boldsymbol m).
\end{equation}
Due to our filter design, the $L_2$ product above as well as all convolutions, integrals and sums are understood in the principal value sense
$$\lim_{\epsilon \to 0} \int_{\|\boldsymbol y - \boldsymbol m\|\geq \epsilon} f(\boldsymbol y) 
       \cdot \phi_l^{\gamma, n} (\boldsymbol y - \boldsymbol m) d \boldsymbol y
        ~=~ \langle f(\cdot), \phi_l^{\gamma, n}(\cdot - \boldsymbol m) \rangle^{pv}_{L_2} ~=~
(f\pvc \phi_l^{\gamma, n,\vee})(\boldsymbol m)\,.$$
Having clarified this, the symbol ``pv'' will be dropped in the following.
At the core of (\ref{eq8}) is the DHBB filter conveyed by $\phi_l^{\gamma, n}$ ($l$ counts directions, $n$ and $\gamma$ are tuning parameters providing sharpness). In fact, we suitably factorize the filter conveyed by $\phi_l^{\gamma, n}*\phi_l^{\gamma, n,\vee}$ in the Fourier domain 
 where $\phi^{\gamma,n,\vee}_l(\boldsymbol x) := \phi^{\gamma,n}_l(-\boldsymbol x)$ with the argument reversion operator ``$^\vee$'' and apply a thresholding procedure $\mathbf T$ ``in the middle''. Underlying this factorization is a factorization of the bandpass filter involved. The precise filter design will be detailed in the following.  
Note that the directional Hilbert transform is also conveyed by a non-symmetric kernel. 
Reversing this transform (as well as the factor of the Butterworth) 
restores symmetry. 
It is inspired 
by the steerable wavelet \cite{UnserVandeville2010,HeldStorathMassopustForster2010,UnserChenouardVandeville2011}
and to some extend similar in spirit to the curvelet transform \cite{MaPlonka2010}, \cite{CandesDemanetDonohoYing2006} 
and the curved Gabor filters \cite{Gottschlich2012}. We deal with curvature by analyzing single directions~$l$ separately before the final synthesis.

Via factorization, possible phase shifts are compensated and unwanted frequencies introduced by the thresholding operator are eliminated, yielding a sparse smoothed feature image. This allows for easy binarization and segmentation via subsequent morphological methods, leading to the ROI. 
 
Note that (\ref{eq8}) can be viewed as an analog to the projection operator in sampling theory with the analysis and synthesis steps (e.g. \cite{Unser2000}). In this vein we have the following three steps:

\begin{enumerate}
 \item[] \emph{Forward analysis (prediction)}: A first application of the argument reversed DHBB filter 
 to a fingerprint image $f$ corresponds to a number of directional selections 
 in certain frequency bands of the fingerprint image giving $c_l[\boldsymbol m]$ above. 
 \item[] \emph{Proximity operator (thresholding)}: In order to remove intermediate coefficients 
 due to spurious patterns (cf. \cite{HaltmeierMunk2014}) we perform soft tresholding on the filtered grey values yielding $d_l[\boldsymbol m]$ above. 
 \item[] \emph{Backward synthesis}: Subsequently we apply the filter (non-reversed) again giving $\tilde f$ assembled from all subbands. 
 A numerical comparison to other synthesis methods, summation 
 (corresponding to a naive reconstruction) and maximal response in the appendix \ref{AppendixC}, shows the superiority of this smoothing step. 
\end{enumerate}

 Due to the discrete nature of the image $f[\boldsymbol k] = f(\boldsymbol x) \mid_{\boldsymbol x = \boldsymbol k \in \mathbb Z^2}$, we work with
 the discrete version of $\tilde f (\boldsymbol x)$ at $\boldsymbol x = \boldsymbol k \in \mathbb Z^2$
 in Eq. (\ref{eq8}).

\subsection{Filter Design for Fingerprint Segmentation} \label{secFilterDesign}

\begin{figure*}[ht] 
\begin{center}
    \subfigure[$\widehat h_l^n(\boldsymbol \omega)$]{ \includegraphics[width=0.22\textwidth]{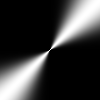} }		
    \subfigure[ 1D: $\widehat g^\gamma(\omega)$ ]{ \includegraphics[width=0.22\textwidth,height=0.22\textwidth]{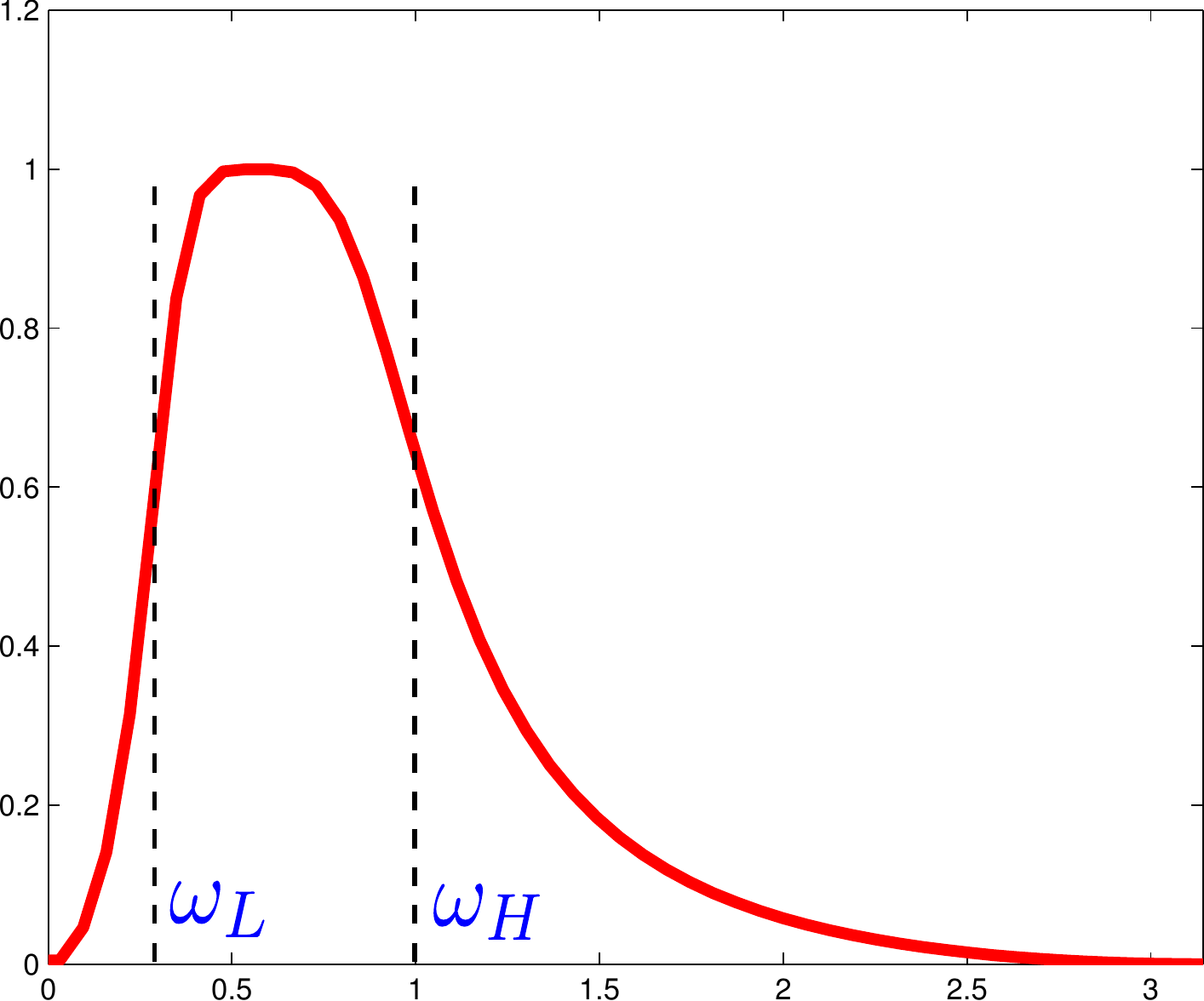} }				
    \subfigure[ 2D: $\widehat g^\gamma(\boldsymbol \omega)$ ]{ \includegraphics[width=0.22\textwidth]{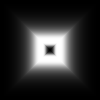} }
    \subfigure[ $\widehat \phi^{\gamma, n}_l(\boldsymbol \omega)$ ]{ \includegraphics[width=0.22\textwidth]{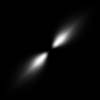} }		
    \subfigure[ $ \text{Re}\{ \phi^{\gamma, n}_l(\boldsymbol x) \} $ ]{ \includegraphics[width=0.22\textwidth]{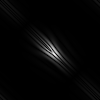} }		    
    \subfigure[ $ \text{Im}\{ \phi^{\gamma, n}_l(\boldsymbol x) \} $ ]{ \includegraphics[width=0.22\textwidth]{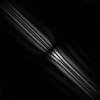} }	
    \subfigure[ $ \abs{ \widehat \phi^{\gamma, n}_l(\boldsymbol \omega) }^2$ ]
              { \includegraphics[width=0.22\textwidth]{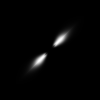} }		
    \subfigure[ $ (\phi^{\gamma, n}_l \ast \phi^{\gamma, n, \vee}_l)(\boldsymbol x)$ ]{\includegraphics[width=0.22\textwidth]{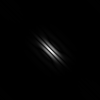} }		            
    \caption{ Image (a) displays the angularpass filter $\widehat h_l^n(\boldsymbol \omega)$ 
              with $\theta = 13\pi/16$, $n = 20$.
              Images (b-c) show the 1D and 2D Butterworth bandpass filters $\widehat g^\gamma(\omega)$ and $\widehat g^\gamma(\boldsymbol \omega)$ with $\omega_L = 0.3, \omega_H = 1, \gamma = 2$,
              (d) the spectrum of the DHBB filter $\widehat \phi^{\gamma, n}_l(\boldsymbol \omega)$. 
              Images (e-f) visualize the real and imaginary part of the DHBB filter $\phi^{\gamma, n}_l(\boldsymbol x)$.
              Images (g-h) display the squared magnitude of the spectrum of the DHBB in the frequency and spatial domains which acts somewhat like a Gabor filter. }
    \label{figFilterDesign}
\end{center}
\end{figure*}

The features of interest in a fingerprint image are repeated (curved) patterns which are concentrated in a particular range of frequencies after a Fourier transformation. 
In principle, the frequencies lower than these range's limits correspond 
to homogeneous regions and those higher to small scale objects, i.e. noise, respectively. 
Taking this prior knowledge into account, we design an algorithm that captures these fingerprint patterns in different directional subbands in the frequency domain for extracting the texture. 

In this section, we design angularpass and bandpass filters. The angularpass filter builds on iterates of the directional Hilbert transformation, 
a multidimensional generalization of the Hilbert transform called the Riesz transform. It can be represented via principal value convolution kernels. 
The bandpass filter builds on the Butterworth transform which can be represented directly via a convolution kernel. We follow here a standard technique designing a bandpass filter 
from a lowpass filter which has an equivalent representation in analog circuit design.

\subsubsection{The $n^\text{th}$ Order Directional Hilbert Transform of a Butterworth Bandpass}

Although a fingerprint image
$$ \boldsymbol k = [k_1,k_2]\mapsto f[\boldsymbol k],\quad \{-M,\ldots,M]\times \{-N,\ldots,N\} \to \{0,\ldots,255\}$$
is a discrete signal observed over a discrete grid, $M,N \in \mathbb N$ we start our considerations with a signal  
$$ \boldsymbol x = (x_1,x_2)\mapsto f(\boldsymbol x),\quad D:=[-a,a]\times [-b,b] \to [0,1]$$
assuming values in a continuum $a,b>0$. 
The frequency coordinates in the spectral domain will be denoted by $\boldsymbol \omega = (\omega_1, \omega_2) \in \mathbb R^2$. %[-\pi, \pi]^2$.

As usual, the following operators are defined first for functions $f$ in the Schwartz Space 
of rapidly-decaying and infinitely differentiable test functions: 
$$
S(\mathbb R^d) = \left\{ f \in C^\infty(\mathbb R^d) \mid 
\displaystyle \sup_{\boldsymbol x \in \mathbb R^d} ~ \abs{(1 + \abs{\boldsymbol x}^m) \frac{d^n}{d \boldsymbol x^n} f(\boldsymbol x)}
< +\infty ~,~~ \forall m, n \in \mathbb Z_+  \right\},
$$  
and continuously extended onto 
$$
L_2(\mathbb R^d) = \left\{ f \in S(\mathbb R^d) \mid \| f \|_{L_2} = \int_{\mathbb R^d} \abs{f(\boldsymbol x)}^2 d \boldsymbol x < +\infty \right\} .
$$

In our context we only need $d=1,2$. Further, we denote the Fourier and its inverse transformations by 
\begin{eqnarray*}
 \cF[f](\boldsymbol \omega) ~=~ \int_{\mathbb R^d} f(\boldsymbol x)\,e^{-j\langle \boldsymbol \omega, \boldsymbol x \rangle}\,d \boldsymbol x
 ~=~ \widehat f(\boldsymbol \omega) ~,~~~
 \cF^{-1}[f](\boldsymbol \omega) 
 ~=~ \frac{1}{(2\pi)^d}\int_{\mathbb R^d} \hat f(\boldsymbol \omega)\,e^{j\langle \boldsymbol\omega, \boldsymbol x\rangle}\,d \boldsymbol x
\end{eqnarray*}
where $j$ denotes the imaginary unit with $j^2=-1$.     

% -------------------------------------------------------------------------------------------
\paragraph{Butterworth bandpass} 

For $\gamma \in\mathbb N$ and frequency bounds $0<\omega_L<\omega_H$, setting $\Delta = \omega_H-\omega_L$, $p^2=\omega_H\omega_L$, 
the one-dimensional ($d=1$) Butterworth bandpass transform is defined via
\begin{eqnarray*}
 \cB[f](x) &=& \cF^{-1}\Bigg[\omega \mapsto 
 \underbrace{ \sqrt{\frac{(\omega\Delta)^{2\gamma} }{(\omega\Delta)^{2\gamma} + (\omega^2-p^2)^{2\gamma}}} }
 _{ := \hat b(\omega) } \,\hat f(\omega) \Bigg](x)\,,
\end{eqnarray*}
cf. \cite{DigitalSignalProcessing2007}.
It is easy to verify that $\hat b(\omega)$ tends to zero for $\omega\to 0$ and $\omega \to \infty$ and has unique maximum at 
the geometric mean $p$ with value $1$. In consequence, for high values of $\gamma$, this filter approximates 
the ideal filter
$$ \hat b_\text{ideal} (\omega) =\left\{\begin{array}{ll}
   1 &\mbox{ if } \omega_L\leq \omega\leq \omega_H\\
   0 &\mbox{ else }
\end{array}\right..$$
The ideal filter, however, suffers from the Gibbs effect. Letting $t = \frac{(j\omega)^2 + p^2}{j\omega\Delta}$, 
we factorize the bandpass Butterworth as
$$
\hat b^2(\omega) = \frac{1}{ (-1)^\gamma (t^2)^\gamma + 1} =
\frac{1}{(-1)^\gamma \prod_{k=1}^\gamma (t^2 - t^2_k)} =
\underbrace{ \prod_{k=1}^\gamma \frac{1}{t-t_k} }_{ H(t) }
\cdot 
\underbrace{ \prod_{k=1}^\gamma \frac{1}{-t-t_k} }_{ H(-t) },
$$
with 
$t_k = e^{\pi j (\gamma + 2k-1)/2\gamma}$ ($k=1,\ldots,\gamma$), the negative squares of which representing the $\gamma$ different complex roots
of $(-1)$.
Then, with the below complex valued factor of $0 \leq \hat b^2 (\omega) = B(j\omega) \cdot B(-j\omega)$ called the transfer function,
$$
H(t) ~=~ H\left( \frac{(j\omega)^2 + p^2}{j\omega\Delta} \right)
~=~ \prod_{k=1}^\gamma \frac{\Delta (j\omega)}{(j\omega)^2 - \Delta t_k (j\omega) + p^2}
~:=~ B(j\omega)\,,
$$
we use the approximation:
$
j\omega ~=~ \log e^{j\omega} ~\approx~ 
2 \frac{e^{j\omega} - 1}{e^{j\omega} + 1}
$
to obtain
$$
B(j\omega) ~\approx~
\prod_{k=1}^\gamma  
\frac{ 2\Delta(e^{2j\omega}-1) }{ \big(4 + p^2-2\Delta t_k\big)e^{2j\omega} + \big( 2p^2 - 8 \big) e^{j\omega} + \big( 4 + p^2 + 2\Delta t_k \big) }
:= B^\gamma (e^{j\omega}).
$$
% -------------------------------------------------------------------------------------------	       
This approximation is often called the bilinear transform, which turns out to reduce the frequency bandwidth of interest, cf. Figure \ref{figCompareOriginalApproximation1}.
	       
The 1D filter $B^\gamma (e^{i\omega})$ is then generalized to a 2D domain. The McClellan transform \cite{McClellan1973}, 
\cite{MersereauMecklenbraukerQuatieri1976}, \cite{MecklenbraukerMersereau1976}, \cite{Tseng2001}, would be one favorable method. Also, recently, bandpass filtering with a radial filter in the Fourier domain has been proposed by
\cite{KocevarKotnikChowdhuryKacic2014}, \cite{GhafoorTajAhmadJafri2014} and \cite{YangXiongVasilakos2013} et al. 
for enhancing fingerprint images.
However, for a simpler reconstruction of 2D filter and a data-friendly alternative to the polar tiling of the frequency plane, 
a Cartesian array is used instead (see \cite{MaPlonka2010}, \cite{CandesDemanetDonohoYing2006}, \cite{YiLabateEasleyKrim2009}, \cite{DoVetterli2005}). 
      
\begin{figure*}[ht] 
\begin{center}
 \subfigure[]{ \includegraphics[width=0.15\textwidth]{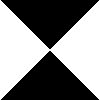} }  \qquad \qquad \qquad
 \subfigure[]{ \includegraphics[width=0.15\textwidth]{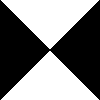} }
 \caption{ The indicator functions in the horizontal direction (a) and vertical direction (b).
         }  
 \label{fig:Indicator}
\end{center} 
\end{figure*}

Thus, on a rectangular domain $D=[-a,a]\times [-b,b]$ with common cuttoff frequencies $0<\omega_L<\omega_H$ and the two characteristic functions
$$ \chi_h(\omega_1,\omega_2) := \left\{\begin{array}{ll}
   1&\mbox{ if } b|\omega_1| \geq a |\omega_2|\\ 0&\mbox{ else}
   \end{array}\right.\,,\quad 
   \chi_v(\omega_1,\omega_2) := \left\{\begin{array}{ll}
   1&\mbox{ if } b|\omega_1| \leq a |\omega_2|\\ 0&\mbox{ else}
   \end{array}\right.\,$$
(see Figure \ref{fig:Indicator}), define
\begin{eqnarray}\label{eq:2D-ButterworthFilter}
\widehat g^\gamma(\omega_1,\omega_2 ) = B^\gamma(e^{j\omega_1})  \chi_h(\omega_1,\omega_2) +  B^\gamma(e^{j\omega_2})  \chi_v(\omega_1,\omega_2)
\end{eqnarray}	      
as the spectrum of our two-dimensional Butterworth filter $g^\gamma(\boldsymbol x)$.
Note that since $\widehat g^\gamma(\boldsymbol \omega) \in L_2(\mathbb R^2)$, there is a well defined $g^\gamma(\boldsymbol x) = \cF^{-1}\left[ \widehat g^\gamma(\boldsymbol \omega) \right](\boldsymbol x). $

Figure \ref{figFilterDesign} (b) and \ref{figFilterDesign} (c) show an example of the 1D and 2D Butterworth bandpass filters.

% ---------------------------------------------------------------------------------------------------------------------

\paragraph{$n$-th order directional Hilbert transformations} For more detail on the Hilbert transform $\mathcal H$ and the Riesz transform $\mathcal R$, 
we refer the reader to the literature for an in-depth discussion \cite{UnserVandeville2010}, 
\cite{UnserChenouardVandeville2011}, \cite{ChaudhuryUnser2009}, \cite{FelsbergSommer2001}, 
\cite{Hahn1996}, \cite{LarkinBoneOldfield2001,Larkin2001, LarkinFletcher2007}, 
\cite{UnserSageVandeville2009}, and \cite{HaeuserHeiseSteidl2014}.

Consider a vector $\boldsymbol u\in \mathbb R^d$ and set and compute, respectively, for $\boldsymbol x\in \mathbb R^d$
\begin{eqnarray}\label{Riesz-transform:def1} 
 \cR[f](\boldsymbol x) 
  &:=& \cF^{-1}\Bigg[\boldsymbol \omega \mapsto \underbrace{ -j\,\frac{\boldsymbol \omega}{\|\boldsymbol \omega\|} }_{ :=~ \widehat h(\boldsymbol \omega) }
      \,\hat f(\boldsymbol \omega)\Bigg](\boldsymbol x)\\ \nonumber
 \cH_{\boldsymbol u}[f](\boldsymbol x) 
  &:=& \langle \boldsymbol u,\cR[f](\boldsymbol x)\rangle ~=~
 \cF^{-1}\Bigg[\boldsymbol \omega \mapsto \underbrace{ -j\,\frac{\langle \boldsymbol u, \boldsymbol \omega\rangle}{\|\boldsymbol \omega\|} }_{:=~ \widehat h_{\boldsymbol u} (\boldsymbol \omega)}
 \,\hat f(\boldsymbol \omega)\Bigg](\boldsymbol x)
 \\ \label{Riesz-transform:def}
 \underbrace{\cH_{\boldsymbol u}\ldots \cH_{\boldsymbol u}}_{n-\text{times}} [f](\boldsymbol x) ~=~ 
 \cH_{\boldsymbol u}^n[f](\boldsymbol x)&=& 
 \cF^{-1}\Bigg[\boldsymbol \omega \mapsto 
 \underbrace{ (-j)^n\,\frac{\langle \boldsymbol u, \boldsymbol \omega\rangle^n}{\|\boldsymbol \omega\|^n} }_{ \widehat h_{\boldsymbol u}^n (\boldsymbol \omega) }
 \,\hat f(\boldsymbol \omega)\Bigg](\boldsymbol x), 
 \qquad n \in \mathbb N
\end{eqnarray}
The first line (\ref{Riesz-transform:def1}) called the Riesz transform has a representation as a principal value integral
\begin{eqnarray*}
 \cR[f](\boldsymbol x) &=& (f \pvc h)(\boldsymbol x) ~=~\lim_{\epsilon \to 0} \int_{\|\boldsymbol y - \boldsymbol x\|\geq \epsilon} 
 f(\boldsymbol x - \boldsymbol y) \,h(\boldsymbol y)\,d \boldsymbol y
\end{eqnarray*}
where 
$$h(\boldsymbol y) := \left\{\begin{array}{rcl} \frac{\boldsymbol y}{\|\boldsymbol y\|^{d+1}}\frac{\Gamma\left(\frac{d+1}{2}\right)}{\pi^{\frac{d+1}{2}}}&\mbox{ for } d>1\\\frac{1}{\pi y}&\mbox{ for }d=1\end{array}\right.$$
Setting $h_{\boldsymbol u}(\boldsymbol y) = \langle \boldsymbol u, h(\boldsymbol y)\rangle$ and 
$h^n_{\boldsymbol u}(\boldsymbol y) =  \big( h_{\boldsymbol u}\pvc\ldots\pvc h_{\boldsymbol u} \big) (\boldsymbol y) $ we have for the third line  (\ref{Riesz-transform:def}) called the $n$-order directional Hilbert transform that
$$ \cH^n_{\boldsymbol u}[f](\boldsymbol x) ~=~f\pvc h^n_{\boldsymbol u} (\boldsymbol x)\,.$$
Since $-1\leq \frac{\langle \boldsymbol u, \boldsymbol \omega\rangle}{\|\boldsymbol \omega\|}\leq 1$ and high powers preserve the values near $\pm 1$ while forcing all other values in $(-1,1)$ towards  $0$, 
this filter gives roughly the same result as an inverse Fourier transform of a convolution of the signal's Fourier transform with
$$ A_{\alpha, \boldsymbol u}(\boldsymbol \omega) =\left\{\begin{array}{ll}
   1 &\mbox{ if } \frac{|\langle \boldsymbol u, \boldsymbol \omega\rangle|}{\| \boldsymbol \omega\|} \geq \cos \alpha\\
   0 &\mbox{ else }
\end{array}\right.$$
for small $\alpha >0$.
The directional Hilbert transform, however, suffers less from a Gibbs effect than this sharp cutoff filter. 
	     
In 2D, the direction vector is $\boldsymbol u = [\cos(\theta) , \sin(\theta)]^T$ with the discretized 
$\theta = \frac{\pi l}{L} \in [0, \pi)$ and $l = 0, 1, \ldots, L-1$, where $L \in \mathbb N$ is the total number of orientation. 
Rewrite the impulse response of the $n\mbox{-th} \in \mathbb N$ order directional Hilbert transform 
in (\ref{Riesz-transform:def}) as
\begin{equation} \label{eq:ImpulseResponse-DirectionalHilbert}
\widehat h_{\boldsymbol u}^n (\omega_1, \omega_2) 
~=~ \Bigg[ -j \cos\bigg( \tan_2^{-1}\Big( \frac{\omega_2}{\omega_1} \Big) - \frac{\pi l}{L} \bigg) \Bigg]^n
~:=~ \widehat h_{l}^n (\omega_1, \omega_2).
\end{equation}
% 
% ---------------------------------------------------------------------------------------------------------------------

Putting together (Eq. (\ref{eq:2D-ButterworthFilter}), (\ref{Riesz-transform:def}), (\ref{eq:ImpulseResponse-DirectionalHilbert})), 
for a fixed bandpass $\omega_L<\omega_H$ and $L$ directional subbands we have thus the DHBB filter of order $\gamma,n$:
\begin{equation} \label{eq:DHBB}
 \cH_u^n [g^\gamma] (\boldsymbol x) ~=~
 \cF^{-1} \Bigg[ \boldsymbol \omega \mapsto \underbrace{ \widehat h_l^n(\boldsymbol \omega) \cdot \widehat g^\gamma (\boldsymbol \omega) }
 _{ := \widehat \phi_l^{\gamma, n} (\boldsymbol \omega) } \Bigg](\boldsymbol x) 
 ~:=~ \phi_l^{\gamma, n} (\boldsymbol x) .
\end{equation}

% -----------------------------------------------------------------------------------------------------------
\subsubsection{Thresholding}

\begin{figure*}
\begin{center}
\begin{tabular}[c]{cc}
  \begin{tabular}[c]{c}
      \subfigure{ \includegraphics[width=0.41\textwidth]{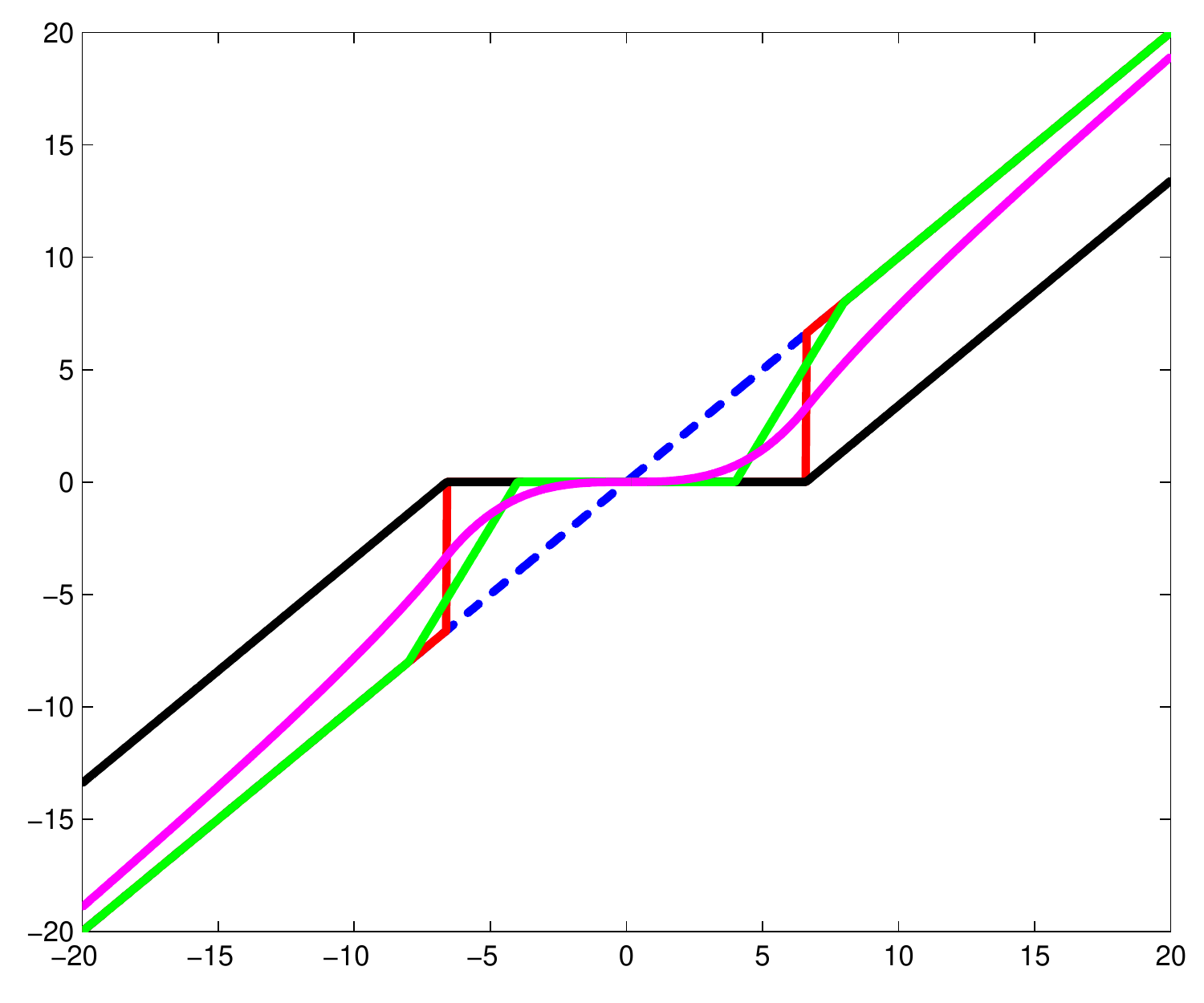} } %% Put figure (b) here
  \end{tabular} &
  \begin{tabular}[c]{c}
        %% Put figure (a)_0 here & Put figure (a)_1 here \\
        \subfigure{\includegraphics[width=0.22\textwidth]{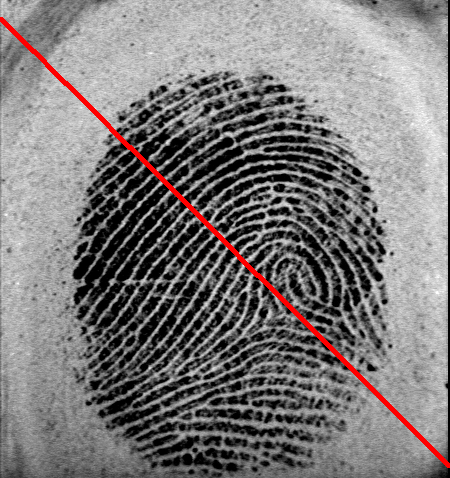}}
        \subfigure{\includegraphics[width=0.22\textwidth]{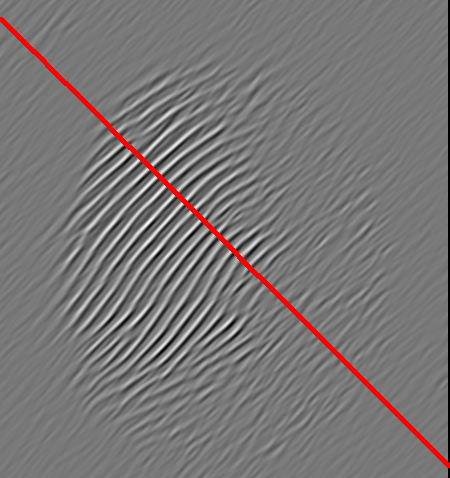}}        
        \\ 
        %% Put figure (a)_0 here & Put figure (a)_1 here \\
        \subfigure{ \includegraphics[width=0.22\textwidth]{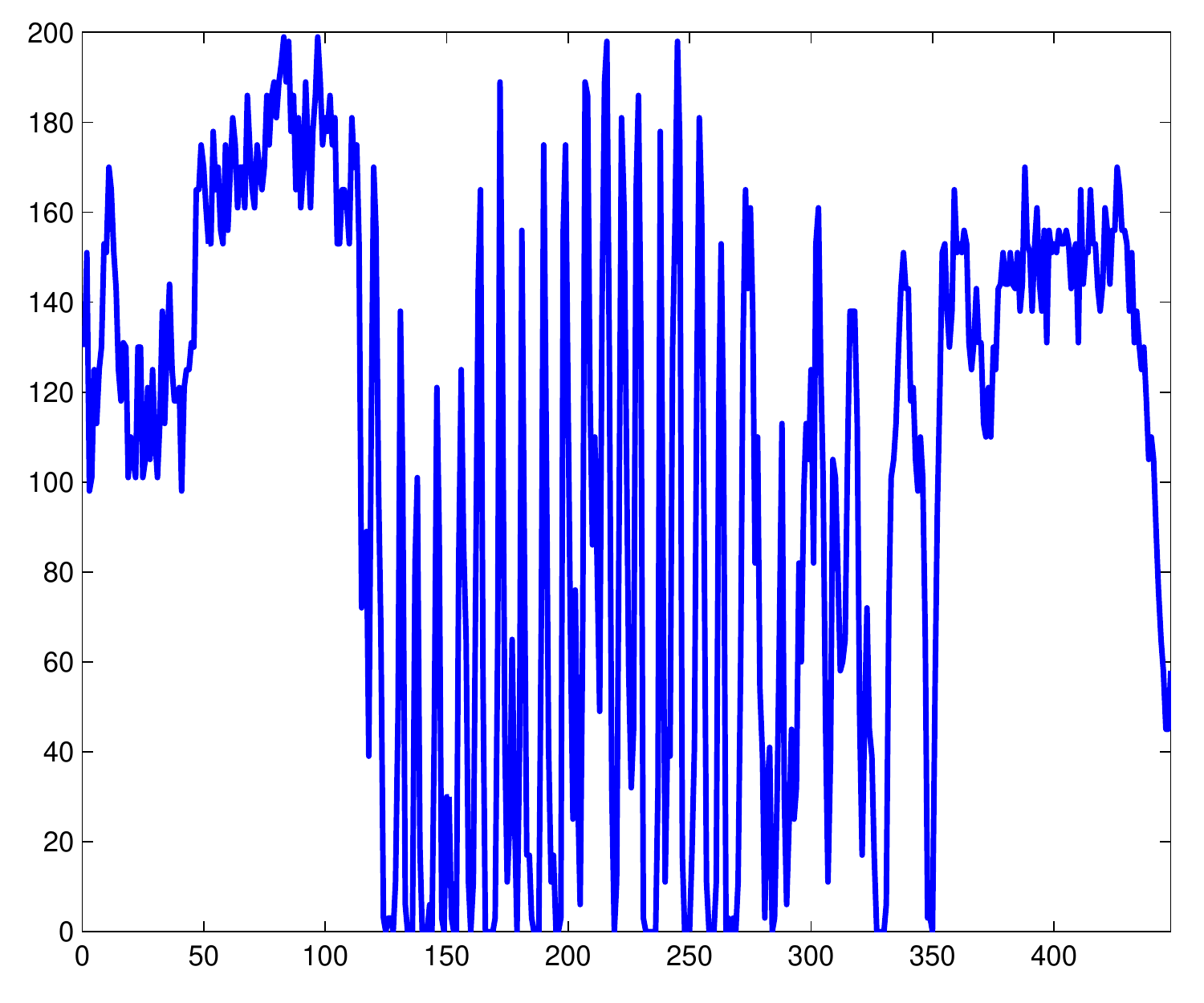} }
        \subfigure{ \includegraphics[width=0.22\textwidth]{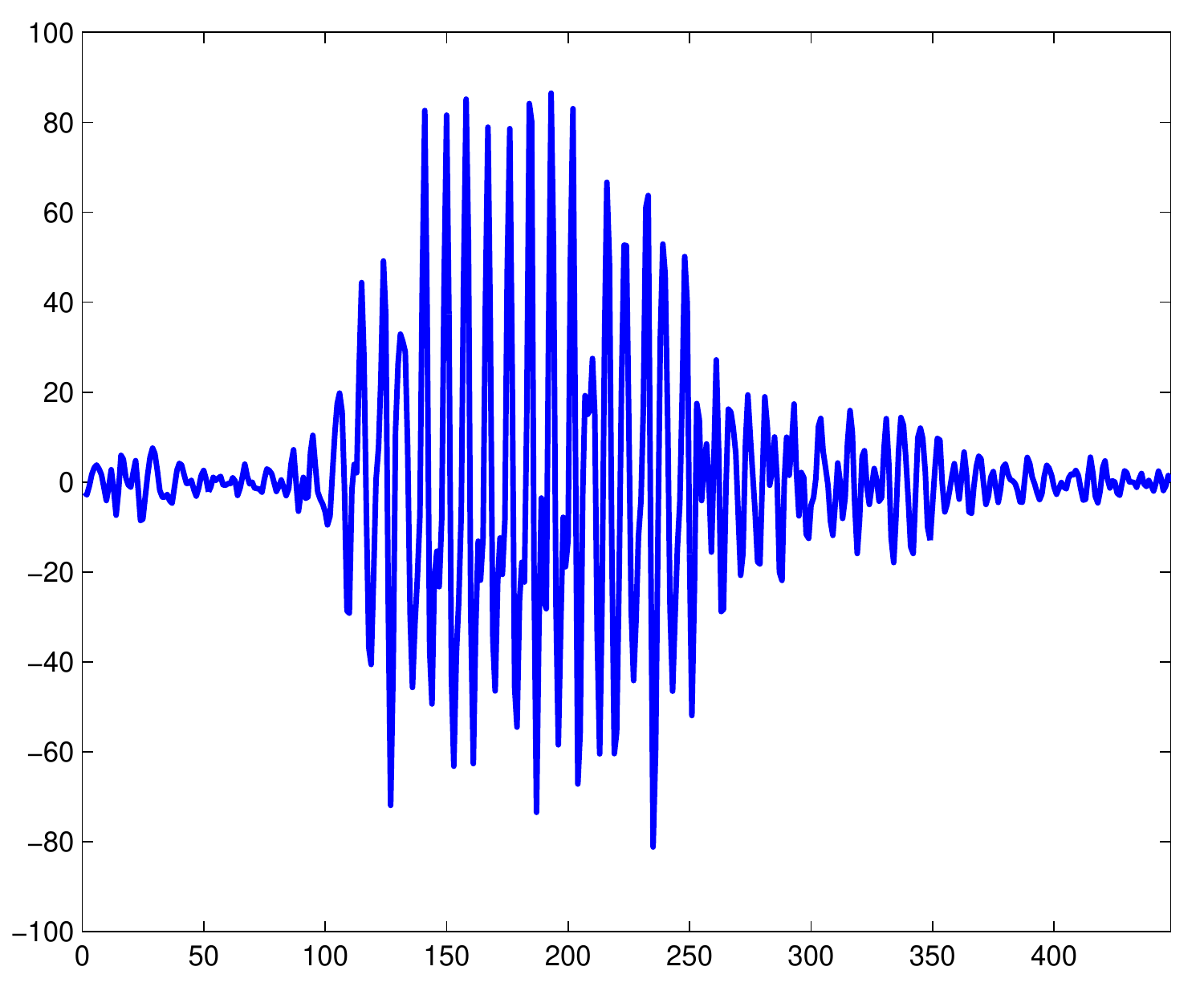} }        
  \end{tabular} \\  
%   (a) & (b) \\
\end{tabular}
          \subfigure{ \includegraphics[width=0.22\textwidth]{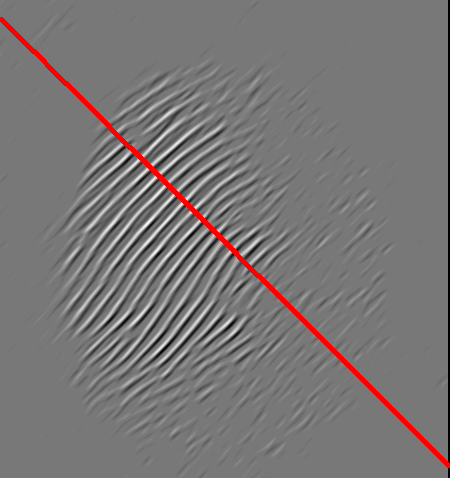} }
          \subfigure{ \includegraphics[width=0.22\textwidth]{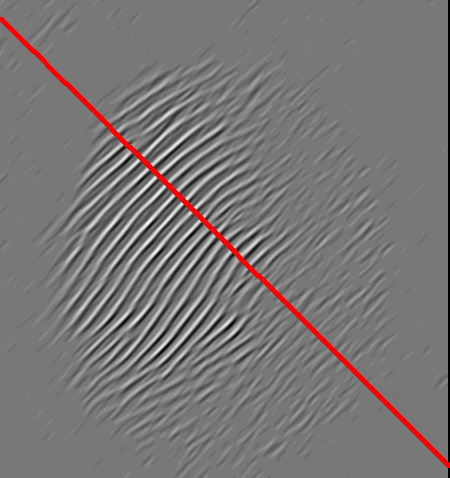} }
          \subfigure{ \includegraphics[width=0.22\textwidth]{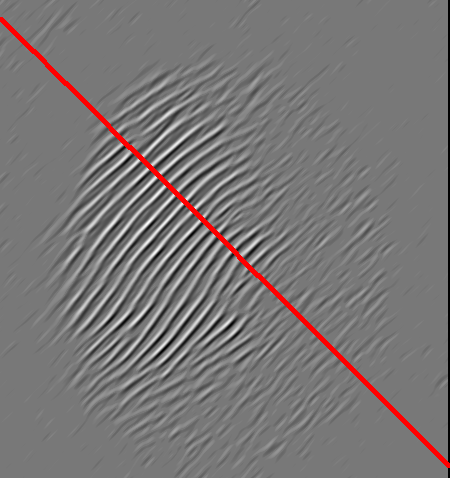} }    
          \subfigure{ \includegraphics[width=0.22\textwidth]{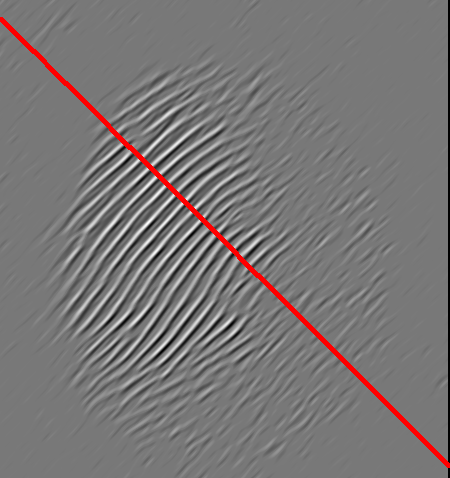} }
          \\
          \subfigure{ \includegraphics[width=0.22\textwidth]{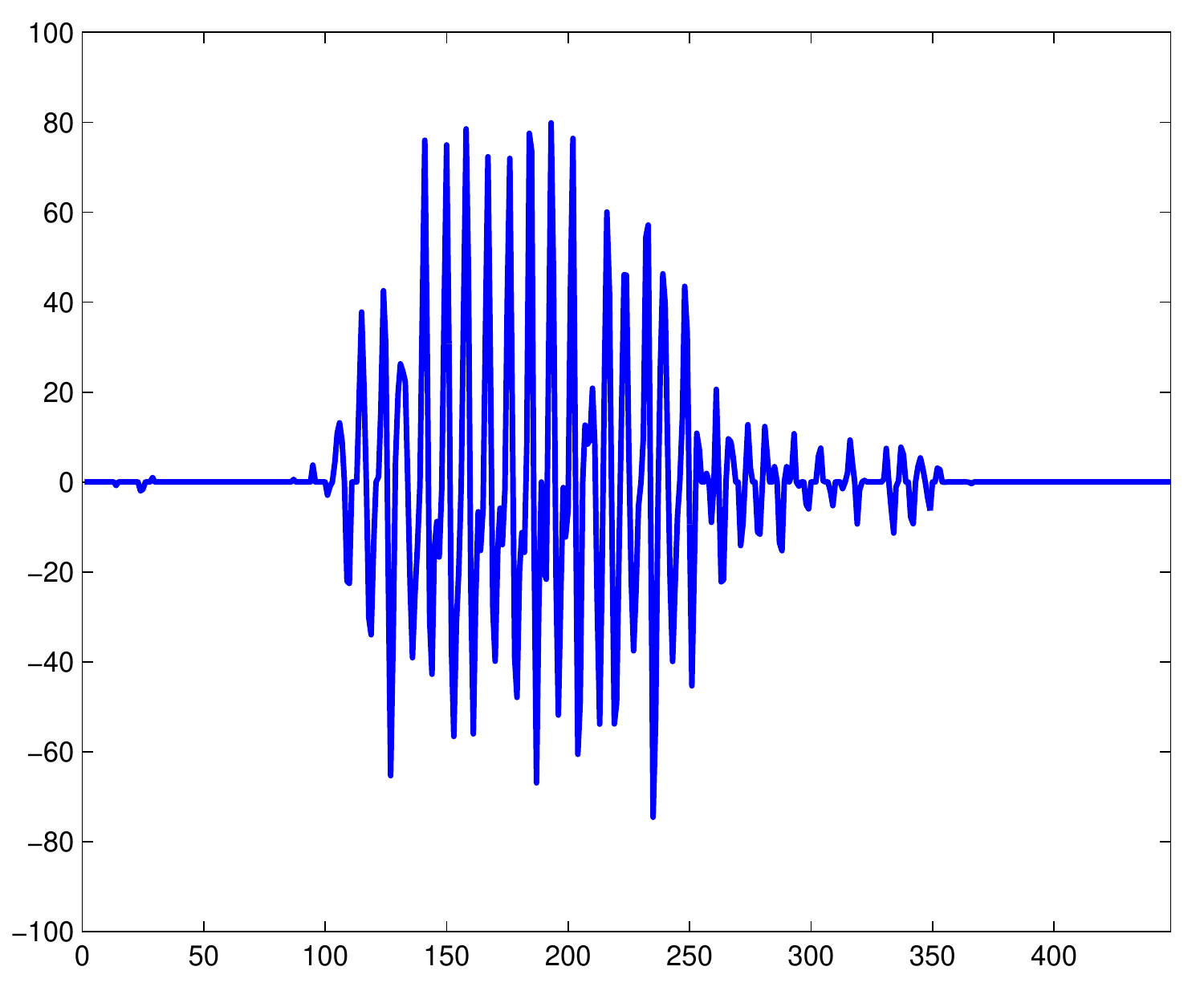} }
          \subfigure{ \includegraphics[width=0.22\textwidth]{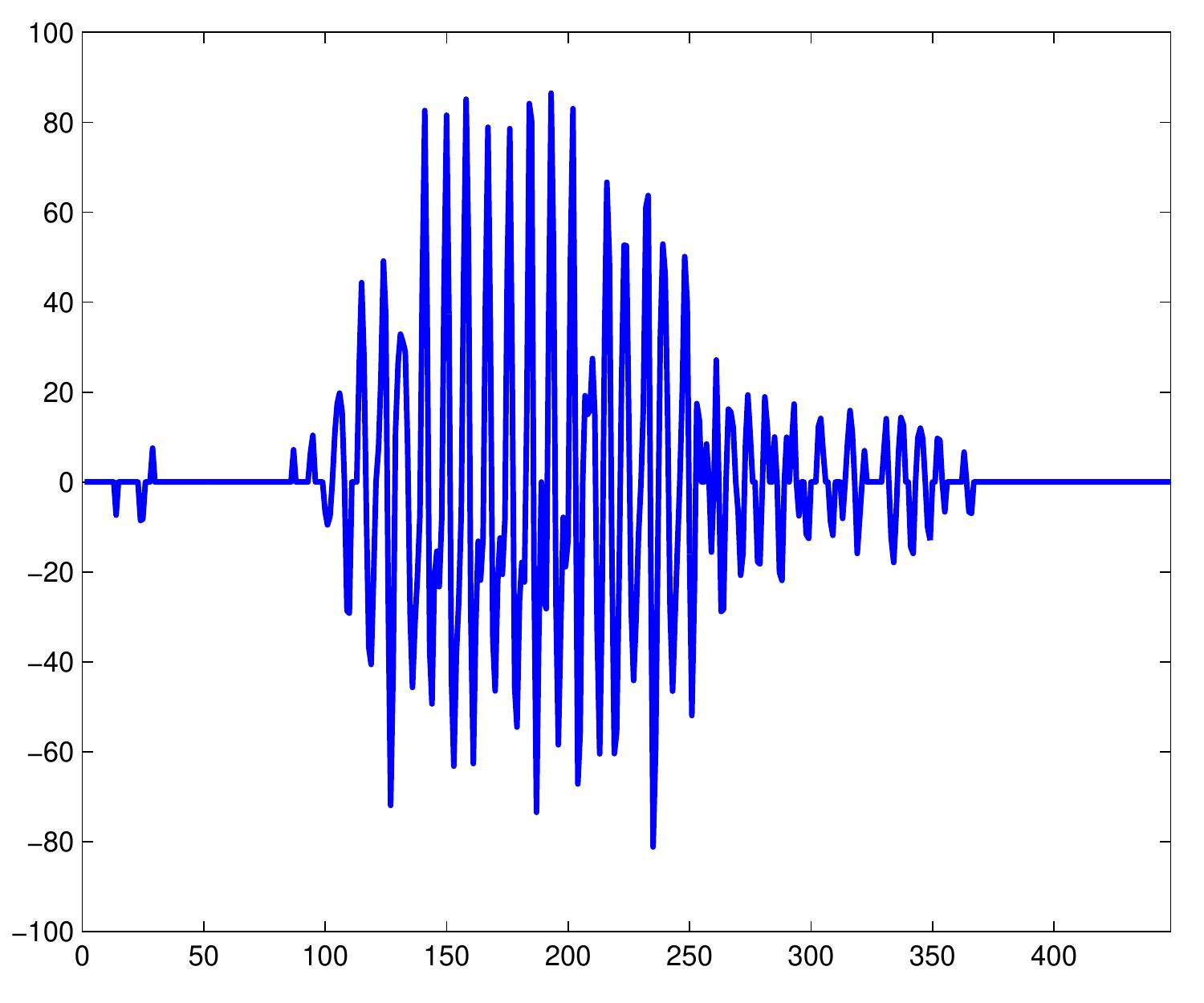} }
          \subfigure{ \includegraphics[width=0.22\textwidth]{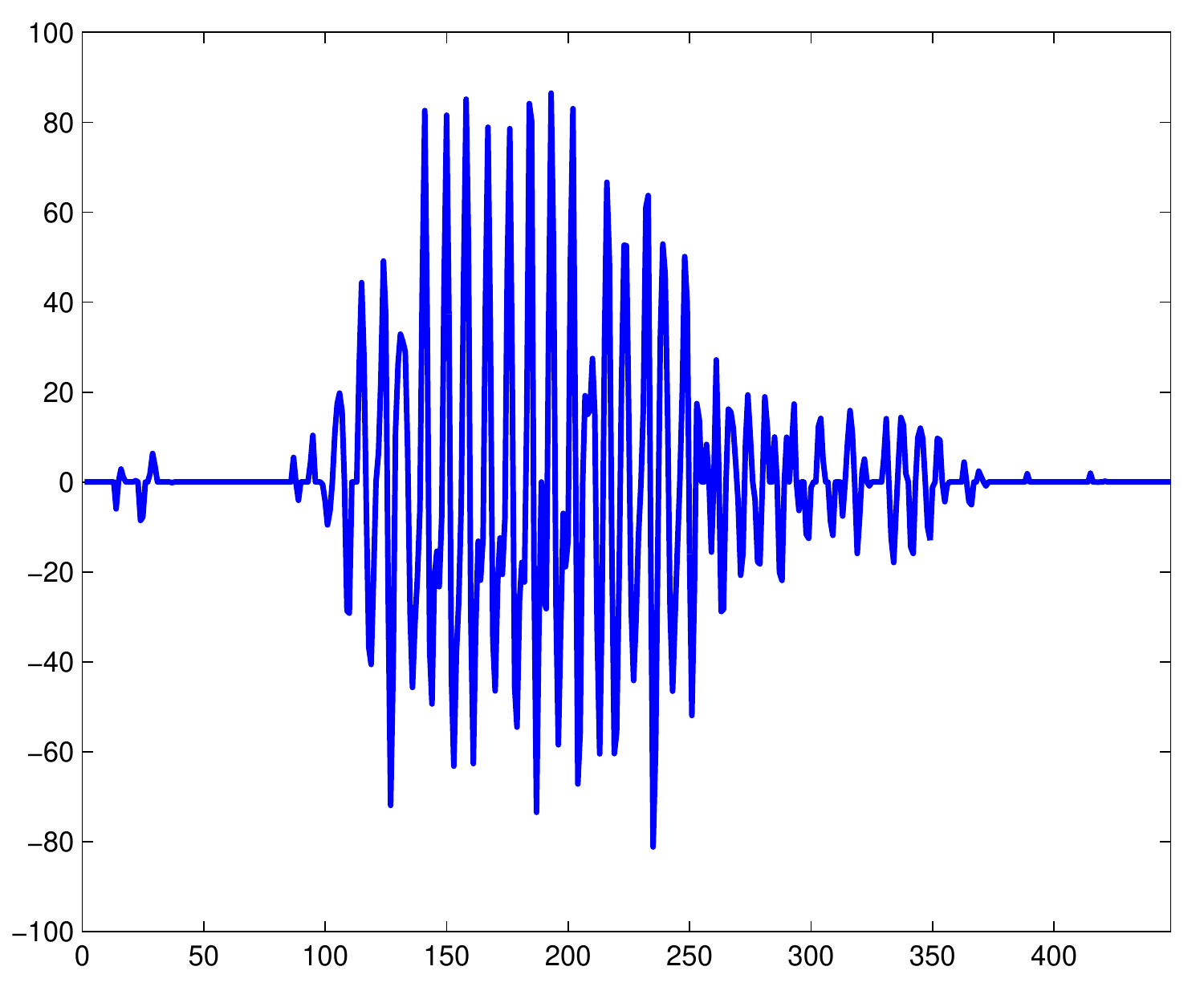} }
          \subfigure{ \includegraphics[width=0.22\textwidth]{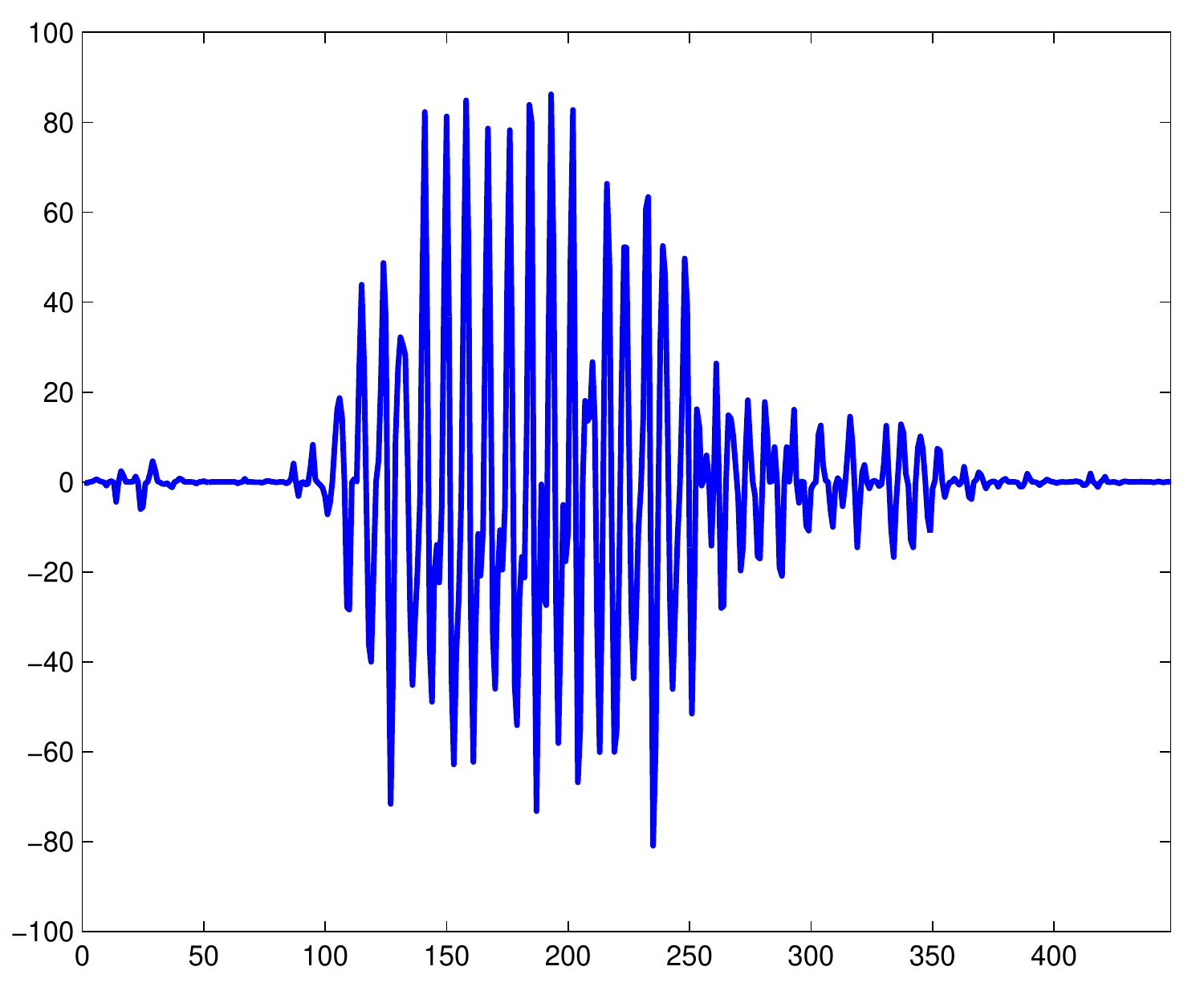} } 
          \caption{
                   Four typical thresholding functions (red: hard, black: soft, green: semi-soft, magenta: nonlinear)
                   are compared (top left).
                   The following six pairs show an image 
                   and the visualization of the corresponding 1D cross section along the red line.
                   F.l.t.r and top to bottom:  
                   the original image $f[\boldsymbol k]$, the coefficient $c_l[\boldsymbol k]$ and
                   the thresholded coefficients $d_l[\boldsymbol k]$ for the soft, hard, semisoft 
                   and nonlinear thresholding operators.
                   Comparing the four cross sections in the bottom row, we observe that 
                   soft-thresholding achieves the sparsest solution.
                    \label{fig:Threshold} 
                  }           
\end{center}
\end{figure*}

         For given $\beta >0$, soft-thresholding is defined as follows
         \begin{equation} \label{eq:SoftThresholding}
          x\mapsto \mathbf T (x, \beta) = \frac{x}{|x|} \cdot \max \left( |x| - \beta, 0 \right ).
         \end{equation}
         Thus, the thresholded coefficients are
         $
         d_l[\boldsymbol m] =  \mathbf T\{ c_l[\boldsymbol m], \beta \}. 
         $
         Note that $d_l[\boldsymbol m]$ is a solution of the $\ell_1$-shrinkage minimization problem
         $$   \min_{u} \left \{ \beta \parallel u \parallel_{\ell_1} + \frac{1}{2} \parallel u - c_l \parallel^2_{\ell_2} \right \}\,$$
         yielding soft-thresholding (cf. \cite{DonohoJohnstone1994}). 
         %(see the proposition \ref{prop_dl} in Appendix \ref{AppendixD}). 
         Figure \ref{fig:Threshold} visualizes the effect of the soft-thresholding and the comparison with the others 
         (such as: hard \cite{DonohoJohnstone1994}, semi-soft \cite{SanamShahnaz2013}
         and nonlinear \cite{NasriNezamabadi2009} thresholding operators).

% -----------------------------------------------------------------------------------------------------

\subsection{Fingerprint Segmentation} \label{secSegmentation}

After having designed the FDB filter, let us now ponder on parameter selection, image binarization and morphological processing.

\subsubsection{Parameter Choice for Texture Extraction}

A fingerprint image will be rescaled such that its oscillation pattern stays in a specific range in the Fourier domain, the 
coordinates of which are $ \omega_i = [-\pi, \pi], i \in \{ 1, 2 \} $. 
For choosing the cutoff frequencies $\omega_L$ and $\omega_H$,
we incorporate our prior knowledge about adult fingerprint images at resolution of 500 DPI:
Valid interridge distances remain in a known range approximately from 3 to 25 pixels \cite{Gottschlich2012}.
This corresponds exactly to $\omega_H = 1$ as a limit for high frequencies. 
A limit of $\omega_L = 0.3$ for low frequencies of the Butterworth bandpass filter
corresponds to an interridge distance of about 12 pixels. 
The range $|\omega_i| \in [\omega_H , \pi]$ contains the small scale objects which are considered as noise. 
The range $|\omega_i| \in [0, \omega_L]$ contains 
the low frequency objects, corresponding to homogeneous regions.

\begin{figure*}
\begin{center}
    \subfigure[$\widehat h^n_l(\boldsymbol \omega), n=3$]{ \includegraphics[width=0.22\textwidth]{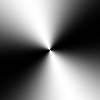} }		
    \subfigure[$\widehat\phi^{\gamma,n}_{l}(\boldsymbol \omega)$]{ \includegraphics[width=0.22\textwidth]{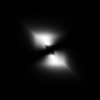} }
    \subfigure[$\abs{\phi^{\gamma,n}_{l}( \boldsymbol x )}$]{ \includegraphics[width=0.22\textwidth]{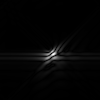} }
    \subfigure[$ (\phi^{\gamma, n}_l \ast \phi^{\gamma, n, \vee}_l)(\boldsymbol x)$]{ \includegraphics[width=0.22\textwidth]{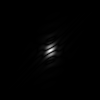} }
    
    \subfigure[n=20]{ \includegraphics[width=0.22\textwidth]{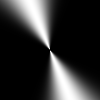} }
    \subfigure[]{ \includegraphics[width=0.22\textwidth]{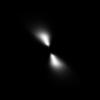} }
    \subfigure[]{ \includegraphics[width=0.22\textwidth]{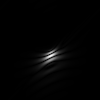} }
    \subfigure[]{ \includegraphics[width=0.22\textwidth]{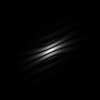} }
    
    \subfigure[n=100]{ \includegraphics[width=0.22\textwidth]{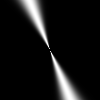} }	
    \subfigure[]{ \includegraphics[width=0.22\textwidth]{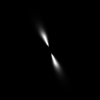} }
    \subfigure[]{ \includegraphics[width=0.22\textwidth]{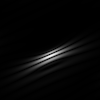} }
    \subfigure[]{ \includegraphics[width=0.22\textwidth]{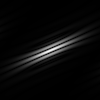} }    
    
    \subfigure[n=3]{ \includegraphics[width=0.2\textwidth]{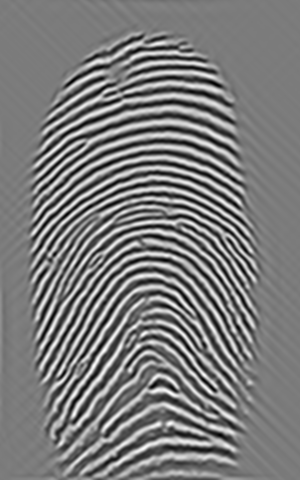} }		
    \subfigure[n=20]{ \includegraphics[width=0.2\textwidth]{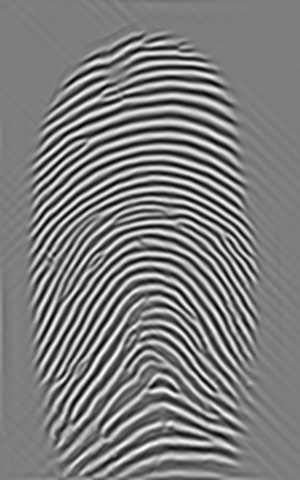} }		
    \subfigure[n=100]{ \includegraphics[width=0.2\textwidth]{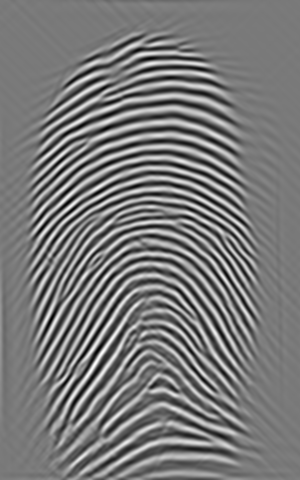} }    

    \caption{Angular bandpass $\widehat h^n_l(\boldsymbol \omega)$ at $\theta = \frac{7\pi}{16}, \gamma = 3$ 
		and different orders $ n \in \{ 3, 20, 100 \} $ and their responses (last row).}    
    \label{figCompareAngularBandpassNumberDirections}
\end{center}
\end{figure*}

The number of directions $L$ in and the order $n$ of the directional Hilbert transform involves a tradeoff between the following effects. We observe that with increased order $n$ the filter's shape becomes thinner in the Fourier domain. Although this sparsity smooths  the texture image in the spatial domain, in order to fully cover all FOTIs, $L$ needs to grow with $n$. However, a disadvantage of choosing large $n$ and $L$
is that errors occur on the boundary due to the 
over-smoothing effect as illustrated in Figure \ref{figCompareAngularBandpassNumberDirections} (o).

\begin{figure*}
\begin{center}
    \subfigure[ $\gamma=1$ ]{ \includegraphics[width=0.20\textwidth]{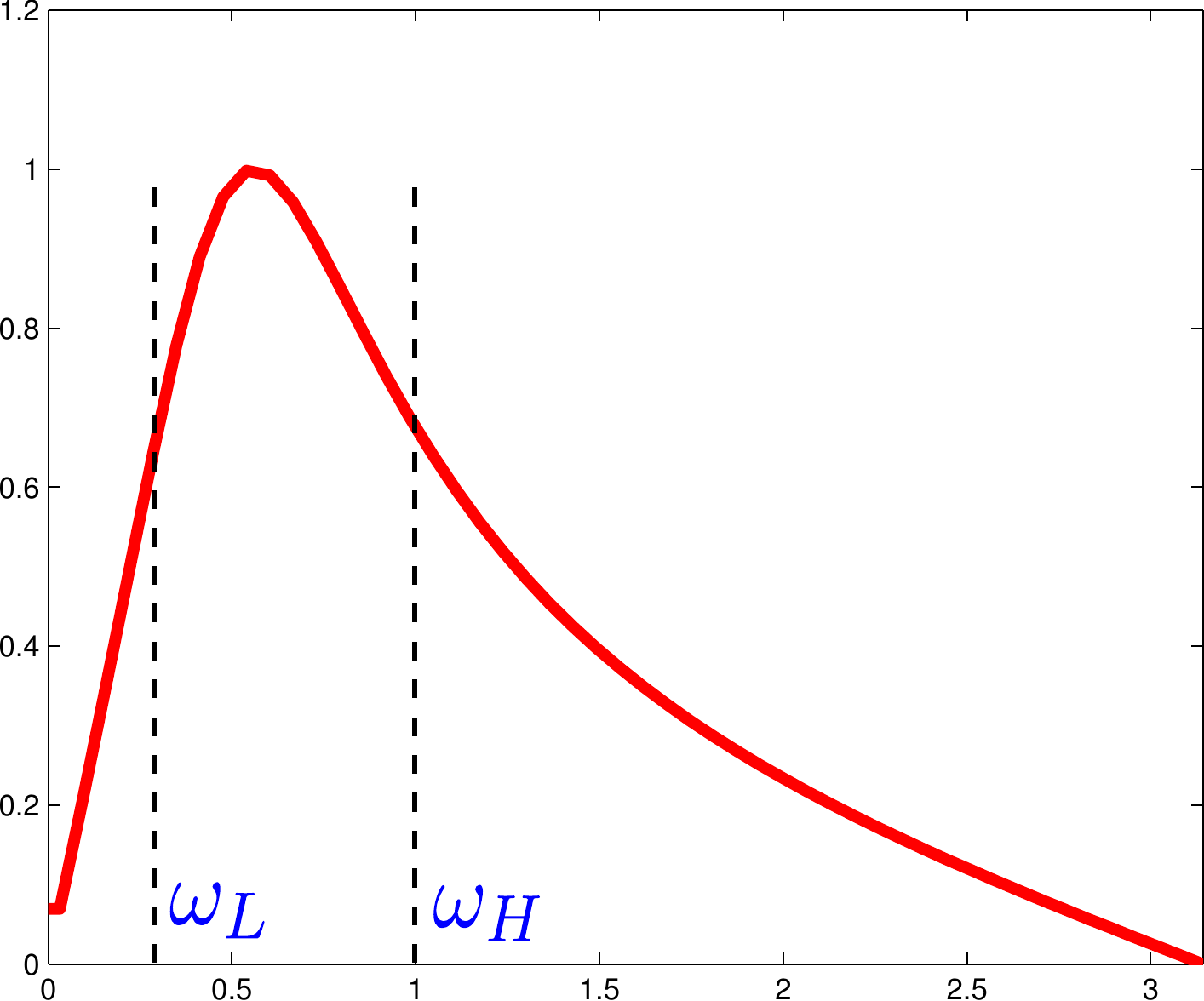} }		
    \subfigure[ $\gamma=2$ ]{ \includegraphics[width=0.20\textwidth]{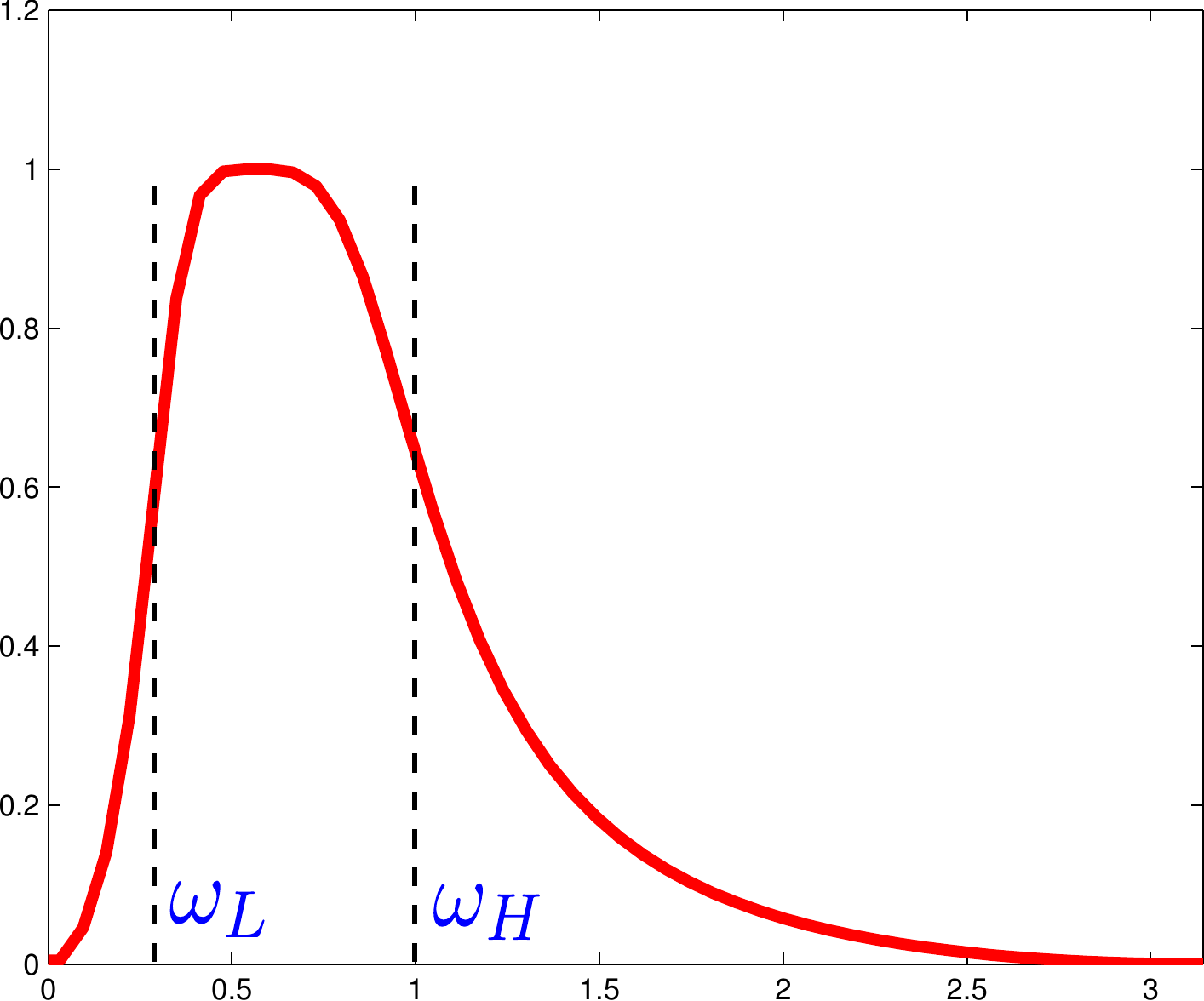} }		
    \subfigure[ $\gamma=3$ ]{ \includegraphics[width=0.20\textwidth]{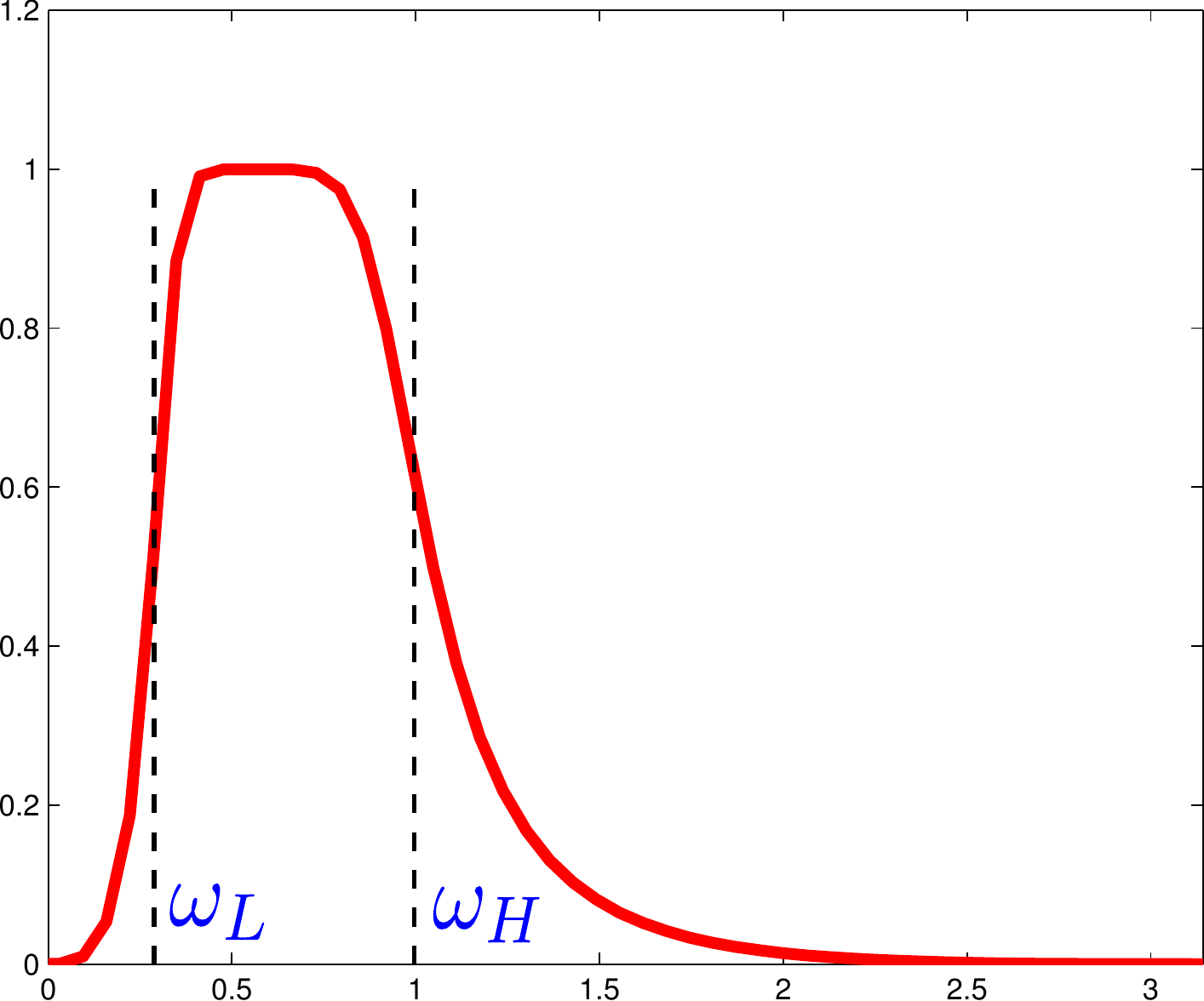} }	
    \subfigure[ $\gamma=10$ ]{ \includegraphics[width=0.20\textwidth]{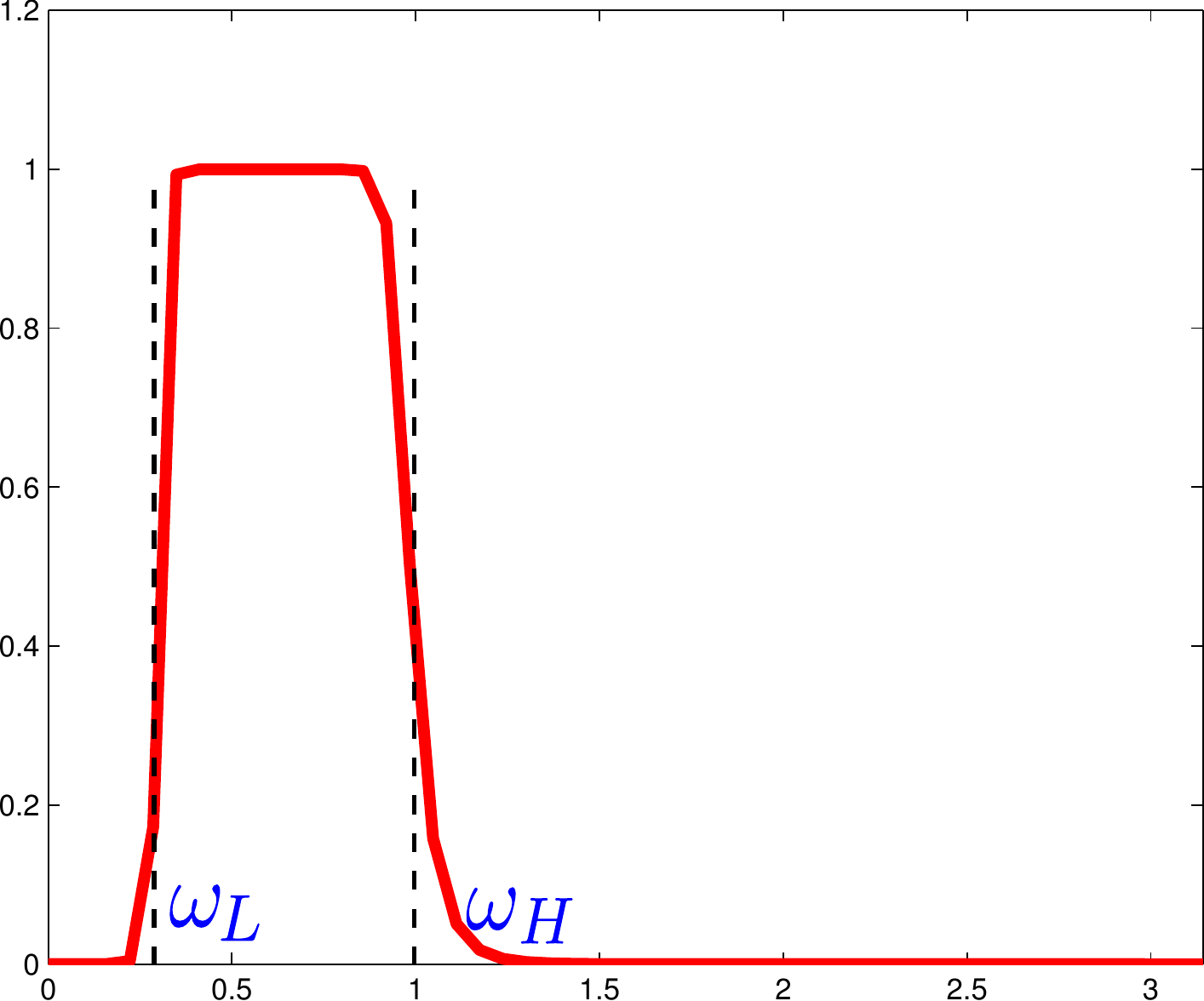} }	
    \\
    \subfigure{ \includegraphics[width=0.15\textwidth]{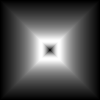} }		
    \subfigure{ \includegraphics[width=0.15\textwidth]{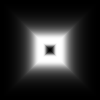} }		
    \subfigure{ \includegraphics[width=0.15\textwidth]{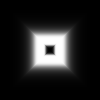} }	
    \subfigure{ \includegraphics[width=0.15\textwidth]{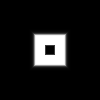} }	
    \\
    \subfigure{ \includegraphics[width=0.15\textwidth]{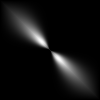} }		
    \subfigure{ \includegraphics[width=0.15\textwidth]{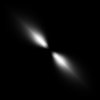} }		
    \subfigure{ \includegraphics[width=0.15\textwidth]{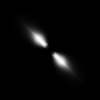} }	
    \subfigure{ \includegraphics[width=0.15\textwidth]{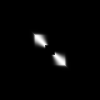} }	
    \\
    \subfigure{ \includegraphics[width=0.15\textwidth]{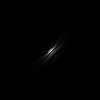} }		
    \subfigure{ \includegraphics[width=0.15\textwidth]{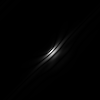} }		
    \subfigure{ \includegraphics[width=0.15\textwidth]{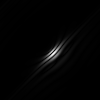} }	
    \subfigure{ \includegraphics[width=0.15\textwidth]{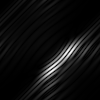} }	
    \\
    \subfigure{ \includegraphics[width=0.15\textwidth]{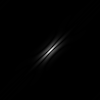} }		
    \subfigure{ \includegraphics[width=0.15\textwidth]{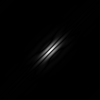} }		
    \subfigure{ \includegraphics[width=0.15\textwidth]{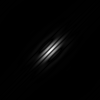} }	
    \subfigure{ \includegraphics[width=0.15\textwidth]{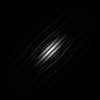} }	 
    \\
    \subfigure{ \includegraphics[width=0.15\textwidth]{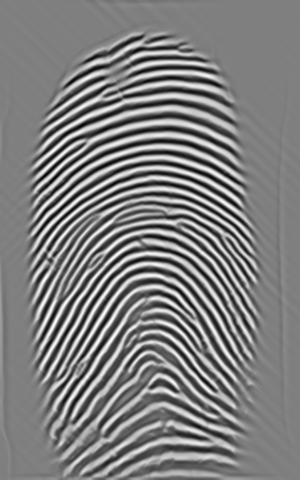} }	
    \subfigure{ \includegraphics[width=0.15\textwidth]{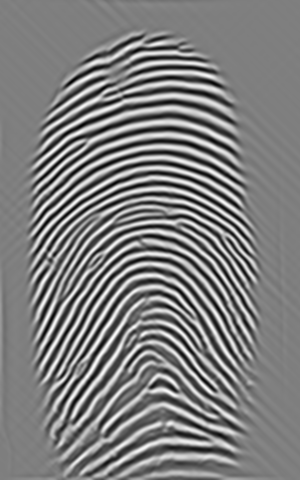} }	
    \subfigure{ \includegraphics[width=0.15\textwidth]{FVC2004_DB3_IM_67_1_n20_gam3_re.png} }	
    \subfigure{ \includegraphics[width=0.15\textwidth]{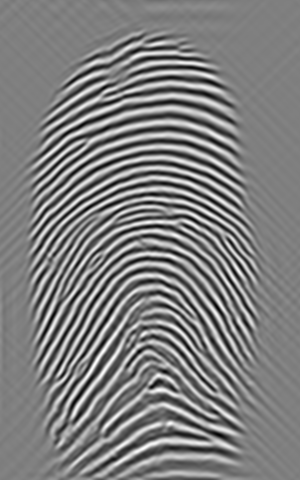} }
    \caption{Butterworth bandpass filter $\widehat g^\gamma(\boldsymbol \omega)$ at $ \omega_L = 0.3, \omega_H = 1$ and different $\gamma$, angular bandpass filter with $n=20, L=16, \theta = \frac{5\pi}{16}$, and their responses. 
             $1^\text{st}$ row: 1D Butterworth, $2^\text{nd}$ row: 2D Butterworth, $3^\text{rd}$ row: $\widehat \phi^{\gamma,n}_{l}(\boldsymbol \omega)$, 
             $4^\text{th}$ row: $\abs{\phi^{\gamma,n}_{l}(\boldsymbol x)}$, 	
             $5^\text{th}$ row: $ (\phi^{\gamma, n}_l \ast \phi^{\gamma, n, \vee}_l)(\boldsymbol x)$, 
             $6^\text{th}$ row: their responses.}		    
    \label{figCompareBandGamma}    
\end{center}
\end{figure*}

\begin{figure*}
\begin{center}
    \subfigure[ $\widehat g^\gamma(\omega)$ ]{\includegraphics[width=0.35\textwidth]{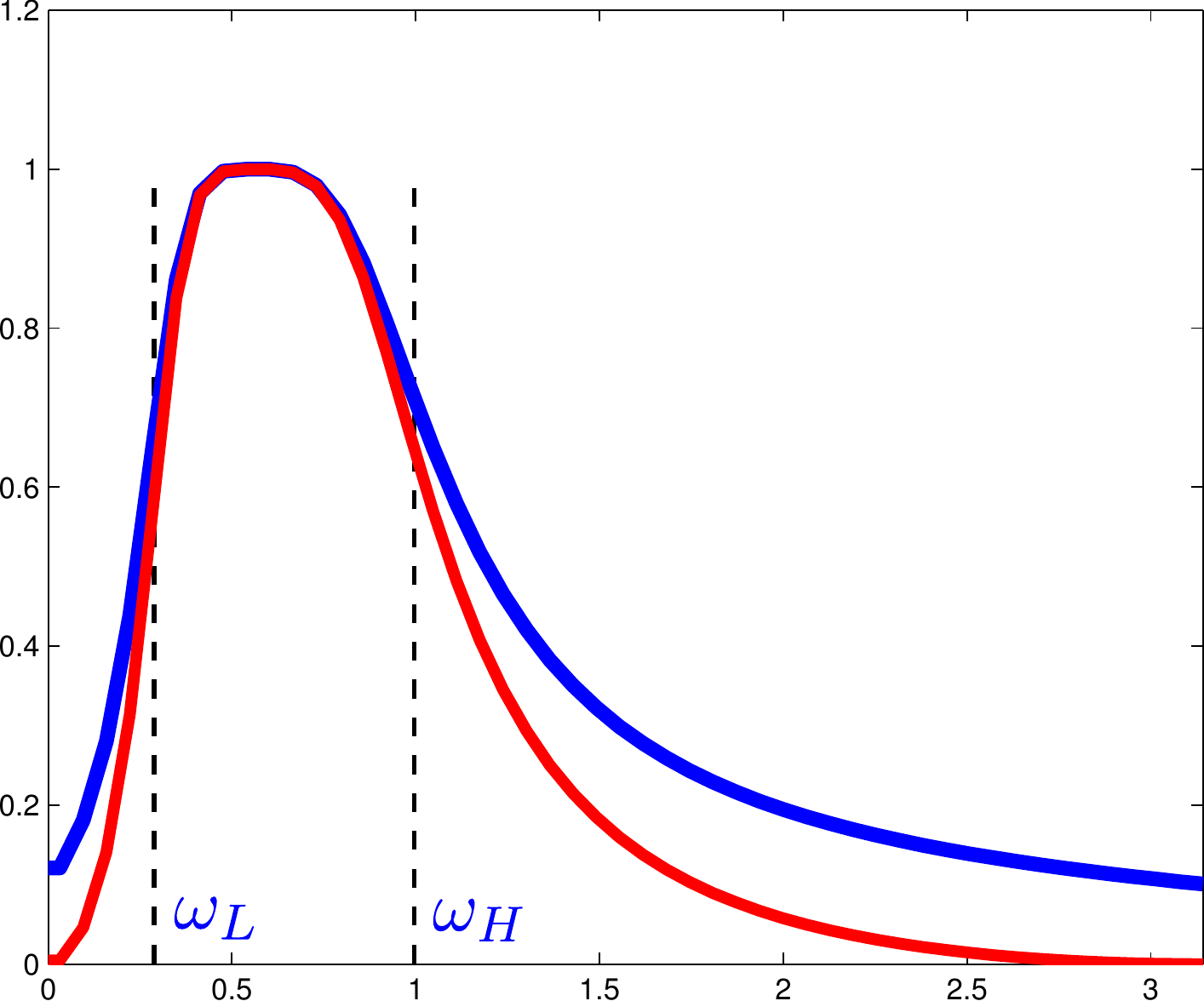} }		
    \\
    \subfigure[ $\widehat g^\gamma(\boldsymbol \omega)$ ]{ \includegraphics[width=0.19\textwidth]{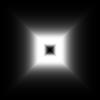} }		
    \subfigure[ $\widehat \phi^{\gamma,n}_l (\boldsymbol \omega)$ ]{ \includegraphics[width=0.19\textwidth]{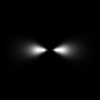} }
    \subfigure[ $\abs{\phi^{\gamma,n}_{l}(\boldsymbol x)}$ ]{ \includegraphics[width=0.19\textwidth]{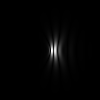} }
    \\
    \subfigure[ $\widehat g^\gamma(\boldsymbol \omega)$ ]{ \includegraphics[width=0.19\textwidth]{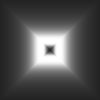} }		
    \subfigure[ $\widehat \phi^{\gamma,n}_l (\boldsymbol \omega)$ ]{ \includegraphics[width=0.19\textwidth]{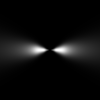} }
    \subfigure[ $\abs{\phi^{\gamma,n}_{l}(\boldsymbol x)}$ ]{ \includegraphics[width=0.19\textwidth]{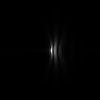} }
    \caption{Image (a) displays a 1D Butterworth bandpass filter (blue) and its approximation (red).
             Image (b) shows the 2D Butterworth bandpass filter $\widehat g^\gamma(\boldsymbol \omega)$ at $n=20, \theta = 0, \gamma = 2$, 
						 and the corresponding DHBB filter in the Fourier and spatial domains (c, d)
             for the approximation by the bilinear transform.
             Image (e) visualizes the 2D version of the original filter 
             and the corresponding DHBB filter (f, g).}		    
    \label{figCompareOriginalApproximation1}
\end{center}
\end{figure*}

The next parameter to select is the order of the Butterworth filter $\gamma$. 
An illustration of the filter for different orders $\gamma \in \{ 1, 2, 3, 10 \}$ 
and with cutoff frequencies $\omega_L=0.3$ and $\omega_H=1$
is shown in Figure \ref{figCompareBandGamma}, its bilinear approximation in Figure \ref{figCompareOriginalApproximation1}.
As $\gamma$ increases the filter becomes sharper.
For very large values of $\gamma$, 
it approaches the ideal filter which is known to cause the unfavorable Gibbs effect.

The thresholding value $\beta$ 
separates large coefficients corresponding to the fingerprint pattern (FOTIs)  
(which are slightly attenuated due to soft-thresholding)
from small coefficients corresponding to non-FOTIS and FOTIs which are not features due to the fingerprint pattern
(these are eliminated).
On the one hand, if $\beta$ is chosen too large,  more prominent parts of true fingerprint tend to be removed. 
On the other hand, if $\beta$ is chosen too small, 
not all all unwanted features (as above) are removed which may  cause segmentation errors. 

In order to find good trade-offs, as described above, $n$, $L$, $\gamma$ and $\beta$ are trained as described in Section \ref{secExperiments}. In fact, since different fingerprint sensors have different properties, $\beta$ is adaptively adjusted to the intensity 
of coefficients in all subbands as
\begin{equation} \label{eq_beta}
  \beta = C \cdot \max_{l, \boldsymbol m} \{ c_l[\boldsymbol m] \}.
\end{equation}
Thus,
instead of $\beta$, $C$ is trained for each sensor.

\subsubsection{Texture Binarization}

In the first step, the texture is decomposed by the operator (\ref{eq8}) to obtain the reconstructed image $\tilde f[\boldsymbol k]$.
Then, $\tilde f[\boldsymbol k]$ is binarized using an adaptive threshold adjusted to the intensity of $\tilde f[\boldsymbol k]$. Thus, 
the threshold is chosen as $C \cdot \displaystyle \max_{\boldsymbol k}(\tilde f[\boldsymbol k])$, with $C$ from (\ref{eq_beta}). If $\tilde f[\boldsymbol k]$ 
is larger than this threshold, it will be set to 1 (foreground), otherwise, it is set to 0 (background) as illustrated in Figure~\ref{figAlgorithmOverview}.
\begin{equation*}
  \tilde f_{\text{bin}}[\boldsymbol k] = \begin{cases}  
                                           1, & \tilde f[\boldsymbol k] \geq C \cdot \displaystyle \max_{\boldsymbol k}(\tilde f[\boldsymbol k])   \\   0, & \text{else}
                                         \end{cases}
  , \qquad \forall \boldsymbol k \in \Omega
\end{equation*}
% -----------------------------------------------------------------------------------------------------

\subsubsection{Morphological Processing}

In this final phase, we apply mathematical morphology (see Chapter 13 in \cite{SonkaHlavacBoyle2008}),
to decide for each pixel whether it belongs to the foreground or background.
Firstly, at each pixel 
$\tilde f_{\text{bin}}[k_1,k_2] \in \{ 0, 1 \}$, we build an $s \times s$ block centered at $(k_1,k_2)$ and 8 neighboring blocks 
(cf. Figure \ref{figMorphologStructureElementy}). Then, for each block, we count
the white pixels and check whether their number exceeds the threshold $\frac{s^2}{t}$ with another parameter $t>0$. 
If at least $b$ blocks are above threshold, the pixel 
$[k_1,k_2]$ %\tilde f_{\text{bin}}[k_1,k_2]$ 
is considered as foreground.\\
\begin{minipage}{0.2\textwidth}
 \centering
 \includegraphics[width=1.0\textwidth]{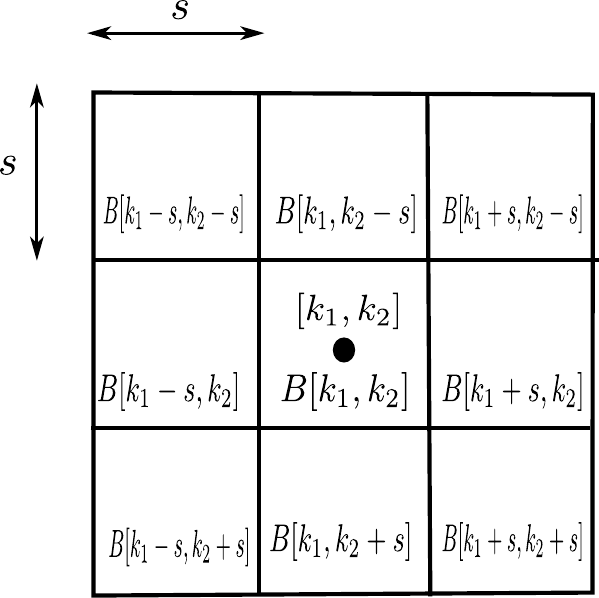}
 \captionof{figure}{The morphological element.}
 \label{figMorphologStructureElementy}
\end{minipage}
\begin{minipage}{0.76\textwidth}
\begin{equation} \label{eq:f_dilate}
  \tilde f_\text{dilate} [k_1,k_2] = 
  \begin{cases}
    1,  & \# \left \{ \sum B_{ [ k_1+m, k_2+m ] } \geq \frac{s^2}{t}, m \in \{ -s, 0, s \} \right \} \geq b		\\
    0,  & else
  \end{cases}
\end{equation}
\end{minipage}
\\

Then, the largest connected white pixel component is selected by a region filling method.
Its convex hull is then the ROI. 
For better visualization we have inverted white and black, i.e. display the background 
by white pixels and the ROI by black pixels, cf. Figure \ref{figAlgorithmOverview}.

\section{Evaluation Benchmark and Results} \label{secBenchmarkResults}

The databases of FVC2000, 2002 and 2004 \cite{FVC2000,FVC2002,FVC2004} are publicly available and established benchmarks 
for measuring the verification performance of algorithms for image enhancement and fingerprint matching.
Each competition comprises four databases: three of which contain real fingerprints acquired 
by different sensors and a database of synthetically generated images (DB 4 in each competition).

It has recently been shown that real and synthetic fingerprints can be discriminated with very high accuracy
using minutiae histograms (MHs) \cite{GottschlichHuckemann2014}.  
More specifically, by computing the MH for a minutiae template 
and then computing the earth mover's distance (EMD) \cite{GottschlichSchuhmacher2014} 
between the template MH and the mean MHs for a set of real and synthetic fingerprints.
Classification is simply performed by choosing the class with the smaller EMD.

The nine databases containing real fingerprints have been obtained by nine different sensors
and have different properties. The fingerprint image quality ranges from 
good quality images (especially FVC2002 DB1 and DB2) 
to low quality images which are more challenging to process (e.g. the databases of FVC2004).
Some aspects of image quality concern both the segmentation step and the overall verification process,
other aspects pose problems only for later stages of the fingerprint verification procedure, 
but have no influence on the segmentation accuracy. 

Aspects of fingerprint image quality which complicate the segmentation:
\begin{itemize}
 \item dryness or wetness of the finger
 \item a ghost fingerprint on the sensor surface
 \item small scale noise
 \item large scale structure noise 
 \item image artifacts e.g. caused by reconstructing a swipe sensor image
 \item scars or creases interrupting the fingerprint pattern
\end{itemize}

Aspects of fingerprint image quality which make an accurate verification
more difficult, but do not have any influence on the fingerprint segmentation step:
\begin{itemize}
 \item distortion, nonlinear deformation of the finger
 \item small overlap area between two imprints
\end{itemize}

Each of the 12 databases contains 110 fingers with 8 impressions per finger.
The training set consists of 10 fingers (80 images) 
and the test set contains 100 fingers (800 images).
In total there are 10560 fingerprint images giving 10560 marked ground truth segmentations
for training and testing.

\subsection{Experimental Results} \label{secExperiments}

\paragraph{Segmentation Performance Evaluation}

Let $N_1$ and $N_2$ be the width and height of image $f[\boldsymbol{k}]$ in pixels.
Let $M_f$ be number of pixels which are marked as foreground by human expert
and estimated as background by an algorithm (missed/misclassified foreground). 
Let $M_b$ be number of pixels which are marked as background by human expert
and estimated as foreground by an algorithm (missed/misclassified background). 
The average total error per image is defined as
\begin{equation}
 Err = \frac{M_f + M_b}{N_1 \times N_2}.
\end{equation}
The average error over 80 training images is basis for the parameter selection.
In Table~\ref{tableResults}, we report 
the average error over all other 800 test images for each database and for each algorithms.

\paragraph{Parameter Selection}

\begin{table}
\begin{center}
  \begin{tabular}{|c|l|}
    \hline
    Parameters		& Description
    \\
    \hline
    			& a constant for selecting the threshold $\beta$ in Eq. (\ref{eq_beta})  	
    \\
    \rb{$C$}		& which removes small coefficients corresponding to noise.
    \\
    \hline
    			& the order of the directional Hilbert transform which 
    \\
    \rb{$n$}		& corresponds to the angularpass filter in Eq. (\ref{Riesz-transform:def}).	
    \\
    \hline
    $L$			& the number of orientations in the angularpass filter in Eq. (\ref{Riesz-transform:def}).	
    \\
    \hline
    $\gamma$		& the order of the Butterworth bandpass filter in Eq. (\ref{eq:2D-ButterworthFilter}).	
    \\
    \hline
    $s$			& the window size of the block in the postprocessing step in Eq. (\ref{eq:f_dilate}).	
    \\
    \hline
    $t$			& a constant for selecting the morphology threshold $T$ in Eq. (\ref{eq:f_dilate}).
    \\    
    \hline
    $b$			& the number of the neighboring blocks in Eq. (\ref{eq:f_dilate}).	
    \\
    \hline
  \end{tabular}    
  \vspace*{8pt}
  \caption{Overview over all parameters for the factorized directional bandpass (FDB) method for fingerprint segmentation. 
	         Values are reported in Table~\ref{tabParameterChoice}.
          \label{tabParameters}
          } 
\end{center}
\end{table}

\begin{table}
\begin{center}
  \begin{tabular}{|c|c|c|c|c|}
    \hline
    FVC		& DB		& $C$		& $\gamma$	& $t$
    \\
    \hline
    2000	& 1		& 0.06		& 4		& 5
    \\
	        & 2		& 0.07		& 2		& 5
    \\
		& 3		& 0.06		& 4		& 4
    \\
	        & 4		& 0.03		& 1		& 5
    \\
    \hline
    2002	& 1		& 0.04		& 1		& 4
    \\
	        & 2		& 0.05		& 1		& 7
    \\
		& 3		& 0.09		& 1		& 5
    \\
		& 4		& 0.03		& 1		& 6
    \\
    \hline
    2004	& 1		& 0.04		& 1		& 7
    \\
	        & 2		& 0.08		& 2		& 5
    \\
		& 3		& 0.07		& 1		& 6
    \\
	        & 4		& 0.05		& 1		& 5
    \\
    \hline
  \end{tabular} 
  \vspace*{8pt}
  \caption{Overview over the parameters learned on the training set.
         The other four parameters are $n = 20$, $L = 16$, $s=9$ and $b = 6$ 
         for all databases.
         \label{tabParameterChoice}}
\end{center}
\end{table}  

Experiments were carried out on all 12 databases and are reported in Table~\ref{tableResults}.
For each method listed in Table~\ref{tableResults}, the required parameters were trained
on each of the 12 training sets: the choice of the threshold values 
for the Gabor filter bank based approach by Shen \textit{et al.} \cite{ShenKotKoo2001},
and the threshold values for the Harris corner response based method 
by Wu \textit{et al.} \cite{WuTulyakovGovindaraju2007}.
The parameters of the method by Bazen and Gerez are chosen as described in~\cite{BazenGerez2001}:
the window size of the morphology operator and 
the weights of the perceptron which are trained in $10^4$ iterations
due to the large number of pixels in the training database. 
For the method of Chikkerur \textit{et al.}, 
we used the energy image computed by the implementation of Chikkerur,
performed Otsu thresholding and mathematical morphology as explained in \cite{SonkaHlavacBoyle2008}.
 
For the proposed FDB method, the involved parameters are summarized in Table \ref{tabParameters}
and the values of the learned parameters are reported in Table \ref{tabParameterChoice}.
Also, the mirror boundary condition with size 15 pixels is used in order to avoid boundary effects.
In a reasonable amount of time, a number of conceivable parameter combinations 
were evaluated on the training set. 
The choice of these parameters balances the smoothing properties of the proposed filter
attempting to avoid both under-smoothing and over-smoothing.

\begin{table}
   \begin{center}
  \begin{tabular}{|c|c|r@{.} l|r @{.} l|r @{.} l|r @{.} l|r @{.} l|} \hline
	FVC	& DB	& \multicolumn{2}{c|}{GFB \cite{ShenKotKoo2001}} & \multicolumn{2}{c|}{HCR \cite{WuTulyakovGovindaraju2007}}	& \multicolumn{2}{c|}{MVC \cite{BazenGerez2001}}	& \multicolumn{2}{c|}{STFT \cite{ChikkerurCartwrightGovindaraju2007}}		& \multicolumn{2}{c|}{FDB}	
	\\
	\hline
	2000	& 1	& 13&26				& 11&15						& 10&01					& 16&70							& 5&51			 
		\\
		& 2	& 10&27				& 6&25						& 12&31					& 8&88							& 3&55			 
		\\
		& 3	& 10&63				& 7&80						& 7&45					& 6&44							& 2&86			 
		\\
		& 4	& 5&17				& 3&23						& 9&74					& 7&19							& 2&31			 
	\\
	\hline
	2002	& 1	& 5&07				& 3&71						& 4&59					& 5&49							& 2&39			 
		\\
		& 2	& 7&76				& 5&72						& 4&32					& 6&27							& 2&91			 
		\\
		& 3	& 9&60				& 4&71						& 5&29					& 5&13							& 3&35			 
		\\
		& 4	& 7&67				& 6&85						& 6&12					& 7&70							& 4&49			
	\\
	\hline
	2004	& 1	& 5&00				& 2&26						& 2&22					& 2&65							& 1&40			 
		\\
		& 2	& 11&18				& 7&54						& 8&06					& 9&89							& 4&90			 
		\\
		& 3	& 8&37				& 4&96						& 3&42					& 9&35							& 3&14		 
		\\
		& 4	& 5&96				& 5&15						& 4&58					& 5&18							& 2&79		 
	\\
	\hline
	Avg.	& 	& 8&33				& 5&78						& 6&51					& 7&57							& 3&30
	\\
	\hline 
  \end{tabular}
  \vspace*{8pt}
  \caption{ %\small 
  Error rates (average percentage of misclassified pixels averaged over 800 test images per database)
  computed using the manually marked ground truth segmentation 
  and the estimated segmentation by these methods:
  a Gabor filter bank (GFB) response based method by Shen \textit{et al.} \cite{ShenKotKoo2001},
  a Harris corner response (HCR) based approach by Wu \textit{et al.} \cite{WuTulyakovGovindaraju2007},
  a method by Bazen and Gerez using local gray-level mean, variance and gradient coherence (MVC)
  as features \cite{BazenGerez2001},
  a method applying short time Fourier transforms (STFT) by Chikkerur \textit{et al.} \cite{ChikkerurCartwrightGovindaraju2007}
  and the proposed method based on the factorized directional bandpass (FDB).
  \label{tableResults}}
  \end{center}  
  \end{table}

\begin{figure*}[ht]
\begin{center}   
    \subfigure{ \includegraphics[width=0.13\textwidth]{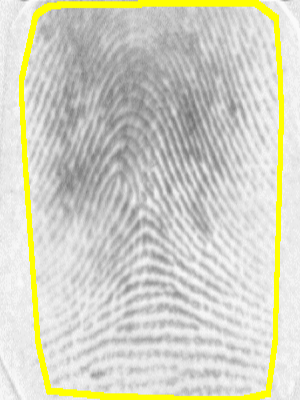} }
    \subfigure{ \includegraphics[width=0.13\textwidth]{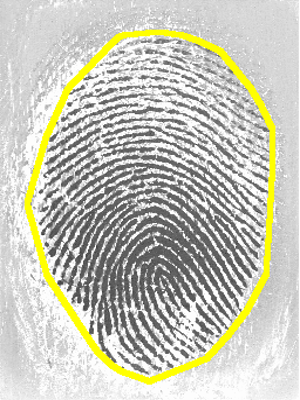} }
    \subfigure{ \includegraphics[width=0.13\textwidth]{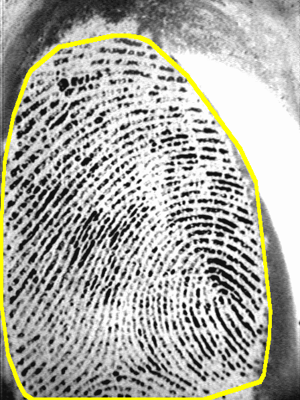} }
    \subfigure{ \includegraphics[width=0.13\textwidth]{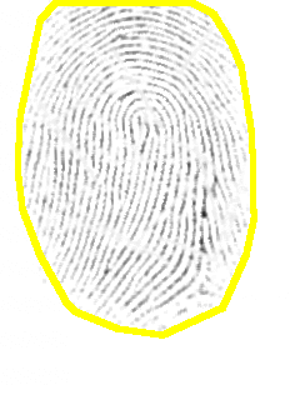} }
    \\
    \subfigure{ \includegraphics[width=0.13\textwidth]{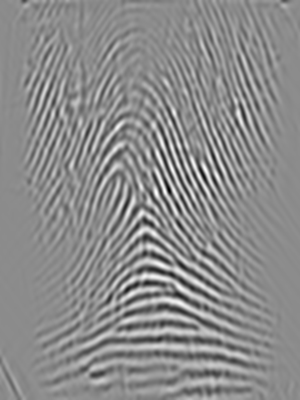} }
    \subfigure{ \includegraphics[width=0.13\textwidth]{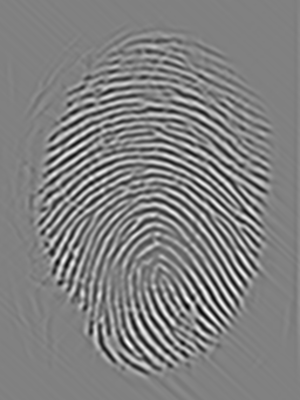} }
    \subfigure{ \includegraphics[width=0.13\textwidth]{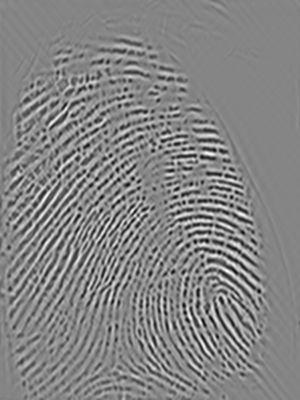} }
    \subfigure{ \includegraphics[width=0.13\textwidth]{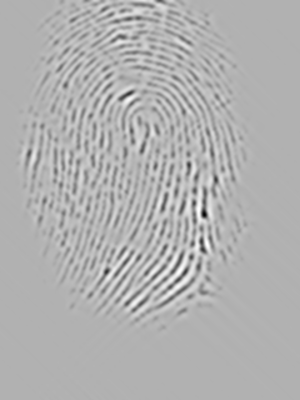} }
    \\
    \subfigure{ \includegraphics[width=0.13\textwidth]{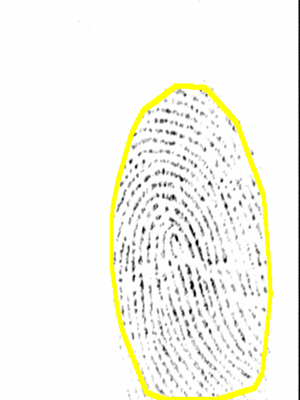} }
    \subfigure{ \includegraphics[width=0.13\textwidth]{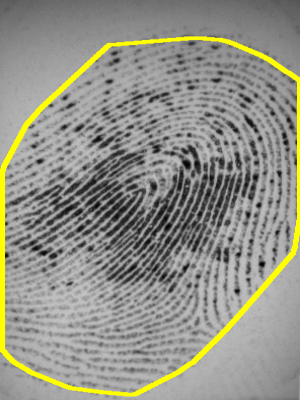} }
    \subfigure{ \includegraphics[width=0.13\textwidth]{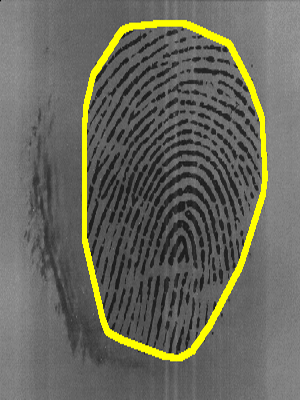} }
    \subfigure{ \includegraphics[width=0.13\textwidth]{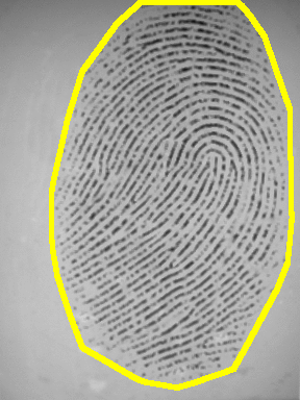} }
    \\
    \subfigure{ \includegraphics[width=0.13\textwidth]{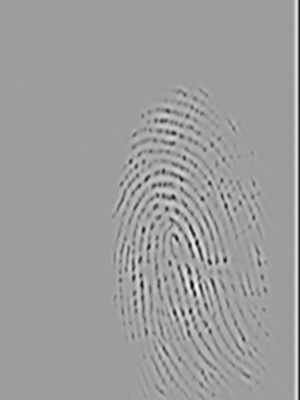} }
    \subfigure{ \includegraphics[width=0.13\textwidth]{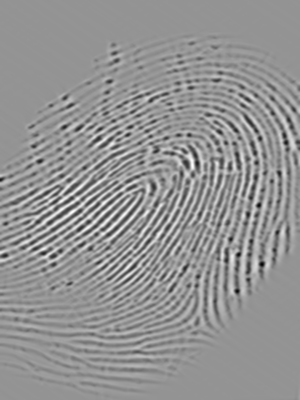} }
    \subfigure{ \includegraphics[width=0.13\textwidth]{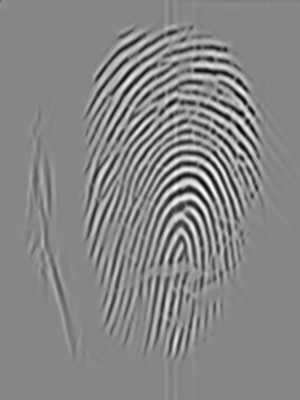} }
    \subfigure{ \includegraphics[width=0.13\textwidth]{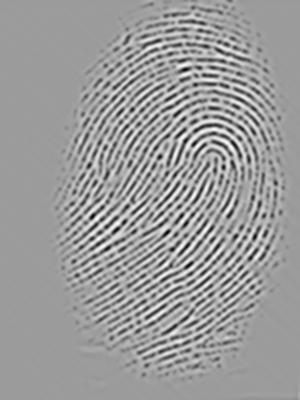} }    
    \\
    \subfigure{ \includegraphics[width=0.13\textwidth]{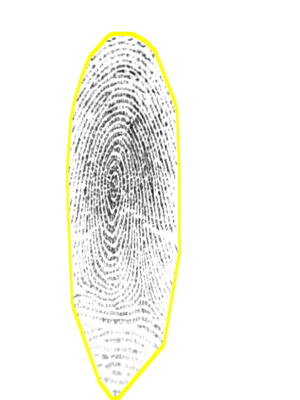} }
    \subfigure{ \includegraphics[width=0.13\textwidth]{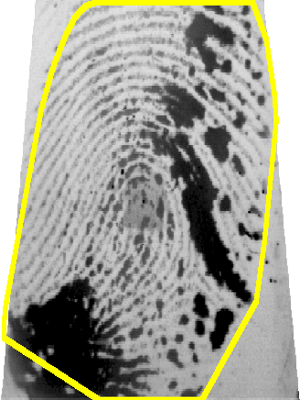} }
    \subfigure{ \includegraphics[width=0.13\textwidth]{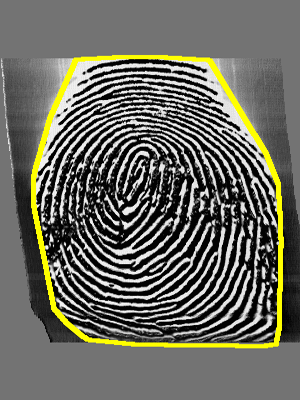} }
    \subfigure{ \includegraphics[width=0.13\textwidth]{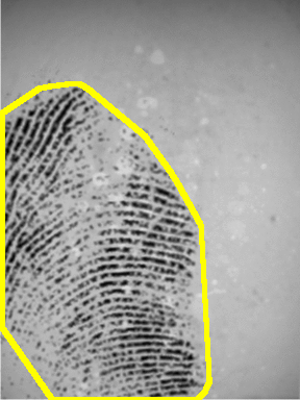} }
    \\
    \subfigure{ \includegraphics[width=0.13\textwidth]{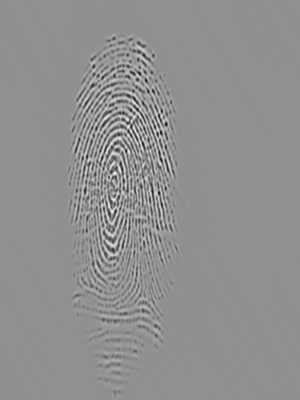} }
    \subfigure{ \includegraphics[width=0.13\textwidth]{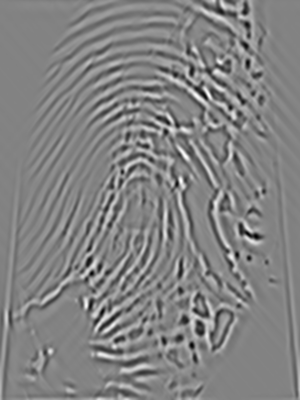} }
    \subfigure{ \includegraphics[width=0.13\textwidth]{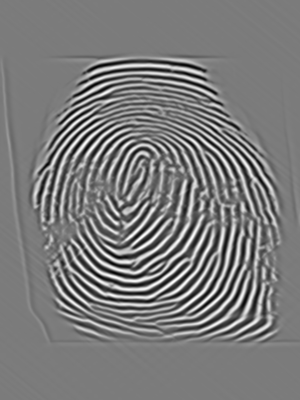} }
    \subfigure{ \includegraphics[width=0.13\textwidth]{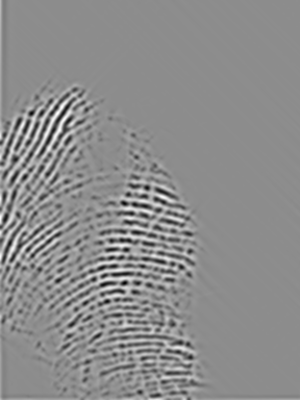} }
    
    \caption{Segmented fingerprint images and the corresponding reconstructed texture images 
              by the FDB method
              for FVC2000 (first and second row), FVC2002 (third and fourth row) 
 	      and FVC2004 (fifth and sixth row).
              Columns f.l.t.r correspond to DB1 to DB4.              
     \label{figSegmentationResultDHBBOverview}}
\end{center}
\end{figure*}

\begin{figure*}[ht]
\begin{center}   
    \subfigure[ Err = 25.23 ]{ \includegraphics[width=0.2\textwidth]{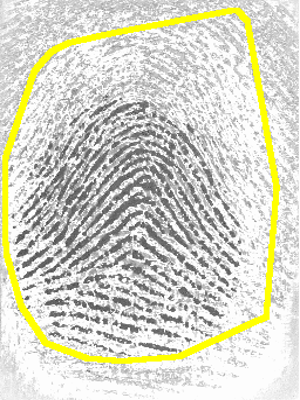} }
    \subfigure[ Err = 18.97 ]{ \includegraphics[width=0.2\textwidth]{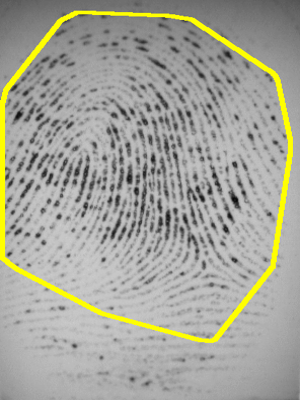} }
    \subfigure[ Err = 11.39 ]{ \includegraphics[width=0.2\textwidth]{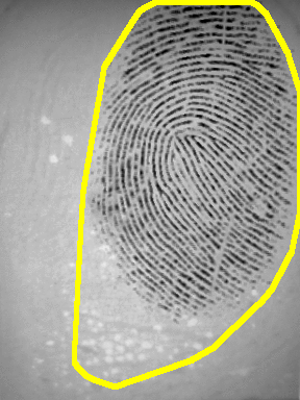} }
    \subfigure[ Err = 5.98 ]{ \includegraphics[width=0.2\textwidth]{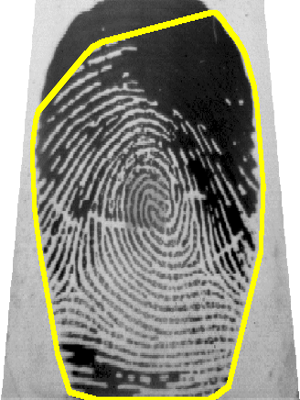} }
    \\    
    \subfigure{ \includegraphics[width=0.2\textwidth]{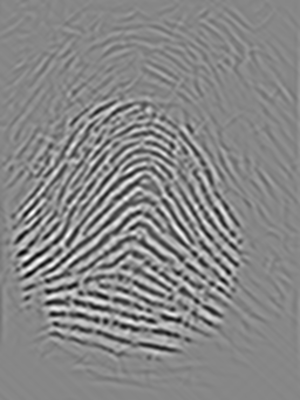} }
    \subfigure{ \includegraphics[width=0.2\textwidth]{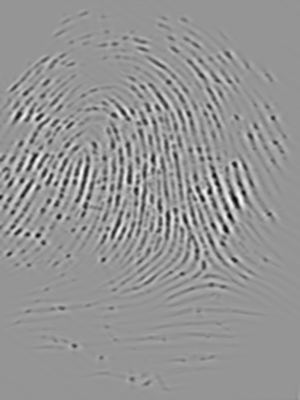} }
    \subfigure{ \includegraphics[width=0.2\textwidth]{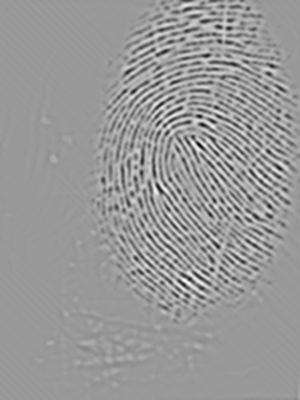} }
    \subfigure{ \includegraphics[width=0.2\textwidth]{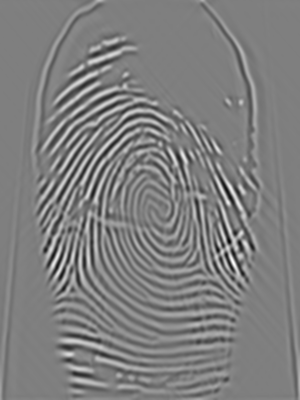} }
    \\
    \subfigure{ \includegraphics[width=0.2\textwidth]{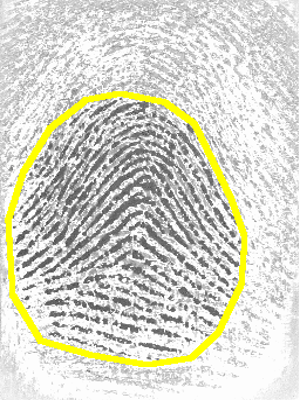} }
    \subfigure{ \includegraphics[width=0.2\textwidth]{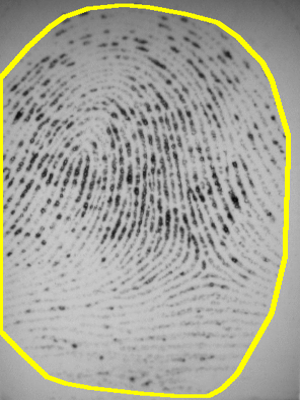} }
    \subfigure{ \includegraphics[width=0.2\textwidth]{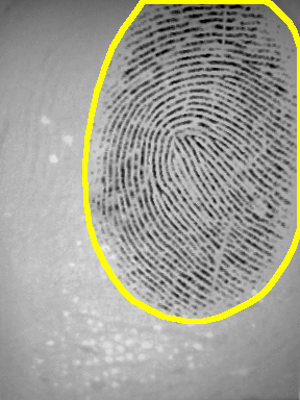} }
    \subfigure{ \includegraphics[width=0.2\textwidth]{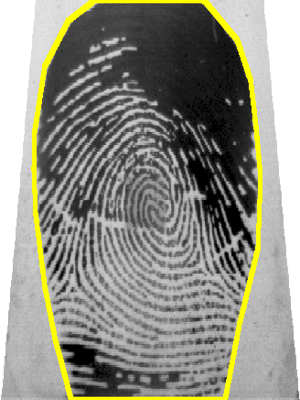} }
    
    \caption{Examples of incorrectly segmented fingerprint images due to: 
             (a) a ghost fingerprint on the sensor surface, 
             (b) dryness of the finger, 
             (c) texture artifacts in the reconstructed image,
					   (d) wetness of the finger.
             The first row shows the segmentation obtained by the FDB method, 
             the second row displays the reconstructed image 
             and the last row visualizes the manually marked ground truth segmentation.
             \label{figExamplesDHBBfail}}   
\end{center}
\end{figure*}

\begin{figure*}[ht]
\begin{center}
    \subfigure[ ]{ \includegraphics[width=0.22\textwidth]{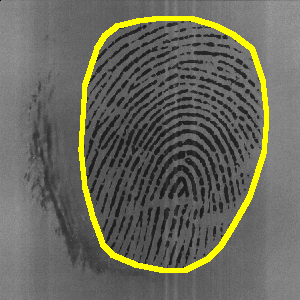} }

    \subfigure[ Err = 2.71 ]{ \includegraphics[width=0.22\textwidth]{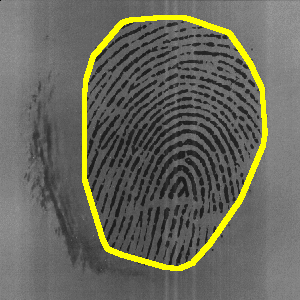} }
    \subfigure[ Err = 10.08 ]{ \includegraphics[width=0.22\textwidth]{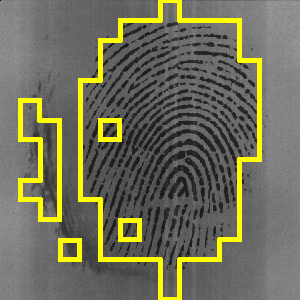} }
    \subfigure[ Err = 11.26 ]{ \includegraphics[width=0.22\textwidth]{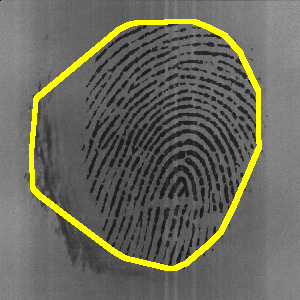} }
    \subfigure[ Err = 5.17 ]{ \includegraphics[width=0.22\textwidth]{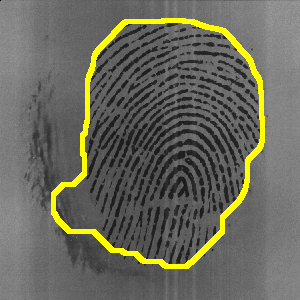} }
 
    \subfigure[ Err = 3.12 ]{ \includegraphics[width=0.22\textwidth]{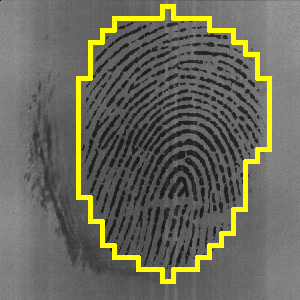} } 
    \subfigure[ ]{ \includegraphics[width=0.22\textwidth]{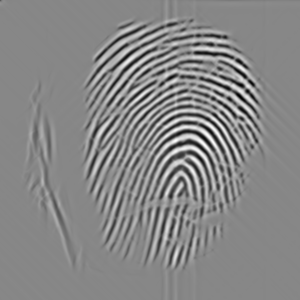} }  
    \subfigure[ ]{ \includegraphics[width=0.22\textwidth]{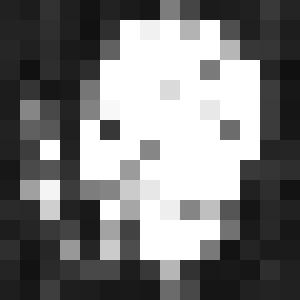} }
    \subfigure[ ]{ \includegraphics[width=0.22\textwidth]{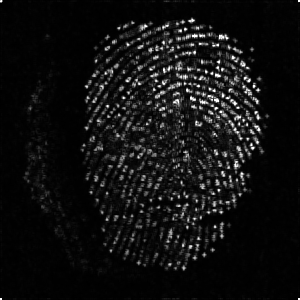} }

    \subfigure[ ]{ \includegraphics[width=0.22\textwidth]{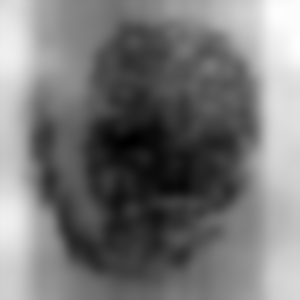} }
    \subfigure[ ]{ \includegraphics[width=0.22\textwidth]{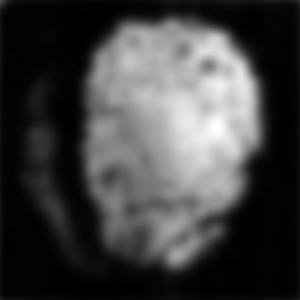} }
    \subfigure[ ]{ \includegraphics[width=0.22\textwidth]{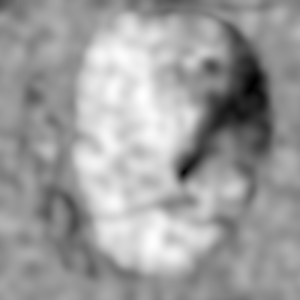} }
    \subfigure[ ]{ \includegraphics[width=0.22\textwidth]{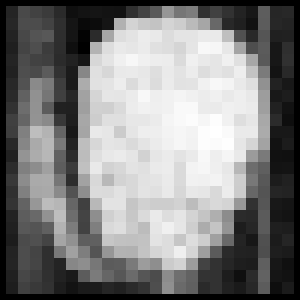} }
       
    \caption{ Segmented fingerprint images and their features of different methods for FVC2002\_DB3\_IM\_15\_1.
                    (a) ground truth; (b, g) FDB, (c, h) Gabor, (d, i) Harris, (e,  j, k, l) Mean-Variance-Coherence, (f, m) STFT. }
    \label{figComparison1}
\end{center}
\end{figure*}

\begin{figure*}[ht]
\begin{center}
    \subfigure[ ]{ \includegraphics[width=0.22\textwidth]{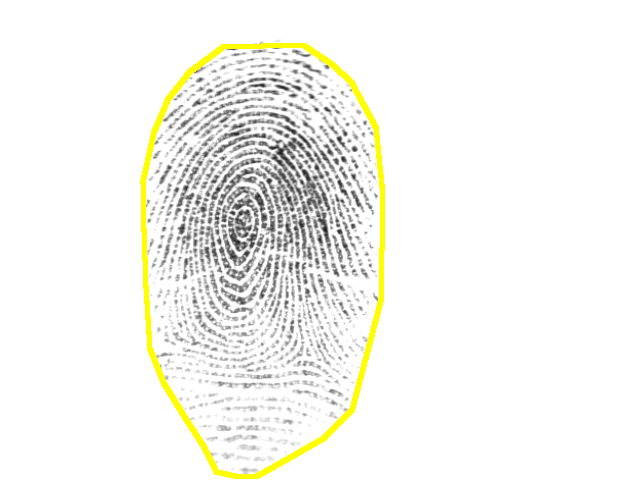} }

    \subfigure[ Err = 0.6 ]{ \includegraphics[width=0.22\textwidth]{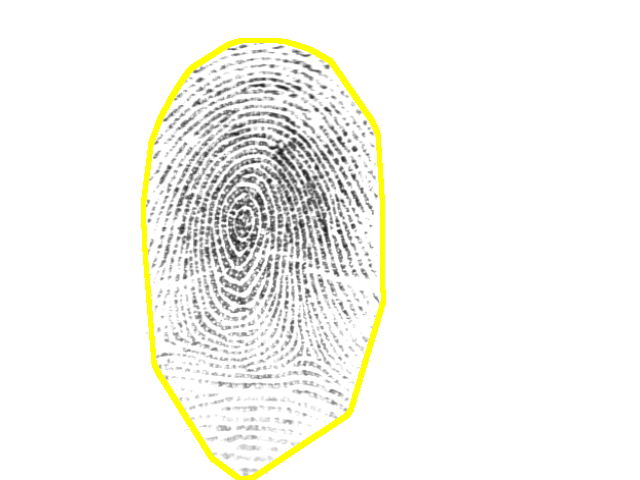} }
    \subfigure[ Err = 2.08 ]{ \includegraphics[width=0.22\textwidth]{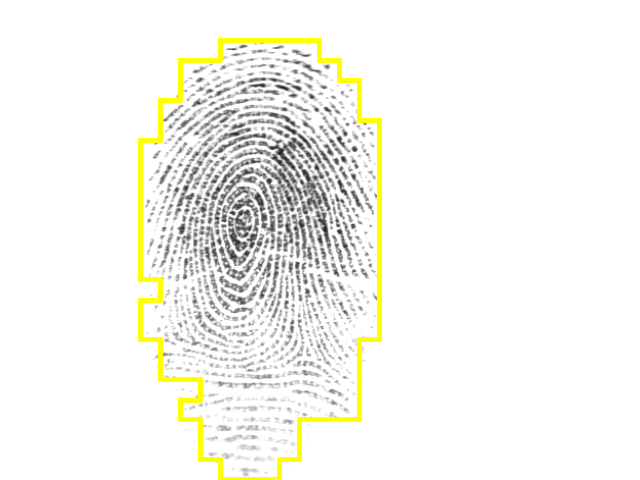} }
    \subfigure[ Err = 1.34 ]{ \includegraphics[width=0.22\textwidth]{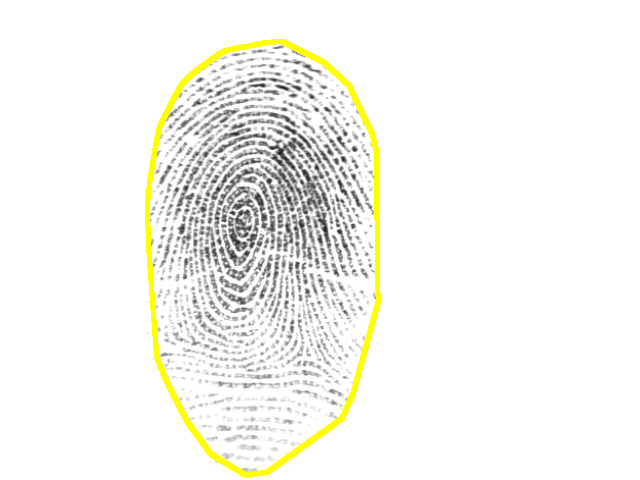} }
    \subfigure[ Err = 3.87 ]{ \includegraphics[width=0.22\textwidth]{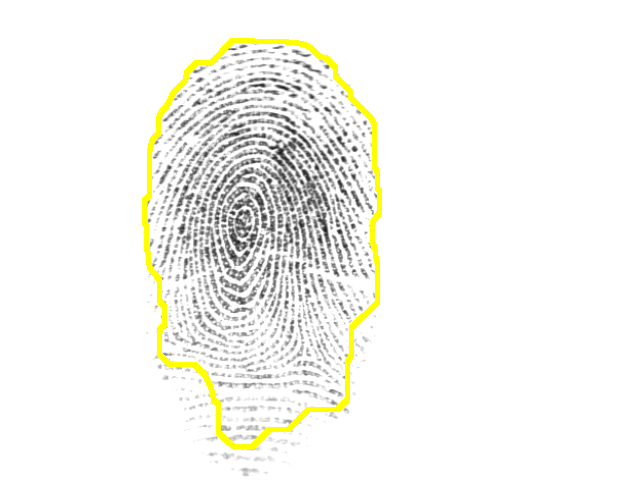} }
 
    \subfigure[ Err = 2.45 ]{ \includegraphics[width=0.22\textwidth]{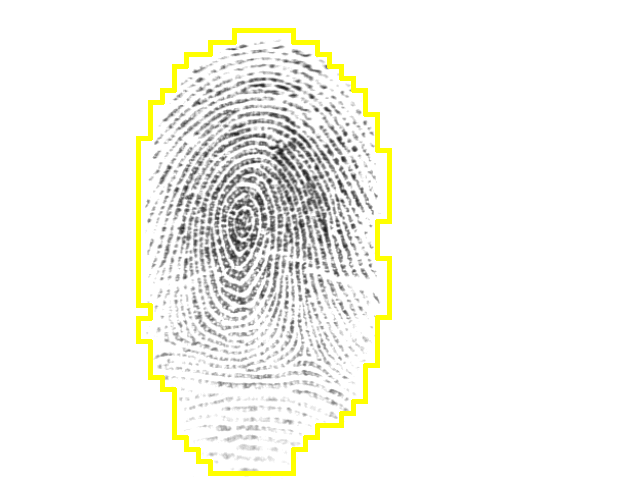} } 
    \subfigure[ ]{ \includegraphics[width=0.22\textwidth]{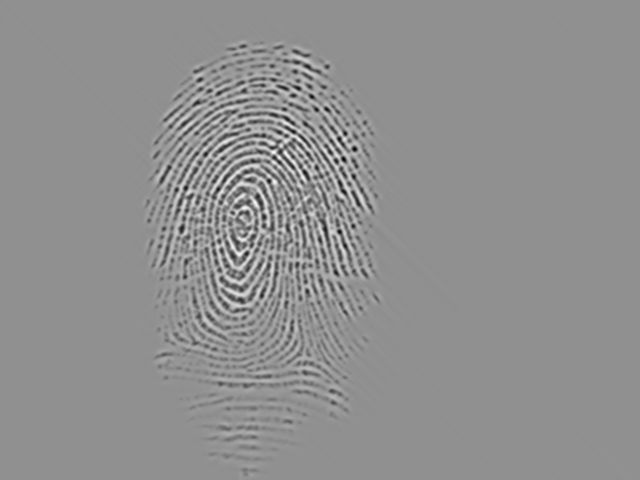} }  
    \subfigure[ ]{ \includegraphics[width=0.22\textwidth]{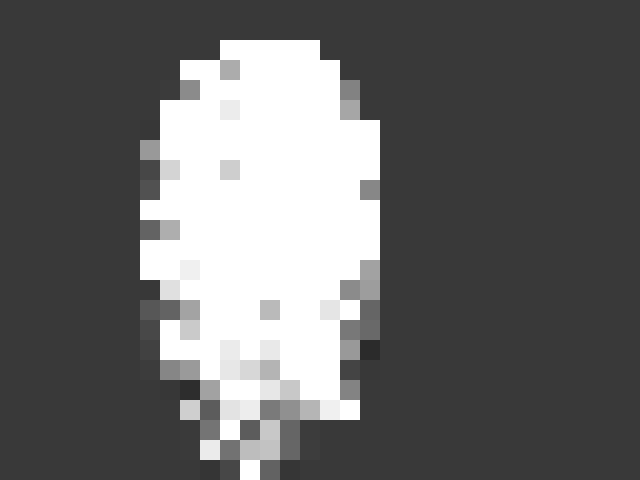} }
    \subfigure[ ]{ \includegraphics[width=0.22\textwidth]{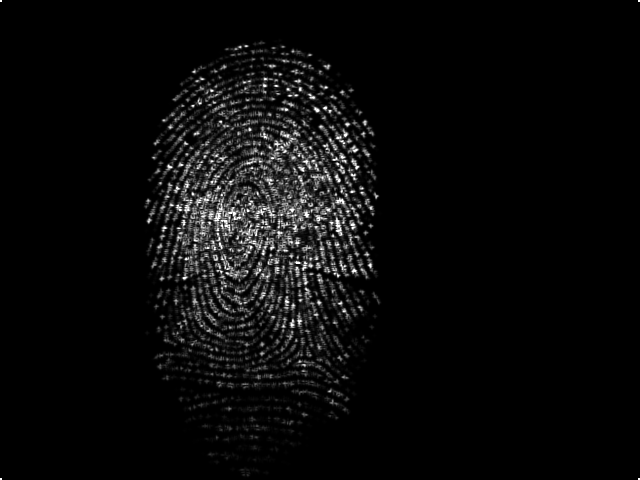} }

    \subfigure[ ]{ \includegraphics[width=0.22\textwidth]{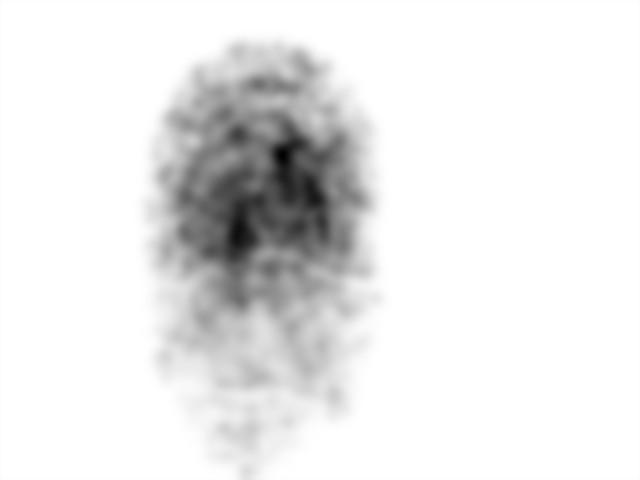} }
    \subfigure[ ]{ \includegraphics[width=0.22\textwidth]{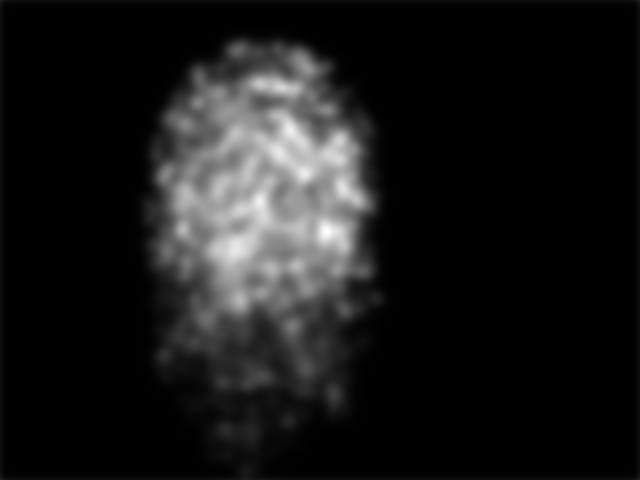} }
    \subfigure[ ]{ \includegraphics[width=0.22\textwidth]{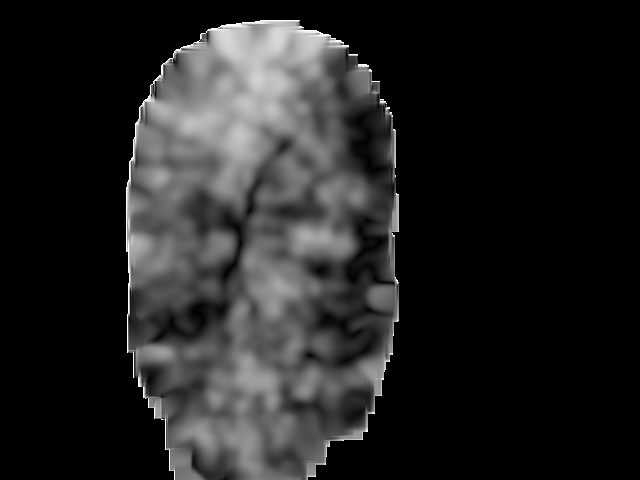} }
    \subfigure[ ]{ \includegraphics[width=0.22\textwidth]{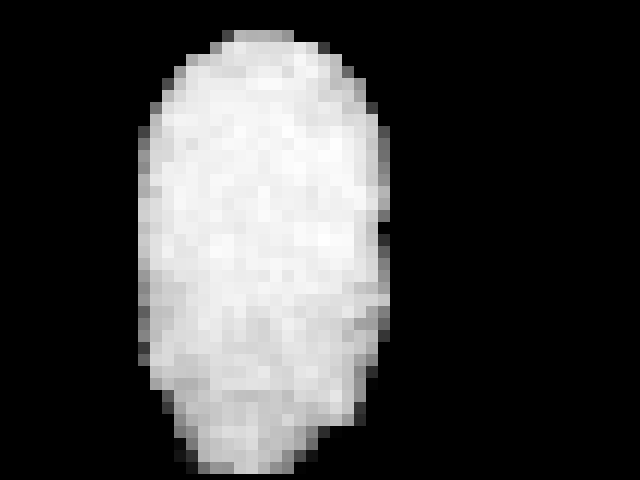} }
       
    \caption{ Segmented fingerprint images and their features of different methods for FVC2004\_DB1\_IM\_24\_7.
                    (a) ground truth; (b, g) FDB, (c, h) Gabor, (d, i) Harris, (e,  j, k, l) Mean-Variance-Coherence, (f, m) STFT. }
    \label{figComparison2}
\end{center}
\end{figure*}

\begin{figure*}[ht]
\begin{center}
    \subfigure[ ]{ \includegraphics[width=0.22\textwidth]{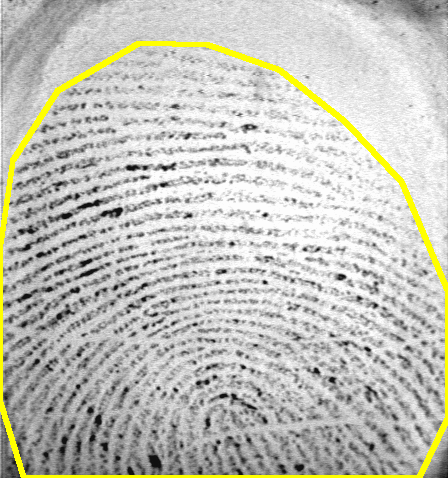} }

    \subfigure[ Err = 7.03 ]{ \includegraphics[width=0.22\textwidth]{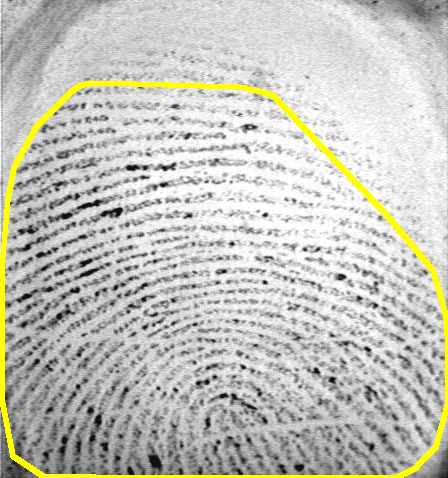} }
    \subfigure[ Err = 20.55 ]{ \includegraphics[width=0.22\textwidth]{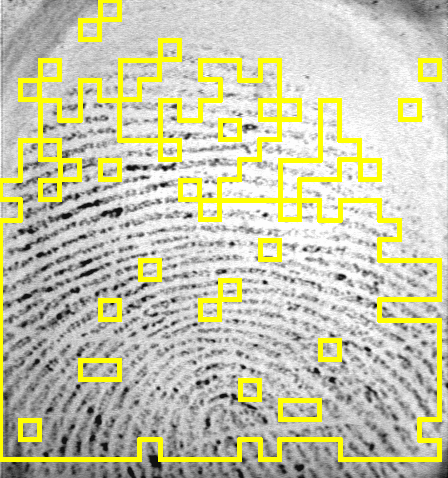} }
    \subfigure[ Err = 14.98 ]{ \includegraphics[width=0.22\textwidth]{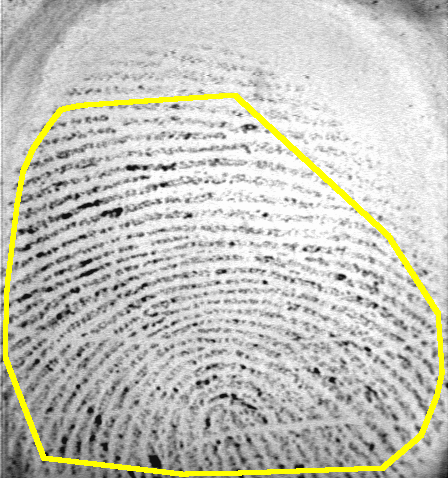} }
    \subfigure[ Err = 21.22 ]{ \includegraphics[width=0.22\textwidth]{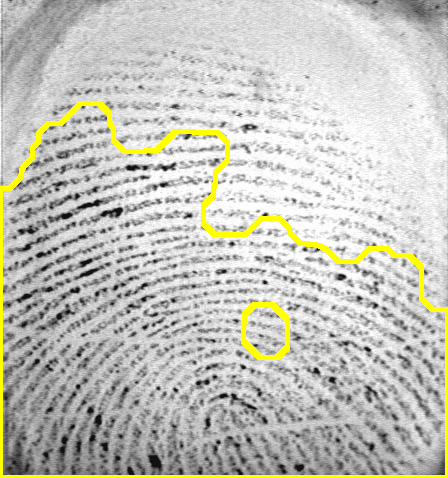} }
 
    \subfigure[ Err = 13.45 ]{ \includegraphics[width=0.22\textwidth]{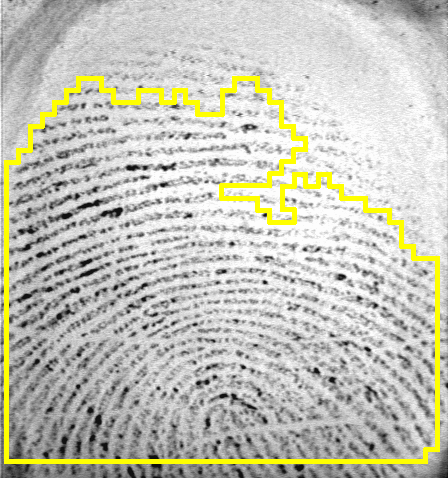} } 
    \subfigure[ ]{ \includegraphics[width=0.22\textwidth]{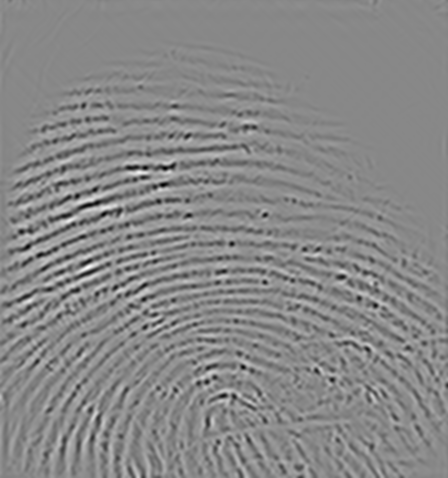} }  
    \subfigure[ ]{ \includegraphics[width=0.22\textwidth]{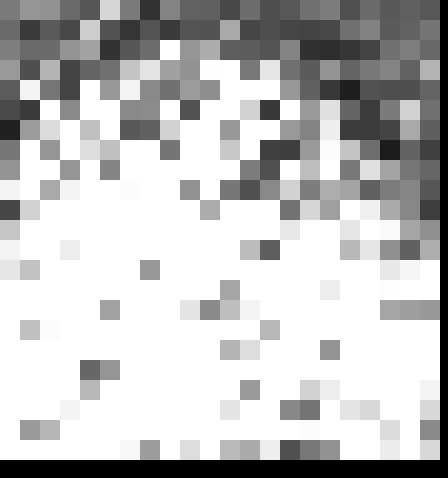} }
    \subfigure[ ]{ \includegraphics[width=0.22\textwidth]{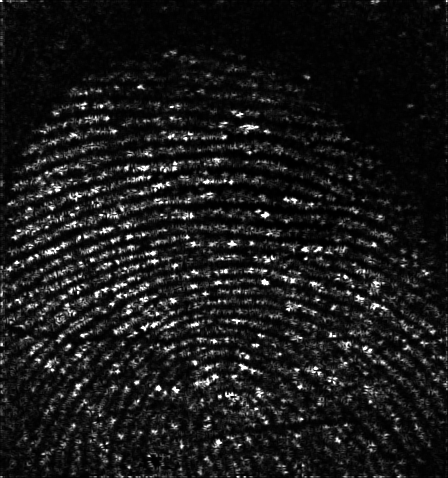} }

    \subfigure[ ]{ \includegraphics[width=0.22\textwidth]{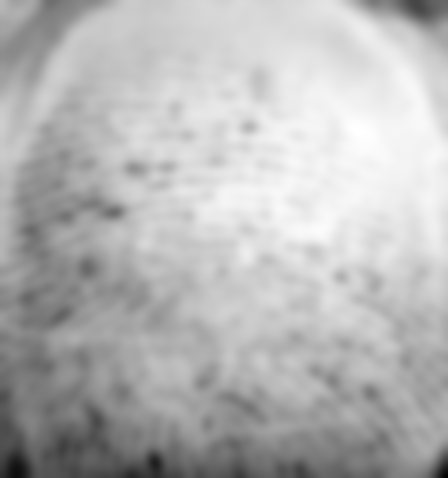} }
    \subfigure[ ]{ \includegraphics[width=0.22\textwidth]{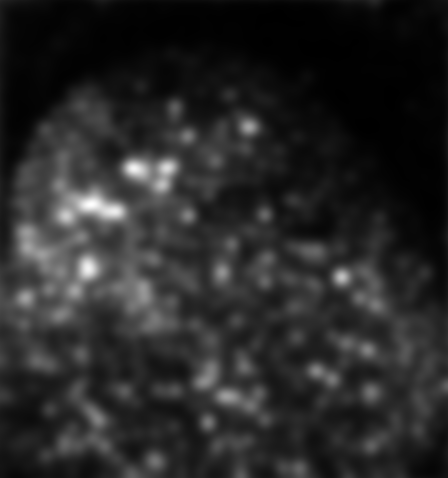} }
    \subfigure[ ]{ \includegraphics[width=0.22\textwidth]{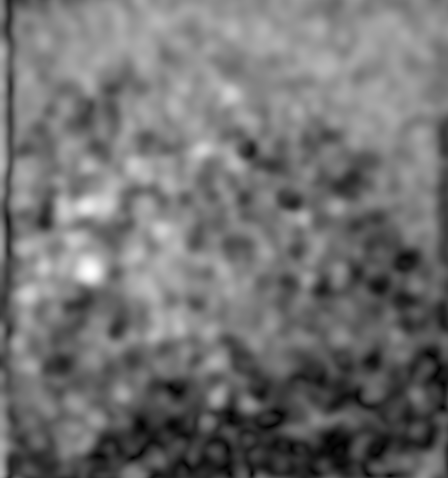} }
    \subfigure[ ]{ \includegraphics[width=0.22\textwidth]{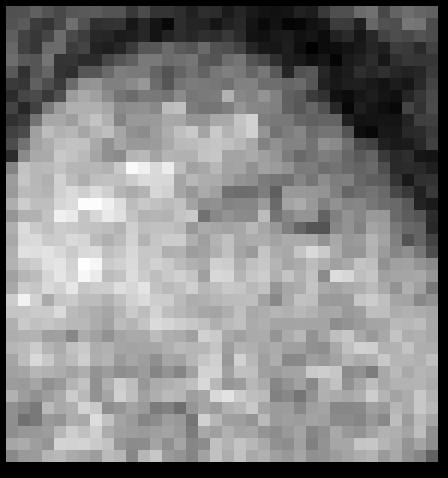} }
       
    \caption{ Segmented fingerprint images and their features of different methods for FVC2000\_DB3\_IM\_17\_3.
                    (a) ground truth; (b, g) FDB, (c, h) Gabor, (d, i) Harris, (e,  j, k, l) Mean-Variance-Coherence, (f, m) STFT. }
    \label{figComparison3}
\end{center}
\end{figure*}

\begin{figure*}[ht]
\begin{center}
    \subfigure[ ]{ \includegraphics[width=0.22\textwidth]{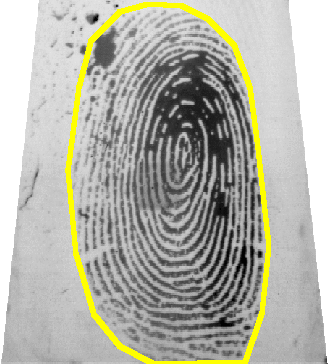} }

    \subfigure[ Err = 5.24 ]{  \includegraphics[width=0.22\textwidth]{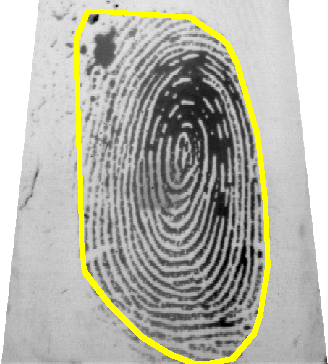} }
    \subfigure[ Err = 13.83 ]{ \includegraphics[width=0.22\textwidth]{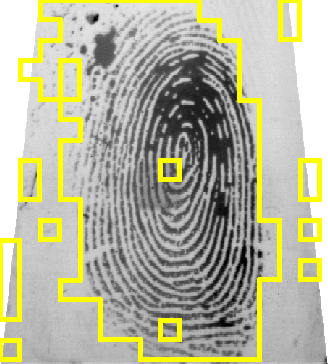} }
    \subfigure[ Err = 15.57 ]{ \includegraphics[width=0.22\textwidth]{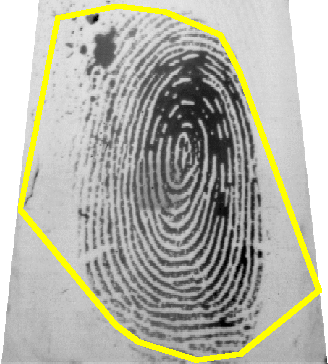} }
    \subfigure[ Err = 6.74 ]{  \includegraphics[width=0.22\textwidth]{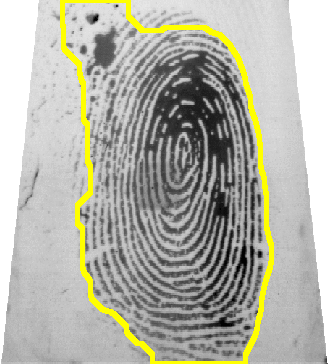} }
 
    \subfigure[ Err = 5.62 ]{ \includegraphics[width=0.22\textwidth]{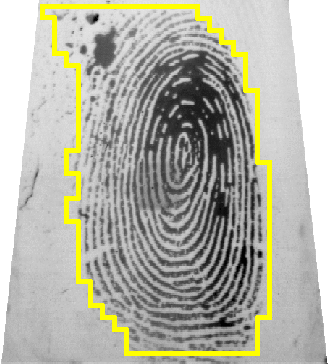} } 
    \subfigure[ ]{ \includegraphics[width=0.22\textwidth]{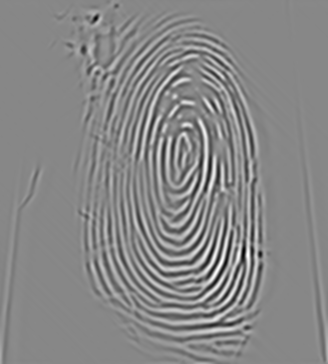} }  
    \subfigure[ ]{ \includegraphics[width=0.22\textwidth]{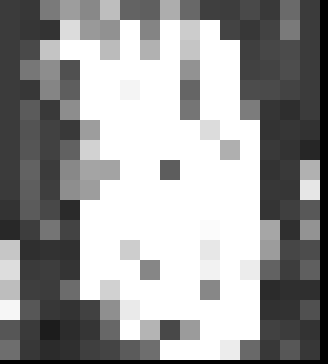} }
    \subfigure[ ]{ \includegraphics[width=0.22\textwidth]{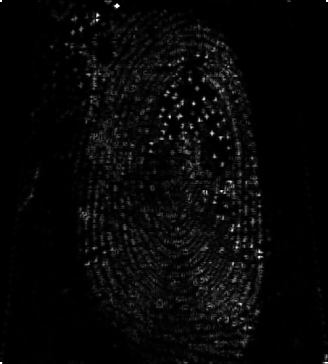} }
    
    \subfigure[ ]{ \includegraphics[width=0.22\textwidth]{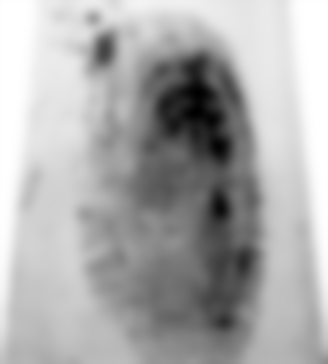} }
    \subfigure[ ]{ \includegraphics[width=0.22\textwidth]{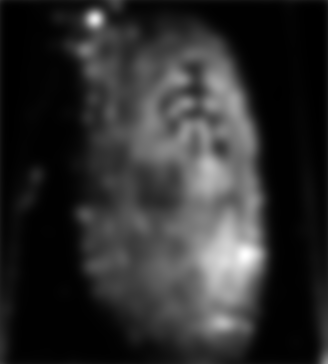} }
    \subfigure[ ]{ \includegraphics[width=0.22\textwidth]{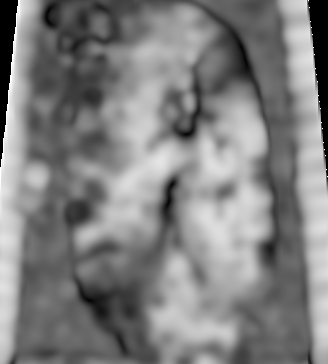} }
    \subfigure[ ]{ \includegraphics[width=0.22\textwidth]{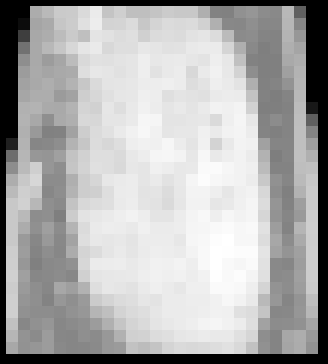} }
       
    \caption{ Segmented fingerprint images and their features of different methods for FVC2004\_DB2\_IM\_56\_8.
                    (a) ground truth; (b, g) FDB, (c, h) Gabor, (d, i) Harris, (e,  j, k, l) Mean-Variance-Coherence, (f, m) STFT. }
    \label{figComparison4}
\end{center}
\end{figure*}

\begin{figure*}[ht]
\begin{center}
    \subfigure[ ]{ \includegraphics[width=0.22\textwidth]{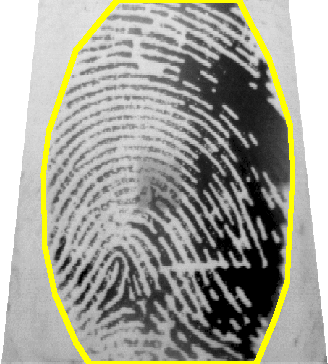} }

    \subfigure[ Err = 5.51 ]{ \includegraphics[width=0.22\textwidth]{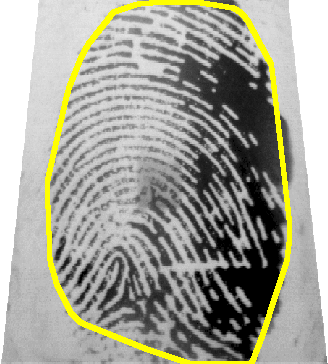} }
    \subfigure[ Err = 12.43 ]{ \includegraphics[width=0.22\textwidth]{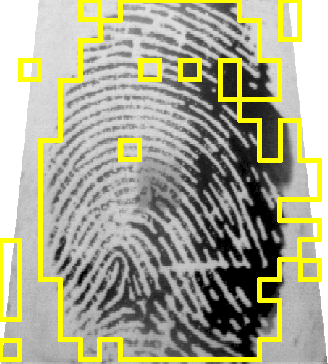} }
    \subfigure[ Err = 6.46 ]{ \includegraphics[width=0.22\textwidth]{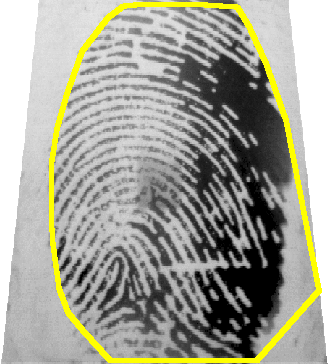} }
    \subfigure[ Err = 4.47 ]{ \includegraphics[width=0.22\textwidth]{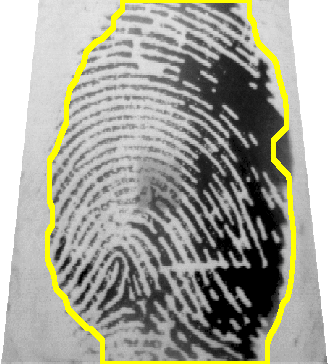} }
 
    \subfigure[ Err = 6.42 ]{ \includegraphics[width=0.22\textwidth]{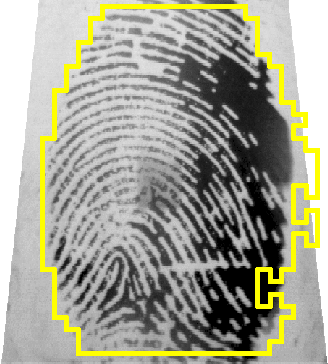} } 
    \subfigure[ ]{ \includegraphics[width=0.22\textwidth]{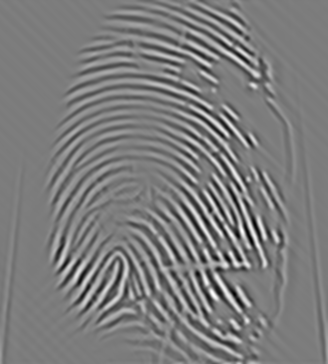} }  
    \subfigure[ ]{ \includegraphics[width=0.22\textwidth]{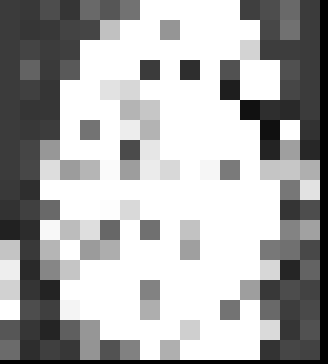} }
    \subfigure[ ]{ \includegraphics[width=0.22\textwidth]{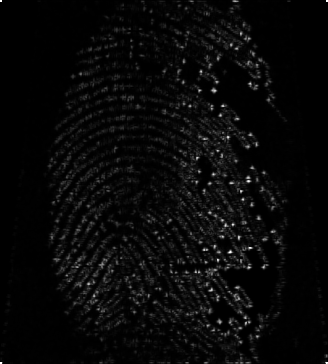} }
    
    \subfigure[ ]{ \includegraphics[width=0.22\textwidth]{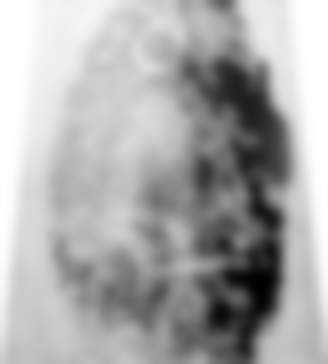} }
    \subfigure[ ]{ \includegraphics[width=0.22\textwidth]{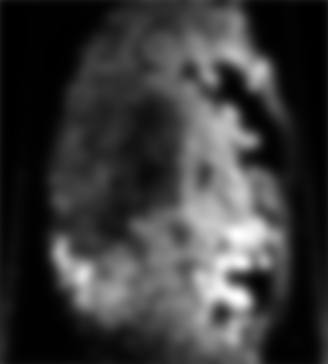} }
    \subfigure[ ]{ \includegraphics[width=0.22\textwidth]{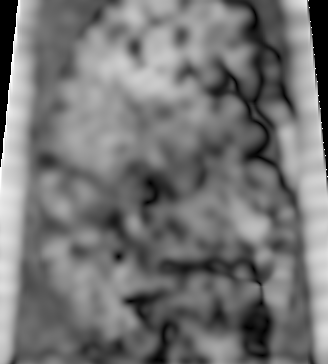} }
    \subfigure[ ]{ \includegraphics[width=0.22\textwidth]{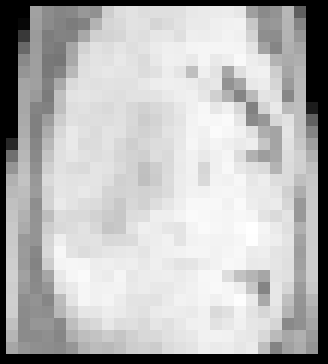} }
       
    \caption{ Segmented fingerprint images and their features of different methods for FVC2004\_DB2\_IM\_65\_7.
                    (a) ground truth; (b, g) FDB, (c, h) Gabor, (d, i) Harris, (e,  j, k, l) Mean-Variance-Coherence, (f, m) STFT. }
    \label{figComparison5}
\end{center}
\end{figure*}

This systematic comparison of fingerprint segmentation methods 
clearly shows that the factorized directional bandpass method (FDB)
outperforms the other four widely used segmentation methods on all 12 databases.
An overview of visualized segmentation results by the FDB method 
is given in Figure~\ref{figSegmentationResultDHBBOverview}.
A few challenging examples for which the FDB method produces a flawed segmentation 
are depicted in Figure \ref{figExamplesDHBBfail}.
Moreover, a comparison of all five segmentation methods and their main features 
for five example images 
are shown in Figure \ref{figComparison1} to \ref{figComparison5}.

\section{Conclusions} \label{secConclusions}

In this paper, we designed a filter specifically for fingerprints which is
based on the directional Hilbert transform of Butterworth bandpass filters.
A systematic comparison with four widely used fingerprint segmentation 
showed that the proposed FDB method outperforms these methods on all 12 FVC databases
using manually marked ground truth segmentation for the performance evaluation.
The proposed FDB method for fingerprint segmentation 
can be combined with all methods for orientation field estimation like 
e.g. the line sensor method \cite{GottschlichMihailescuMunk2009}
or by a global model based on quadratic differentials \cite{HuckemannHotzMunk2008}
followed by liveness detection \cite{GottschlichMarascoYangCukic2014} or
fingerprint image enhancement \cite{Gottschlich2012,GottschlichSchoenlieb2012}.
It can also be used in combination with alternative approaches, 
e.g. as a preprocessing step for locally adaptive fingerprint enhancement 
in the Fourier domain as proposed by Bart\r{u}n\v{e}k \textit{et al.} \cite{BartunekNilssonSallbergClaesson2013}
or before applying structure tensor derived symmetry features
for enhancement and minutiae extraction proposed by Fronthaler \textit{et al.} \cite{FronthalerKollreiderBigun2008}.

Notably, the filter $\phi_l^{n,\gamma}*\phi_l^{n,\gamma,\vee}$ is similar to the Gabor filter which could have been used instead of the DHBB filter. Similarly, Bessel or Chebbychev transforms as well as B-splines as generalizations (\cite{VandevilleBluUnser2005})
could replace the Butterworth. We expect, however, for reasons elaborated, relying on the DHBB filter gives superior segmentation results.

The manually marked ground truth benchmark and the implementation of the FDB method
are available for download at\\
\url{www.stochastik.math.uni-goettingen.de/biometrics/}.

In doing so, we would like to facilitate the reproducibility of the presented results 
and promote the comparability of fingerprint segmentation methods.

\section*{Acknowledgements}

S. Huckemann and C. Gottschlich gratefully acknowledge the support of the 
Felix-Bernstein-Institute for Mathematical Statistics in the Biosciences 
and the Niedersachsen Vorab of the Volkswagen Foundation.
The authors would like to thank Benjamin Eltzner, Axel Munk, Gerlind Plonka-Hoch and Yvo Pokern
for their valuable comments.

%\bibliography{../../FP}
% \bibliographystyle{plain}
%\bibliography{Carsten20141014}
%\bibliographystyle{unsrt}

% ---------------------------------------------------------------------------------------------

\clearpage

\section*{Supplementary Appendix}

\section{Comparison of the Operator in the FDB Method with the Summation and Maximum Operators} \label{AppendixC}

\begin{figure*}[ht] 
\begin{center}
    % Segmented images:
    \subfigure[Err = 1.98]{ \includegraphics[width=0.15\textwidth]{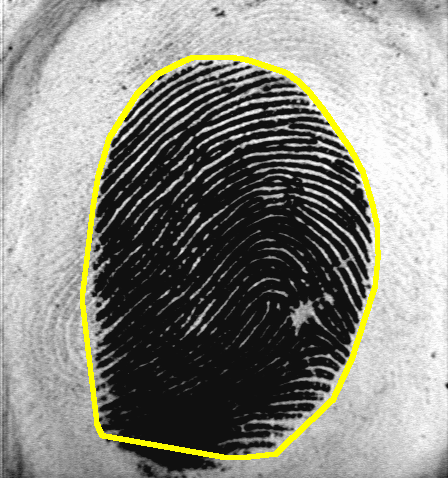} }		
    \subfigure{ \includegraphics[width=0.15\textwidth]{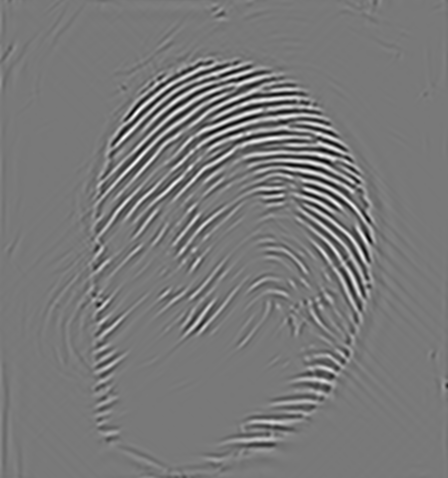} }		
    \subfigure[Err = 1.44]{ \includegraphics[width=0.15\textwidth]{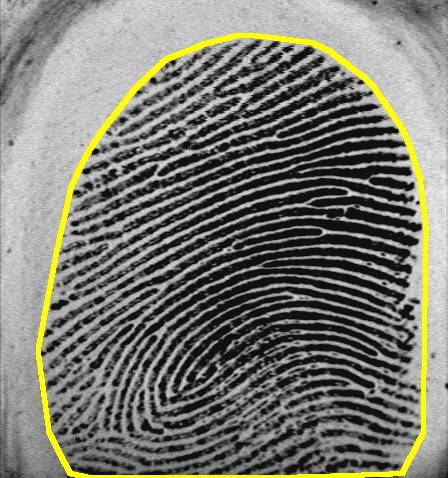} }		
    \subfigure{ \includegraphics[width=0.15\textwidth]{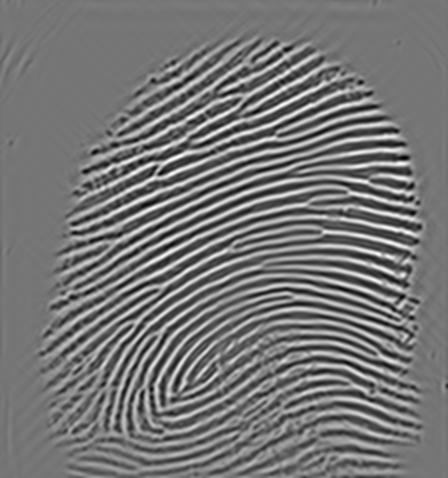} }		
    
    \subfigure[Err = 7.18]{ \includegraphics[width=0.15\textwidth]{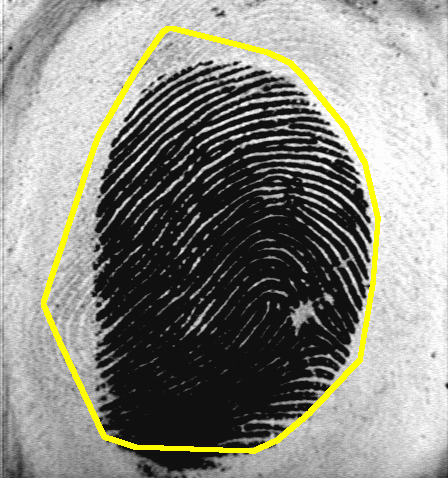} }	
    \subfigure{ \includegraphics[width=0.15\textwidth]{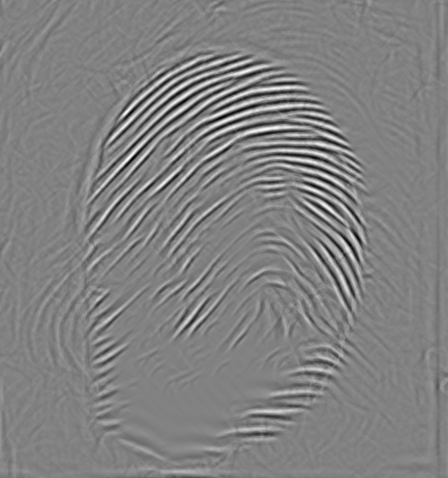} }	
    \subfigure[Err = 5.1]{ \includegraphics[width=0.15\textwidth]{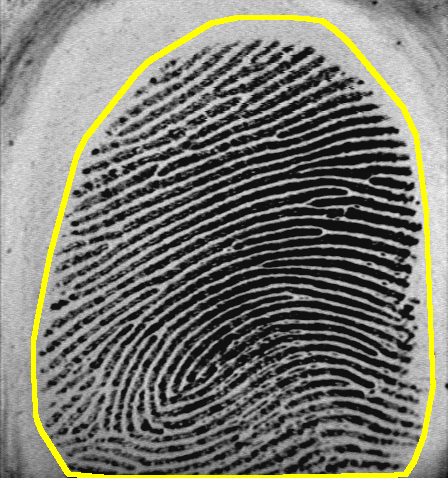} }	
    \subfigure{ \includegraphics[width=0.15\textwidth]{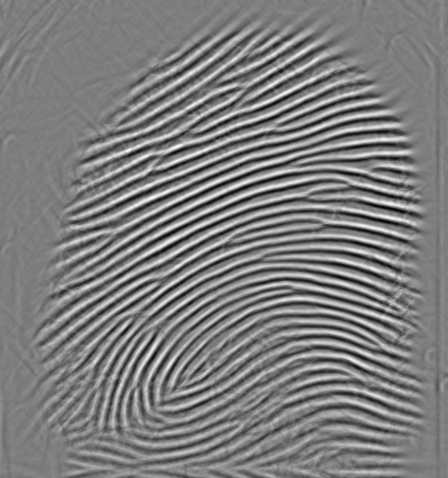} }
    
    \subfigure[Err = 5.15]{ \includegraphics[width=0.15\textwidth]{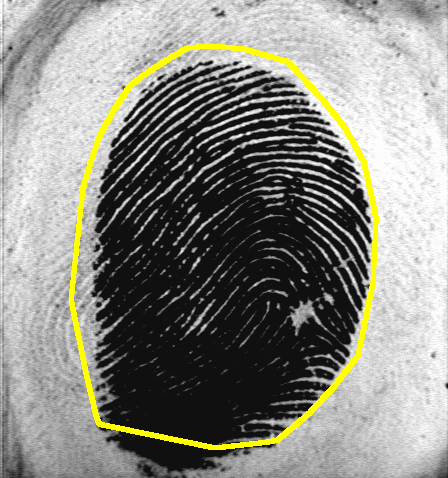} }		
    \subfigure{ \includegraphics[width=0.15\textwidth]{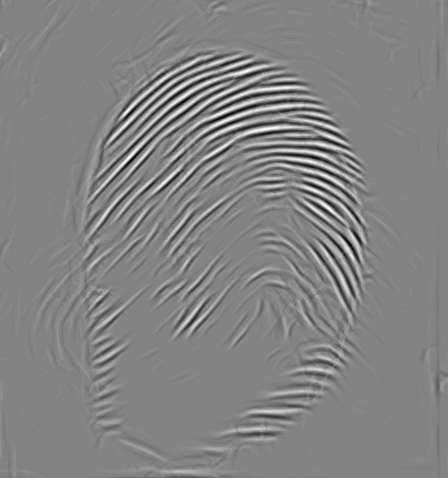} }		
    \subfigure[Err = 4.29]{ \includegraphics[width=0.15\textwidth]{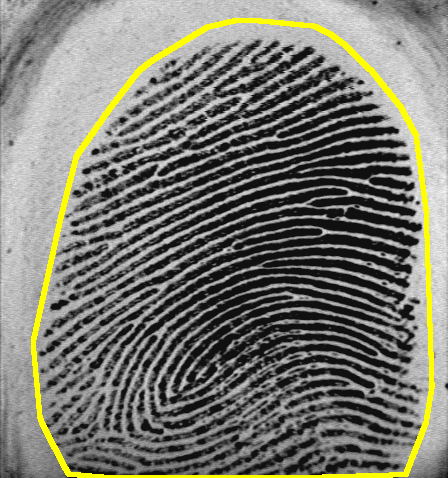} }		    
    \subfigure{ \includegraphics[width=0.15\textwidth]{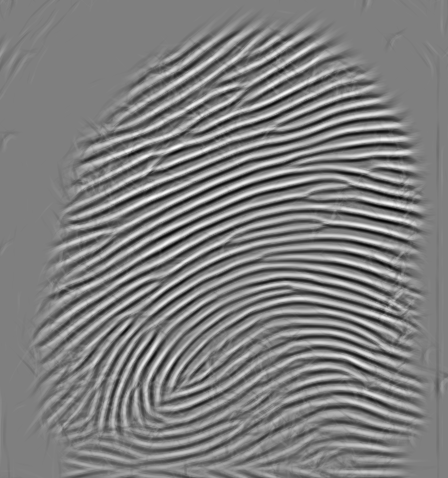} }
    
    \subfigure[Err = 8.67]{ \includegraphics[width=0.15\textwidth]{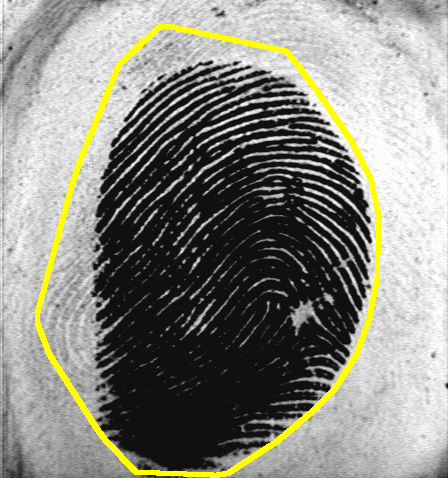} }	
    \subfigure{ \includegraphics[width=0.15\textwidth]{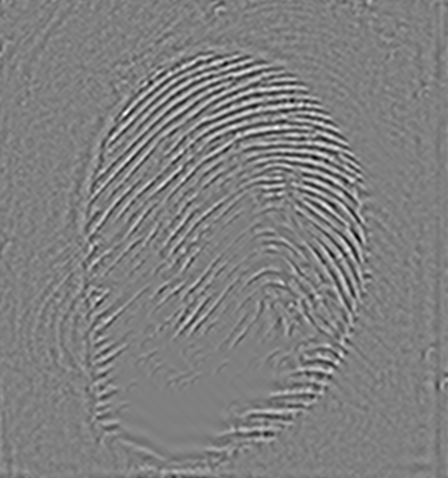} }	
    \subfigure[Err = 6.42]{ \includegraphics[width=0.15\textwidth]{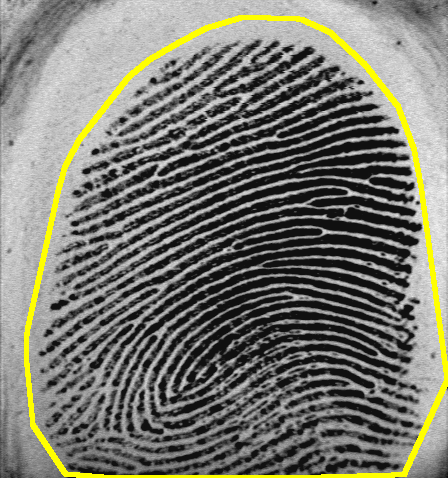} }	
    \subfigure{ \includegraphics[width=0.15\textwidth]{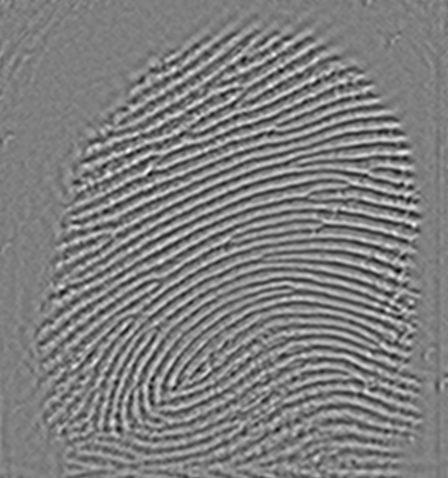} }	
    
    \subfigure[Err = 8.53]{ \includegraphics[width=0.15\textwidth]{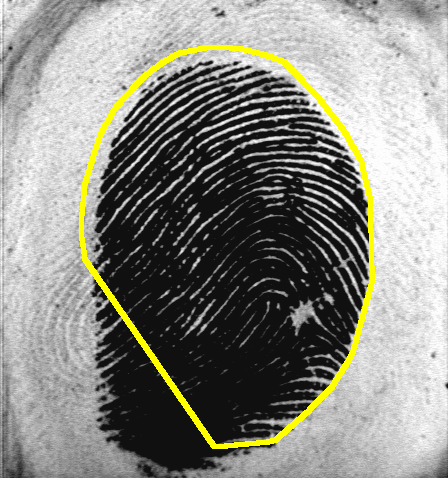} } 
    \subfigure{ \includegraphics[width=0.15\textwidth]{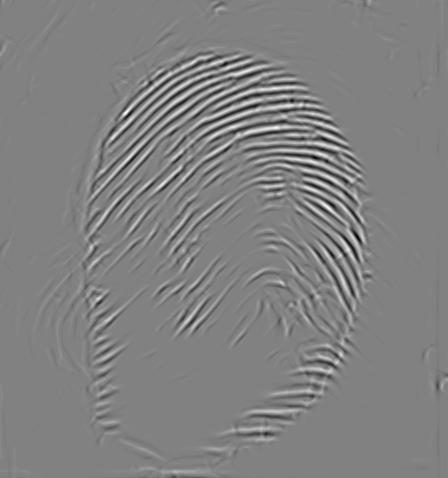} }
    \subfigure[Err = 2.82]{ \includegraphics[width=0.15\textwidth]{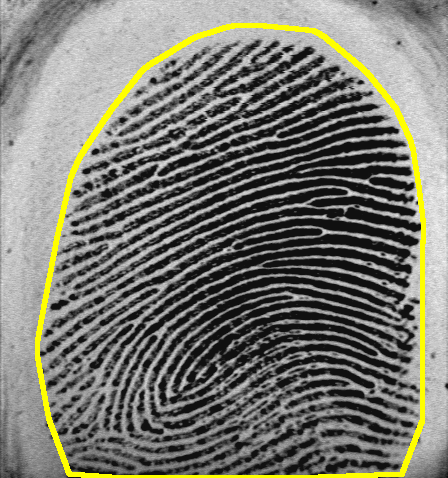} }
    \subfigure{ \includegraphics[width=0.15\textwidth]{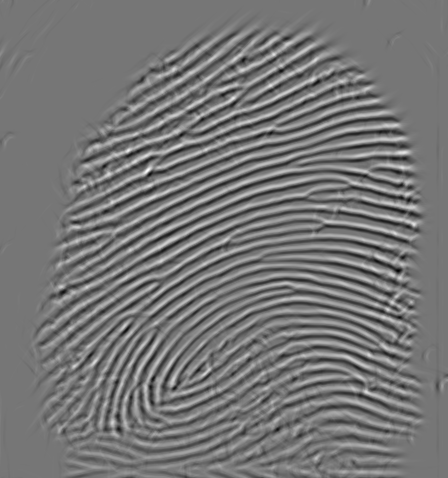} }

    \caption{Comparison of five image reconstruction strategies and their effect on the resulting segmentation.
             $1^\text{st}, 2^\text{nd} $ columns: segmented images (error in percent) 
             and reconstructed images for a low-quality image
             and $3^\text{rd}, 4^\text{th} $ columns for a good quality image.
                    $1^\text{st}$ row: the proposed operator. 
                    $2^\text{nd}, 3^\text{rd}$ rows: maximum operator without 
										and with the shrinkage operator (\ref{eq:SoftThresholding}), respectively.
                    $4^\text{th}, 5^\text{th}$ rows: summation operator without 
										and with the shrinkage operator (\ref{eq:SoftThresholding}), respectively.
    }  
    \label{figComparisonReconstructionOperator}
\end{center}
\end{figure*}

We briefly illustrate the differences between the proposed FDB filter (\ref{eq8}) and the maximum and summation operators for the coefficients 
in all directional subbands. 
Figure \ref{figComparisonReconstructionOperator} compares 
the results of these operators for a low-quality and a good quality example.
The functions are described as follows

\begin{itemize}
\item The maximum operator without and with the shrinkage operator (\ref{eq:SoftThresholding}) (depicted in the second and third row in Figure \ref{figComparisonReconstructionOperator})
	$$
	\tilde f [\boldsymbol k] = 
	\begin{cases}
	  \displaystyle \max_{l} \left\{ c_l[\boldsymbol k] \cdot ( c_l[\boldsymbol k] > 0 ) \right\}  +  \displaystyle \min_{l} \left\{ c_l[\boldsymbol k] \cdot ( c_l[\boldsymbol k] < 0 ) \right\}
	  \qquad \left( \text{without (\ref{eq:SoftThresholding})} \right)
	  \\
	  \displaystyle \max_{l} \left\{ d_l[\boldsymbol k] \cdot ( d_l[\boldsymbol k] > 0 ) \right\}  +  \displaystyle \min_{l} \left\{ d_l[\boldsymbol k] \cdot ( d_l[\boldsymbol k] < 0 ) \right\}
	  \qquad \left( \text{with (\ref{eq:SoftThresholding})} \right) ,
 	\end{cases}
	$$ 
	with $l = 0, 1, \ldots, L-1$.
\item The summation operator without and with the shrinkage operator (\ref{eq:SoftThresholding}) (displayed in the fourth and fifth row in Figure \ref{figComparisonReconstructionOperator})
	$$
	\tilde f [\boldsymbol k] = 
	\begin{cases}
	  \displaystyle \sum_{l=0}^{L-1} c_l[\boldsymbol k]	\qquad \left( \text{without (\ref{eq:SoftThresholding})} \right)
	  \\
	  \displaystyle \sum_{l=0}^{L-1} d_l[\boldsymbol k]	\qquad \left( \text{with (\ref{eq:SoftThresholding})} \right) .
	\end{cases}
	$$	 
\end{itemize}

\section{Additional Figures}

%
% Figure 5
%

%\begin{figure*}
%\begin{center}
		%\includegraphics[width=0.95\textwidth]{Fig7.pdf}
    %\caption{Image (a) displays a 1D Butterworth bandpass filter (blue) and its approximation (red).
             %Image (b) shows the 2D Butterworth bandpass filter $\widehat g^\gamma(\boldsymbol \omega)$ 
						 %at $n=20, \theta = 0, \gamma = 2$, and the 		corresponding DHBB filter in the Fourier and spatial domains (c, d)
             %for the approximation by the bilinear transform.
             %Image (e) visualizes the 2D version of the original filter 
             %and the corresponding DHBB filter (f, g).}		    
    %\label{figCompareOriginalApproximation1}
%\end{center}
%\end{figure*}

% Figure 8
\begin{figure*}
\begin{center}
 \subfigure[]{\includegraphics[width=0.2\textwidth]{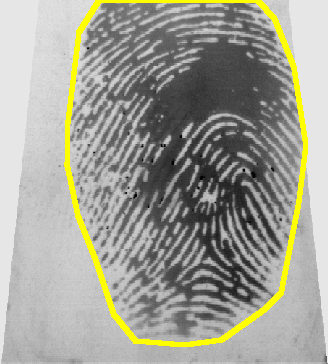}}
 \subfigure[]{\includegraphics[width=0.2\textwidth]{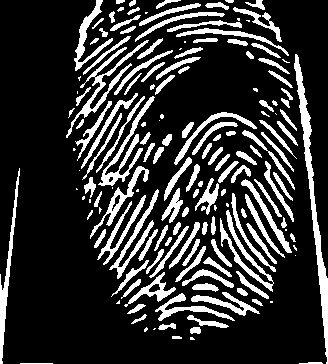}}
 \subfigure[]{\includegraphics[width=0.2\textwidth]{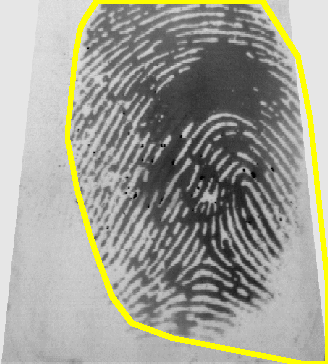}}
 \subfigure[]{\includegraphics[width=0.2\textwidth]{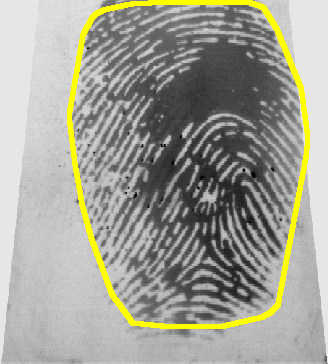}}
 \caption{		
    The ground truth segmentation (a) and the binarized texture image (b) for an example fingerprint.
    Applying a standard morphology operation like closing (dilation followed by erosion)     
    instead of the proposed method connects in this example the 
    white fingerprint texture with structure noise close to the margin of the texture 
    and the result is a defective segmentation (c). 
    The proposed morphology avoids this undesired effect by considering neighborhoods
    on two scales: cells of size $s \times s$ pixels and blocks of $3 \times 3$ cells.
    \label{figMorphology}}
\end{center}
\end{figure*}

% Figure 9

\begin{figure*}
\begin{center}
    \subfigure{ \includegraphics[width=0.15\textwidth]{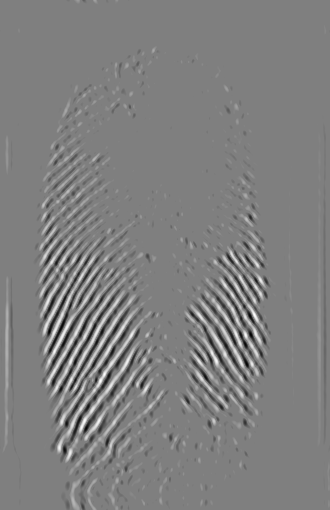} }		
    \subfigure{ \includegraphics[width=0.15\textwidth]{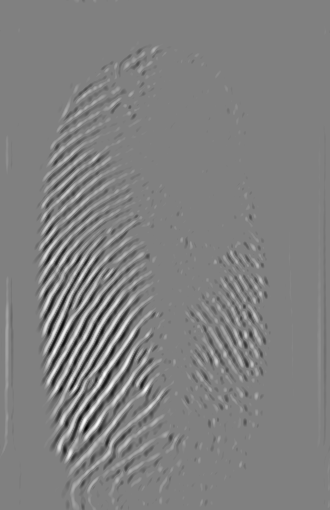} }		
    \subfigure{ \includegraphics[width=0.15\textwidth]{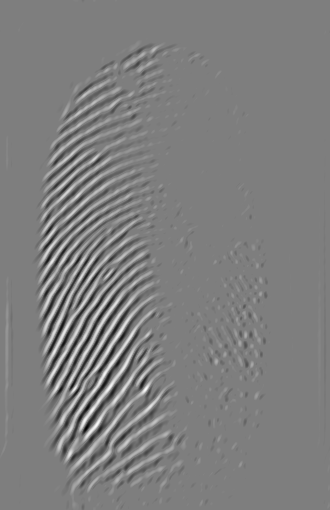} }		
    \subfigure{ \includegraphics[width=0.15\textwidth]{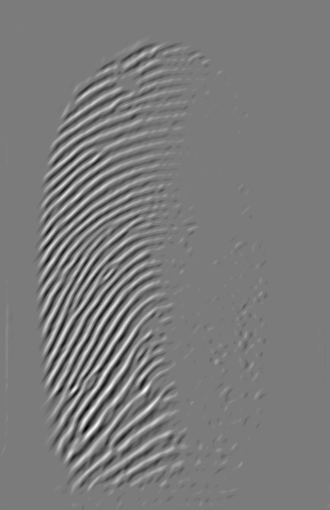} }		
    \\
    \subfigure{ \includegraphics[width=0.15\textwidth]{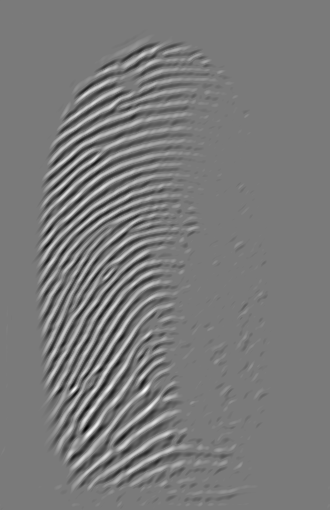} }		
    \subfigure{ \includegraphics[width=0.15\textwidth]{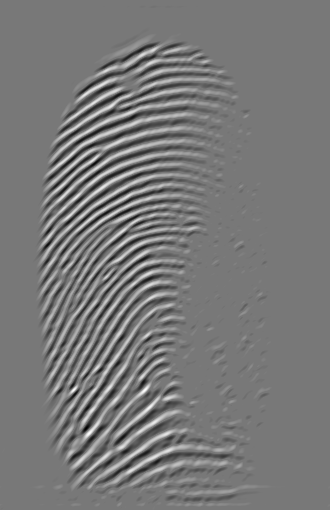} }		
    \subfigure{ \includegraphics[width=0.15\textwidth]{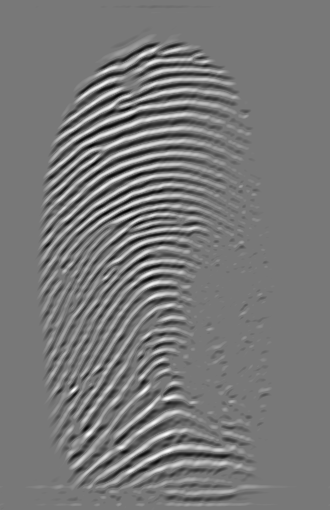} }		
    \subfigure{ \includegraphics[width=0.15\textwidth]{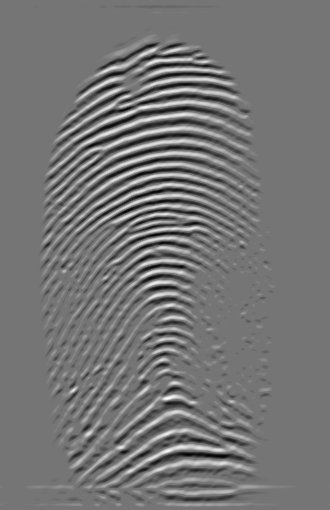} }		
    \\
    \subfigure{ \includegraphics[width=0.15\textwidth]{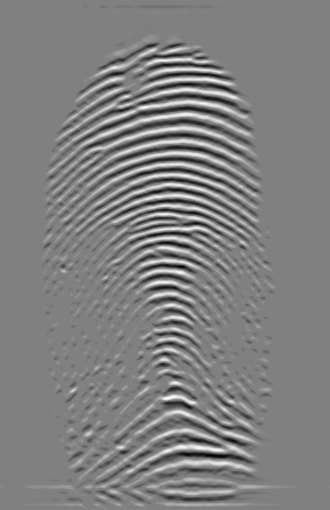} }		
    \subfigure{ \includegraphics[width=0.15\textwidth]{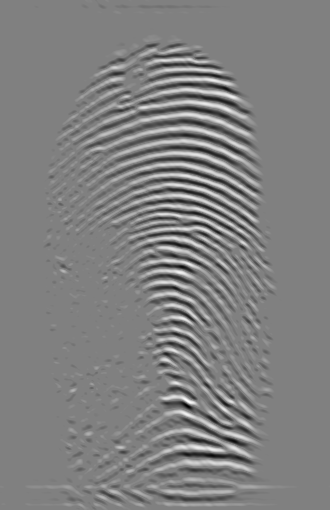} }		
    \subfigure{ \includegraphics[width=0.15\textwidth]{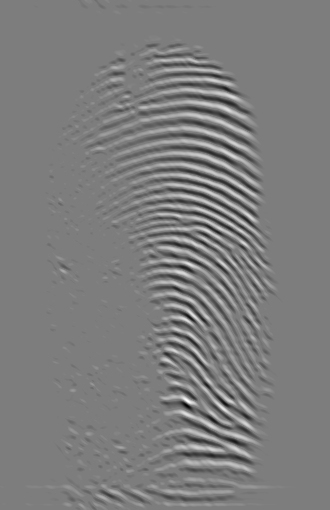} }		
    \subfigure{ \includegraphics[width=0.15\textwidth]{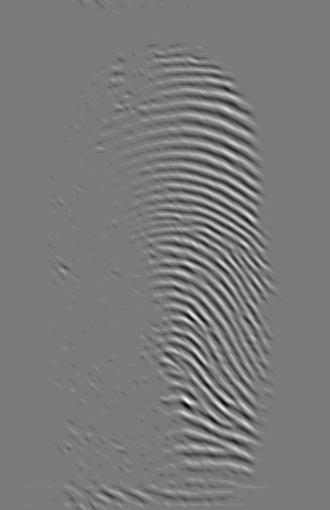} }		
    \\
    \subfigure{ \includegraphics[width=0.15\textwidth]{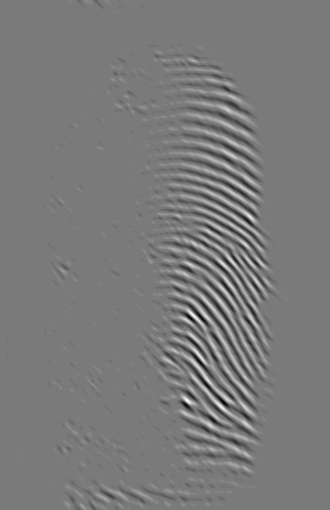} }		
    \subfigure{ \includegraphics[width=0.15\textwidth]{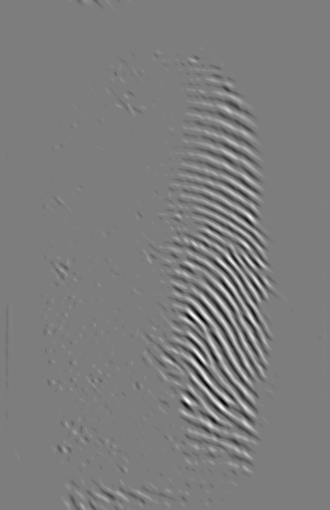} }		
    \subfigure{ \includegraphics[width=0.15\textwidth]{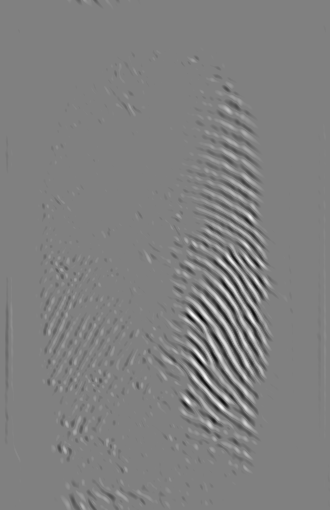} }		
    \subfigure{ \includegraphics[width=0.15\textwidth]{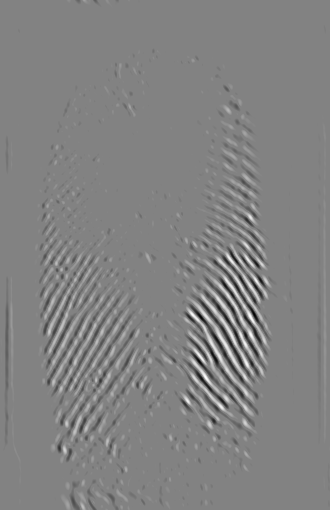} }		
    \caption{Visualization of the coefficients in the 16 subbands 
             of the DHBB filter for 
              $n = 20, \gamma = 3, \omega_L = 0.3, \omega_H = 1$. }    
    \label{fig_16directionCoeff}    
\end{center}
\end{figure*}

\end{document}